\documentclass[lettersize,journal]{IEEEtran}

\usepackage{algorithm}
\usepackage{algorithmic}
\usepackage{amsmath}
\usepackage{amssymb}
\usepackage{booktabs}
\usepackage{graphicx}
\usepackage{mathtools}
\usepackage{multirow}
\usepackage{stfloats}
\usepackage{subfigure}
\usepackage{float}
\usepackage{graphicx}
\usepackage{soul,color,xcolor}
\usepackage{lineno}
\pdfoutput=1
\UseRawInputEncoding

\def\BibTeX{{\rm B\kern-.05em{\sc i\kern-.025em b}\kern-.08em
    T\kern-.1667em\lower.7ex\hbox{E}\kern-.125emX}}
\usepackage{balance}
\begin{document}

\title{\LARGE \bf
A Multi-Modal Approach Based on Large Vision Model for Close-Range Underwater Target Localization}
\author{Mingyang Yang, Zeyu Sha and Feitian Zhang$^*$
\thanks{Mingyang Yang and Zeyu Sha are with the Department of Advanced Manufacturing and Robotics, College of Engineering, Peking University, Beijing, 100871, China,
        {\tt\small mingyangyang@stu.pku.edu.cn} and {\tt\small schahzy@stu.pku.edu.cn}}%

\thanks{Feitian Zhang is with the Department of Advanced Manufacturing and Robotics, and the State Key Laboratory of Turbulence and Complex Systems, College of Engineering, Peking University, Beijing, 100871, China,
        {\tt\small feitian@pku.edu.cn}}%
\thanks{* Send all correspondence to Feitian Zhang.}
}

\maketitle

\begin{abstract}

Underwater target localization uses real-time sensory measurements to estimate the position of underwater objects of interest, providing critical feedback information for underwater robots in tasks such as obstacle avoidance, scientific exploration, and environmental monitoring. While acoustic sensing is the most acknowledged and commonly used method in underwater robots and possibly the only effective approach for long-range underwater target localization, such a sensing modality generally suffers from low resolution, high cost and high energy consumption, thus leading to a mediocre performance when applied to close-range underwater target localization. On the other hand, optical sensing has attracted increasing attention in the underwater robotics community for its advantages of high resolution and low cost, holding a great potential particularly in close-range underwater target localization. However, most existing studies in underwater optical sensing are restricted to specific types of targets, thus lacking generalization capabilities. In addition, these studies typically focus on the design of estimation algorithms and ignore the influence of illumination conditions on the sensing performance, thus hindering wider applications in the real world. To address the aforementioned issues, this paper proposes a novel target localization method that assimilates both optical and acoustic sensory measurements to estimate the 3D positions of close-range underwater targets. The proposed sensing method integrates a large vision model with unique acoustic-based model prompt design to process multi-modal sensor measurements, ensuring the generalizability and robustness of underwater target localization. A test platform with controllable illumination conditions is developed.  Extensive experiments are conducted, the results of which validate the effectiveness of the proposed method.

\end{abstract}


\section{Introduction}
In recent years, a variety of underwater robots have been designed and developed by researchers and engineers worldwide \cite{zhou2023review}, giving birth to a revolutionary paradigm of marine robotics. As a result of technological advancements, the applications of underwater robots grow rapidly covering missions and tasks across the scientific, industrial and military domains \cite{wynn2014autonomous}. Particularly, close-range marine tasks such as marine life monitoring \cite{RAL2022point}, shipwreck surveying \cite{bingham2010robotic}, sea mining exploration \cite{seamining}, searching and rescuing \cite{venkatesan2016auv} become achievable. To successfully accomplish these tasks, accurate target perception, particularly target localization, is the essential cornerstone.
\par To date, many underwater sensing methods have been designed to solve the close-range target localization problem \cite{zhufeng2022present}, \cite{yuan2022survey}. Among these methods, acoustic sensing and optical sensing are the two mainstream types of sensory modalities \cite{cong2021underwater}. Acoustic sensing with sonars is the most acknowledged and commonly-used approach in underwater robots and possibly the only feasible approach for long-range sensing tasks. For instance, Wang et al. \mbox{\cite{acousticmarker}} proposed a method to recognize a specifically designed marker and estimate its relative pose using a forward-looking sonar. However, in close-range scenarios, sonars typically have insufficient resolution and may not provide sufficient details about the target. Furthermore, the cost and compatibility of sonar systems with different types of  underwater robots, especially small-sized autonomous underwater robots, are additional considerations \mbox{\cite{sun2021review}}. 

To resolve the close-range underwater target sensing problem, optical sensing has grasped a growing attention in the science and engineering communities due to its appealing features of high resolution and low cost. Although inappropriate for long-range underwater sensing due to the rapid light attenuation and light refraction \cite{huy2023object}, optical sensors are perfectly fit for close-range underwater tasks. For instance, Wang et al. \mbox{\cite{cameraIEEE}} designed an underwater onboard vision system with a lightweight object detect
or for underwater robotic gripping.

Vision-based underwater target localization methods are classified into two main categories including classical methods and convolutional neural networks (CNNs)-based methods \cite{xu2023systematic}. Classical methods are typically effective when specific shapes or colors are required. For instance, Meng et al. \cite{color-based} leveraged color-based target segmentation and achieved underwater target following for robotic manta. In recent years, CNN-based methods have gained rapidly increasing attention due to advancements in CNN structures and computational resources. These methods have been applied across various domains, including underwater tasks. For example, Sapienza et al. \cite{RAL2023POS} proposed a pipeline leveraging the You Only Look Once (YOLO) model and the augmented autoencoder (AAE) to compute 6-D pose estimates of underwater targets from 2-D images. Furthermore, researchers have investigated and designed a number of target localization methods for close-range underwater targets with dynamic motions \cite{2021Recent}. Wolek et al. \cite{wolek2020tracking} designed and tested a multi-target tracker to actively track nearby surface vessels using a passive sonar. Langis et al. \cite{de2020realtime} proposed a multi-diver tracking method that used camera images to detect human divers and estimate their dynamic motion states. 

 While the design of target localization using a single sensing modality, e.g., acoustic or optical sensing, suffices in estimating the motion states of underwater targets, employing multiple sensory modalities usually leads to higher estimation performances, which attracts a rapidly growing interest within the research community. For example, Remmas et al. \cite{remmas_diver_2021} designed a data fusion scheme using a monocular camera, distributed hydrophones and pressure sensors and achieved accurate tracking of human divers. Jiang et al. \cite{pressureaddflow} proposed a dual-sensor fusion modality integrating pressure sensors and flow velocity sensors to locate a near-field dipole source.

Whereas various underwater target localization approaches both in single and multiple modalities have been designed and tested, there still remain several challenging problems unresolved. First, most of the existing literature focuses on the network architecture design of deep learning models without considering the influence of the illumination condition on the target localization performance. With extremely low illumination, optical sensing is most likely unable to provide sufficiently accurate target estimates. Second, most of the methods are limited to a specific type of task and target. Current camera-based and sonar-based methods require either specially designed markers or large amounts of training data \mbox{\cite{islam2020semantic}}. However, we don't have the luxury of large-scale datasets of underwater targets, thus significantly impeding the wide application of the existing deep learning-based approaches.

\begin{figure}[thpb]
 \centering
 \includegraphics[width=0.48\textwidth]{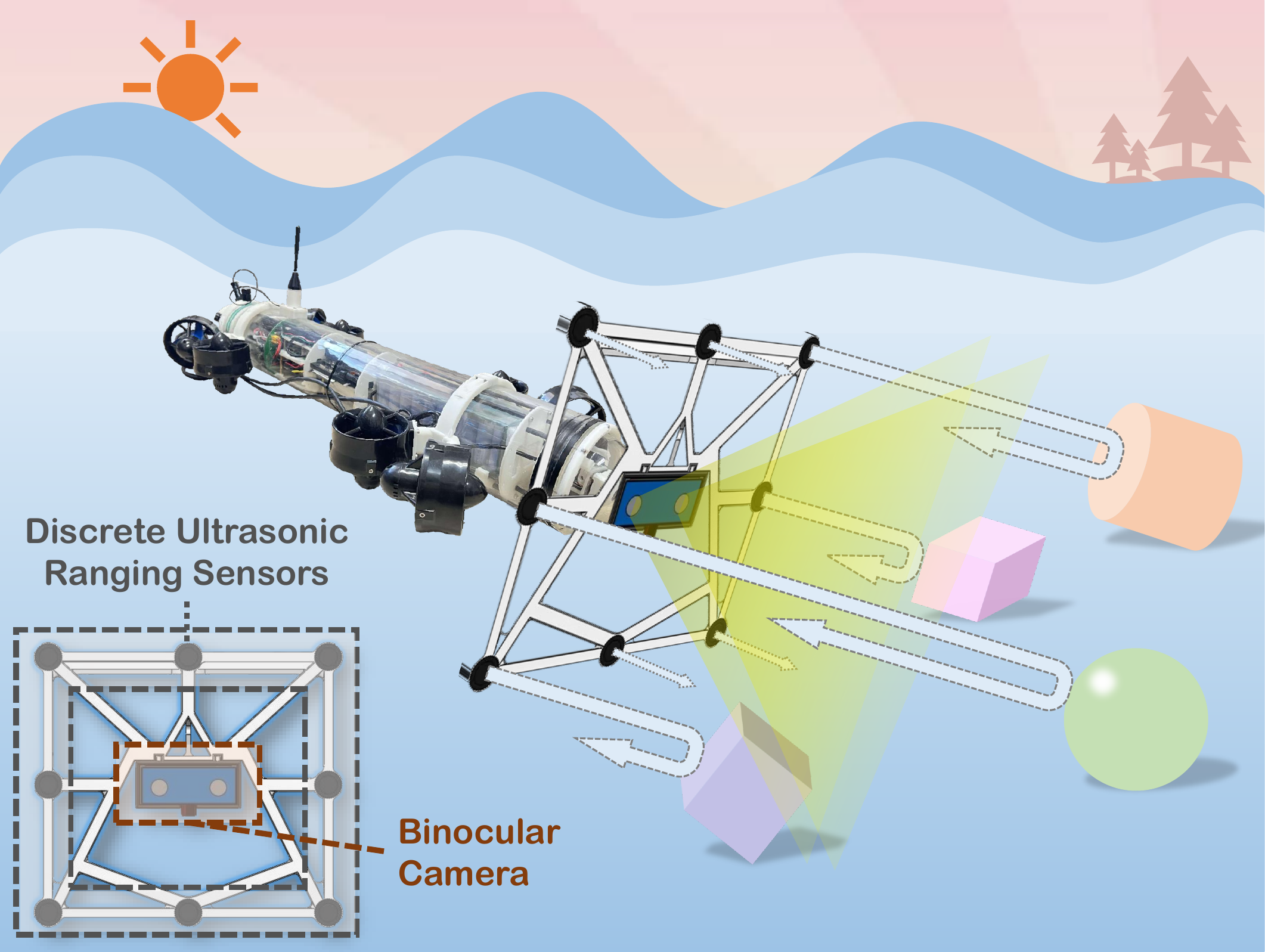}
 \caption{The schematic of the proposed multi-modal close-range target localization framework for underwater robots. When an underwater target appears within the sensor measurement range, multiple optical and acoustic sensors equipped onboard the underwater robot collaboratively estimate the motion states of the target of interest.}
\label{IntroFig}
\end{figure}
\par To address the issues mentioned above, this paper proposes a multi-modal sensing framework (illustrated in Fig. \ref{IntroFig}) that fuses real-time acoustic and optical sensory measurements utilizing a large vision model to achieve a general sensing capability of close-range underwater target localization. To validate the proposed approach, a test platform is designed and constructed with controllable lighting conditions. The test platform consists of a binocular camera taking the advantage of its high-precision and low-cost attributes \cite{huo2018underwater} along with a number of distributed single-beam ultrasonic ranging sensors. 
To segment objects from underwater images, we employ a large vision model --- the Segment Anything Model (SAM) \mbox{\cite{kirillov2023segment}}, which adopts the Transformer architecture \mbox{\cite{vaswani2017attention}}, in our scenario. Trained on the exhaustive SA-1B dataset with over 1 billion masks on 11 million images, SAM is a milestone model in vision history with the ability to segment any object in an image with proper prompt through user interaction \mbox{\cite{zhao2023fast}}.
The emergence of the large vision models sheds light on a new way to solve the underwater target localization problem. Several researches have been attempting to adapt SAM under different scenarios. For instance, Chen et al. \mbox{\cite{chen2023sam}} incorporated domain-specific visual prompts into SAM's segmentation network and proposed simple but effective SAM-Adapter, achieving improved results on medical images. Zhang et al. \mbox{\cite{zhang2023personalize}}  designed a training-free model selector with one-shot image to customize SAM for specific applications, demonstrating significant effectiveness on video segmentation benchmarks.
This paper investigates the feasibility and evaluates the performance of applying SAM to the close-range underwater target localization with zero-shot transfer. Extensive experiments are conducted and experimental results are presented to confirm the multi-modal sensory design and the robustness of the proposed estimation algorithm with respect to illumination conditions. 

The contributions of the paper are twofold.
First, this paper proposes a novel multi-modal sensing method that incorporates a large vision model (SAM) to assimilate acoustic and optical sensory measurements for close-range underwater target localization. Owing to the superior generalization capability of the large vision model, the proposed method is expected to achieve an enhanced robust sensing performance with respect to various underwater targets with no training data required. 
Second, differing from most of the existing studies, this paper takes the illumination variance into consideration and conducts extensive experiments to quantitatively investigate and evaluate the influence of illumination conditions on the performance of the designed target localization algorithm.

\section{Test Platform}
\begin{figure}[thpb]
\centering
\label{testplatform}
\subfigure[Sensing module.]{
\label{sensingmodule}
\includegraphics[width=0.15\textwidth]{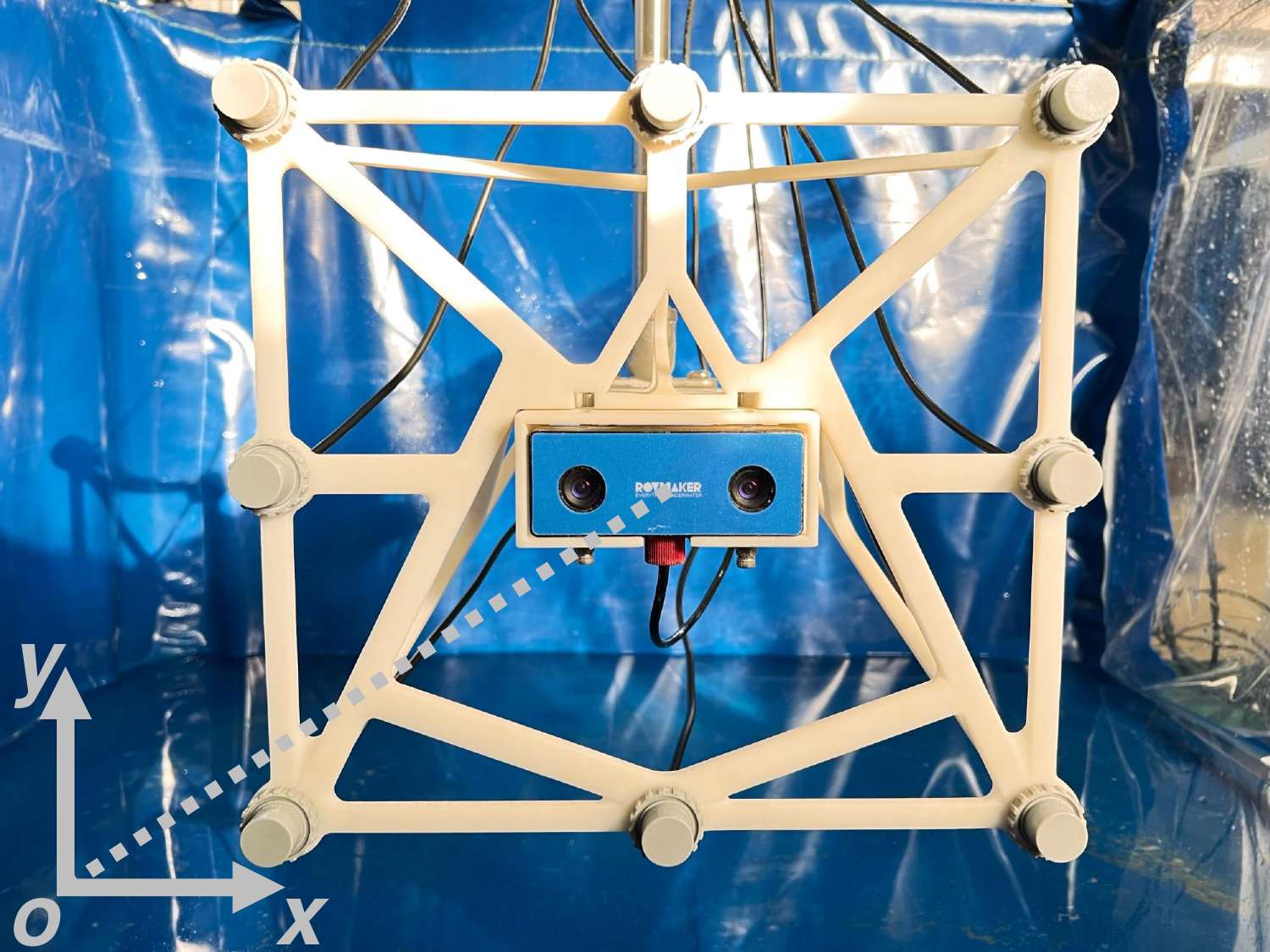}}
\subfigure[Regular-shaped targets.]{
\label{targets}
\includegraphics[width=0.15\textwidth, angle=0]{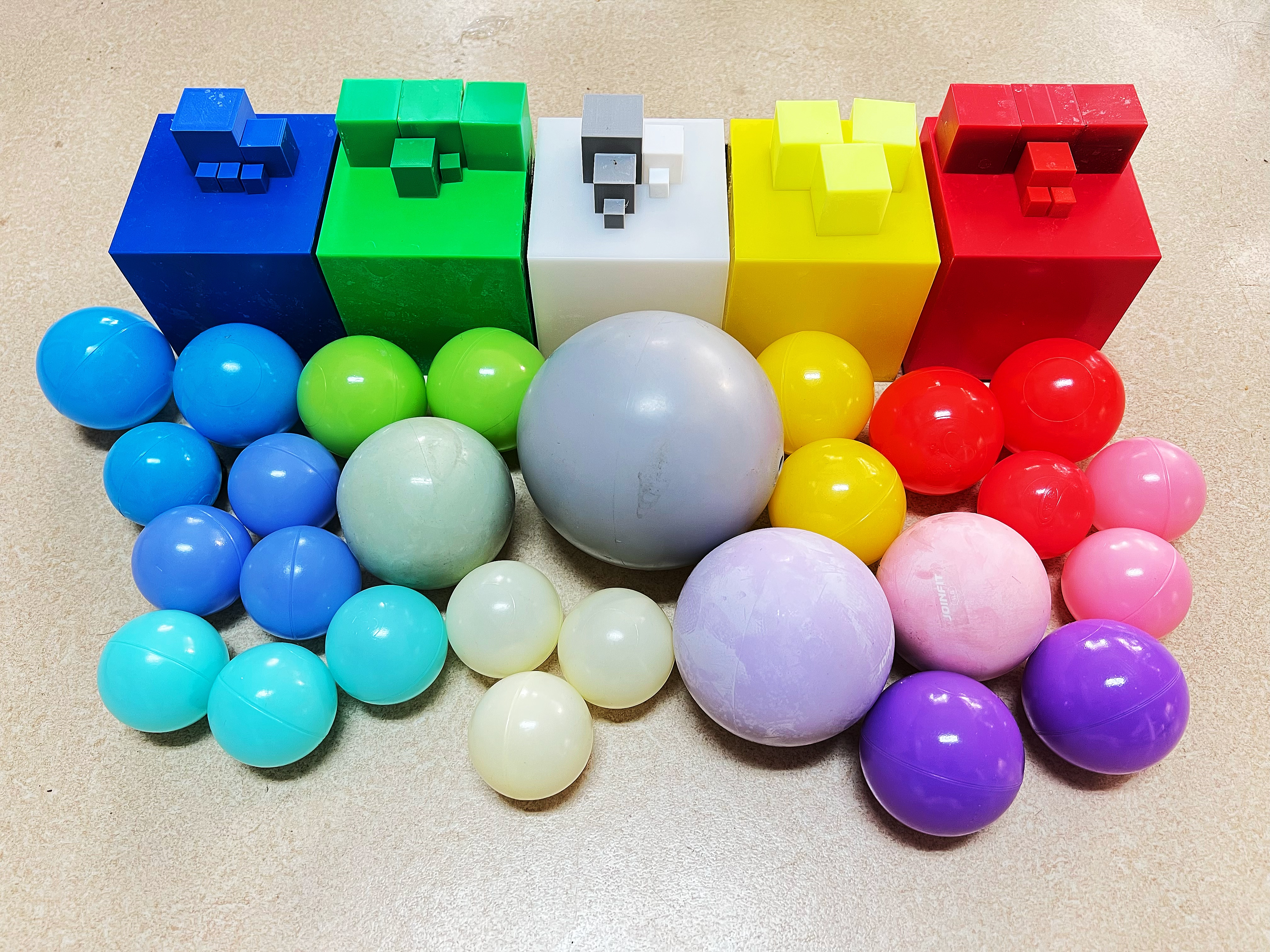}}
\subfigure[Sea animal figure-shaped targets.]{
\label{marinelife}
\includegraphics[width=0.15\textwidth, angle=0]{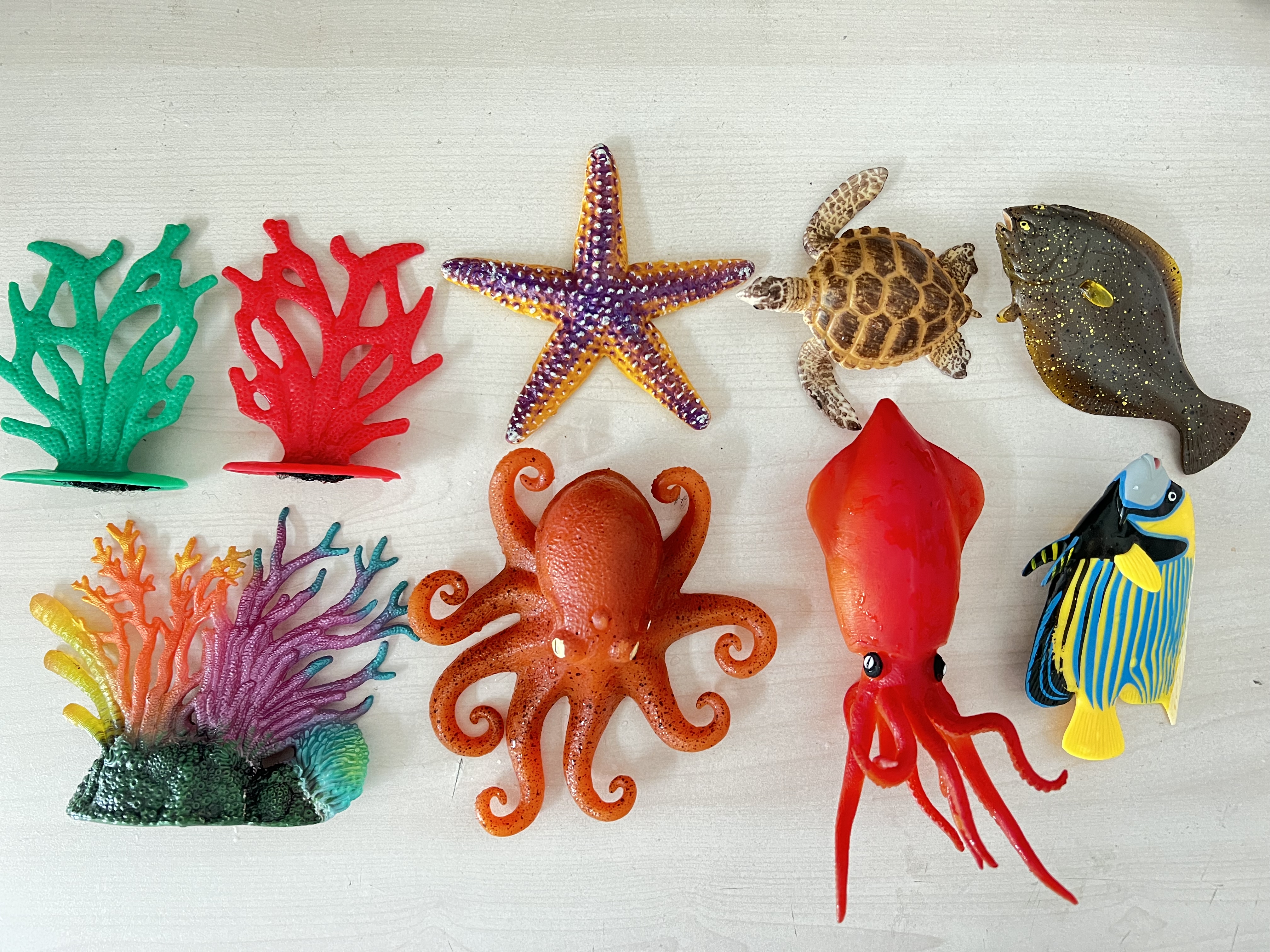}}
\subfigure[Layout of the test platform.]{
\label{slipway}
\includegraphics[width=0.475\textwidth]{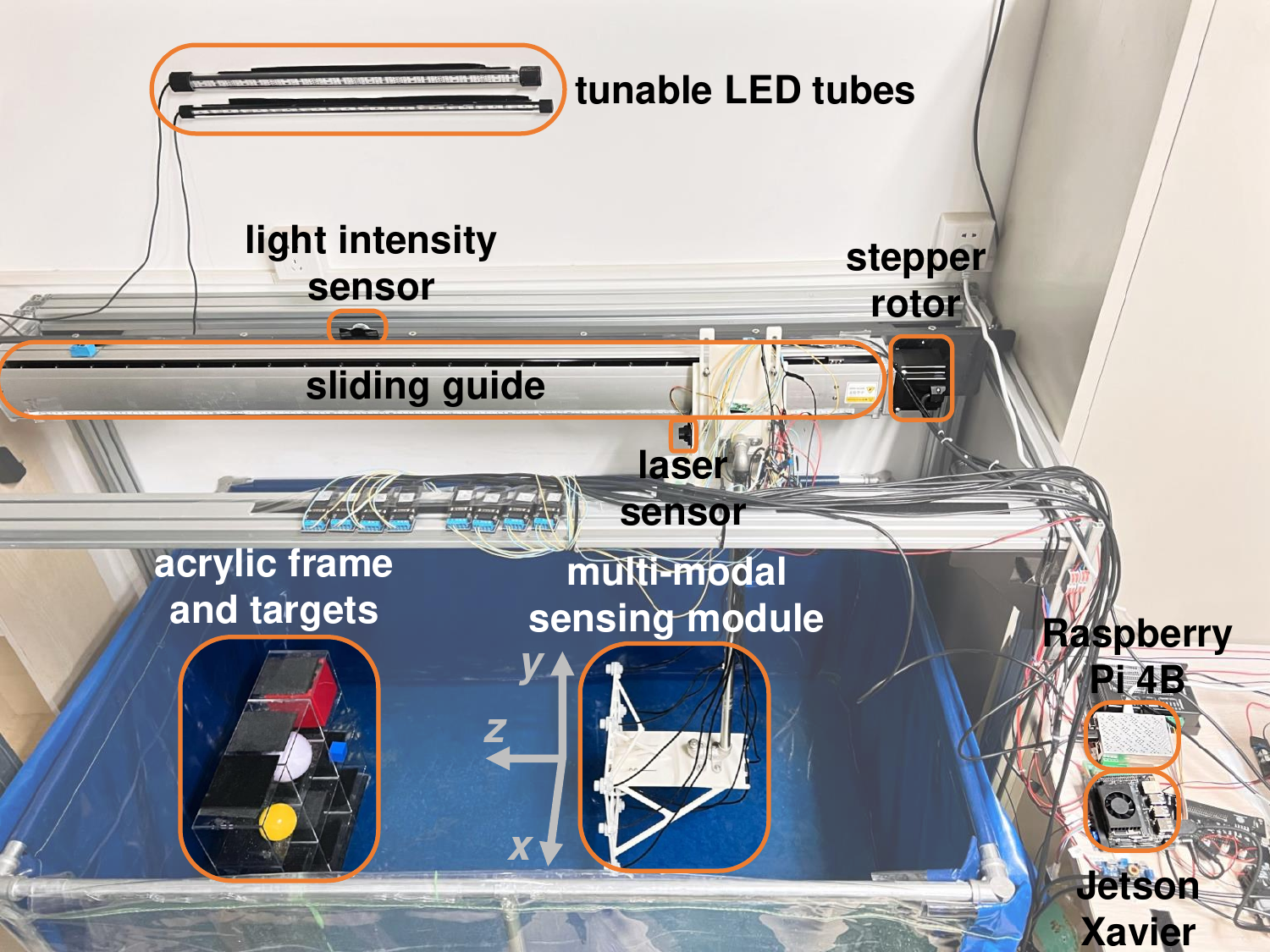}}
\caption{Illustration of the test platform and the relevant components.}
\label{PhyicalExperiment}
\end{figure}

This paper designs and constructs a test platform to investigate the close-range target localization problem which consists of a sensing module, a target module, and a test pool. 
The sensing module, shown in Fig. \ref{sensingmodule}, is comprised of a 3D-printed square-shaped support frame, a binocular camera located at the center of the frame, and eight acoustic ranging sensors located at the four corners and the four midpoints of the edges. We adopt binocular camera from ROVMAKER, supporting several resolutions, with 2560*960 as the highest resolution. The acoustic ranging sensors are the ultrasonic L04 modules by DYP Sensor.
The target module consists of an acrylic frame that holds regular-shaped and sea animal figure-shaped targets. The regular-shaped targets encompass spherical and cubic objects of varying sizes and colors, illustrated in Fig.~\mbox{\ref{targets}}. Additionally, the sea animal figure-shaped targets encompass a diverse range of marine creatures, including a flounder, a butterfly fish, a starfish, a turtle, an octopus, a squid, as well as various types of coral such as green, red, and multi-color coral, illustrated in Fig.~\mbox{\ref{marinelife}}.
The test pool measures 1.5m long, 1m wide and 0.7m deep. Two tunable LED tubes are attached onto the wall above the pool to control the illumination condition of the testing underwater environment. A HOBO MX2202 light intensity sensor is selected and mounted over against the LED tubes, to measure/record the experimental illumination condition.

Along the $z$-axis in Fig. 2, we ensure accurate position control with \mbox{$\pm0.02$~mm} accuracy using stepper motor, specifically Model 86BYG with 4.1N·m holding torque from Mecheltron. At the same time, we apply a laser distance sensor to measure the distance between an L-shape extension beam, mounted on top of the acrylic target frame and exposed to air, and the sensing module at a frequency of 120Hz. The laser sensor module used is the Point LiDAR STP-23L from LDROBOT. With accurate position control via the stepper motor and real-time distance measurement through laser sensor, the measurement precision of the ground-truth data is ensured.

A Jetson Xavier is adopted to collect the sensor measurements and process the data for target localization and tracking in real time.

\section{Sensor Model Preliminaries}
This section defines the variable notations of the camera models used in this paper. The camera models include the pinhole model, the distortion model and the binocular model specifically.

\subsection{Pinhole Camera Model}
Following the conventions in camera modeling \mbox{\cite{szeliski2022computer}}, for a point $\textbf{\textit{P}}$, its coordinates in the world reference frame, the camera reference frame and the pixel reference frame are defined by
$\textbf{\textit{P}}_\text{w}=[X_\text{w},Y_\text{w},Z_\text{w}]^\text{T}$, $\textbf{\textit{P}}_\text{c}=[X_\text{c},Y_\text{c},Z_\text{c}]^\text{T}$ and $\textbf{\textit{P}}_\text{p}=[u,v,1]^\text{T}$, respectively. The transformation relations is calculated as
\begin{equation}
\textbf{\textit{P}}_\text{c} = \textbf{\textit{R}}_\text{c}\textbf{\textit{P}}_\text{w}+\textbf{\textit{t}}_\text{c},
\end{equation}
\begin{equation}
\textbf{\textit{P}}_\text{p}=\frac{1}{Z_\text{c}}\begin{bmatrix}
\textit{f}_\textit{x}&\textit{0}&\textit{c}_\textit{x}\\
\textit{0}&\textit{f}_\textit{y}&\textit{c}_\textit{y}\\
\textit{0}&\textit{0}&\textit{1}
\end{bmatrix}\textbf{\textit{P}}_\text{c}=\frac{1}{Z_\text{c}}\textbf{\textit{KP}}_\text{c},
\end{equation}
where $\textbf{\textit{R}}_\text{c}$ and $\textbf{\textit{t}}_\text{c}$ are the camera extrinsics, $\textbf{\textit{K}}$ is the camera intrinsic matrix, $[c_{x}, c_{y}]^\text{T}$ represents the principal point, $[f_{x}, f_{y}]^\text{T}$ represents the focal lengths, respectively. Ignoring the manufacturing flaws and calibration errors, $f_{x}$ is equal to $f_{y}$ and referred to as $f$ in the following sections.

\subsection{Distortion Model}
\label{imagerectify}
Image distortion occurs with the presence of a lens which alters the light propagation path. There are two types of distortion, namely radial distortion and tangential distortion. Let $\textbf{\textit{D}}_\text{r} = [k_1, k_2, k_3]$ and $\textbf{\textit{D}}_\text{t}=[p_1, p_2]$ represent the radial distortion parameters and the tangential distortion parameters, respectively. Define distortion parameter vector $\textbf{\textit{D}}=[\textbf{\textit{D}}_\text{r}, \textbf{\textit{D}}_\text{t}]$.

\subsection{Binocular Camera Model}
A binocular camera consists of two monocular cameras, namely the left view camera and the right view camera. Each monocular camera has its own intrinsic matrix $\textbf{\textit{K}}_\text{L}$, $\textbf{\textit{K}}_\text{R}$ and distortion parameters $\textbf{\textit{D}}_\text{L}=[\textbf{\textit{D}}_\text{L,r}, \textbf{\textit{D}}_\text{L,t}]$, $\textbf{\textit{D}}_\text{R}=[\textbf{\textit{D}}_\text{R,r}, \textbf{\textit{D}}_\text{R,t}]$. The extrinsic parameters, the rotation matrix $\textbf{\textit{R}}_\text{b}$ and the translation vector $\textbf{\textit{t}}_\text{b}$, are necessary to describe the relative attitude and position between two monocular cameras.

Define $b$ as the baseline of the stereo system. For a point $\textbf{\textit{P}}$, $\textbf{\textit{P}}_\text{L}$ and $\textbf{\textit{P}}_\text{R}$ are the coordinates of $\textbf{\textit{P}}$ in the left and right images, respectively. 
The horizontal coordinates of $\textbf{\textit{P}}_\text{L}$ and $\textbf{\textit{P}}_\text{R}$ are $u_{\text{L}}$ and $-u_{\text{R}}$. The target distance $Z$ is calculated as
\begin{equation}
\label{calcdistance}
Z=\frac{fb}{u_{\text{L}}-u_{\text{R}}}=\frac{fb}{d_u},
\end{equation}
where $d_u$ is the binocular disparity.

Due to calibration and localization error, a difference in the vertical coordinates exists, which is defined as $d_\text{v}$. Define epipolar tolerance $\varepsilon$ as the maximum difference in pixels allowed along the vertical direction. The epipolar matching condition (EMC) is satisfied if
\begin{equation}
\label{EMC}
d_v=|v_\text{L}-v_\text{R}|<\epsilon,
\end{equation}
where $v_\text{L}$ and $v_\text{R}$ are the vertical coordinates of $\textbf{\textit{P}}_\text{L}$ and $\textbf{\textit{P}}_\text{R}$.

\subsection{Ultrasonic Ranging Sensor Model}
The ultrasonic sensor measures distance by utilizing the time of flight (ToF) principle, which calculates the time taken for an ultrasonic wave to travel from the transmitter to the target and back to the receiver, governed by the ToF model, expressed as \mbox{\cite{qiu2022review}}
\begin{equation}
    s = c \Delta t/2
\end{equation}
where $s$ represents the measured distance, $c$ is the velocity of the ultrasonic wave in the medium, and $\Delta t$ is the time interval measured between the transmission and reception of the ultrasonic signal.

\section{Underwater Target Localization Design}
The proposed multi-modal target localization algorithm is illustrated in Algorithm \ref{algo-flow}. The algorithm takes the video stream from the binocular camera and the ranging data stream from the acoustic sensors as inputs to calculate the positions and velocities of the targets of interest in real time. To avoid ambiguity, we use \textit{frame} to refer to the stereo image with both views at a given time step and \textit{image} to refer to either the left or the right view image whose width is half width of a frame.

\floatname{algorithm}{Algorithm}
\begin{algorithm}[!h] 
\caption{Multi-Modal Target Localization.}
\label{algo-flow}
\renewcommand{\algorithmicrequire}{\textbf{Input:}}
\renewcommand{\algorithmicensure}{\textbf{Output:}}
\begin{algorithmic}[1]
\REQUIRE Real-time binocular video frame $I_\text{b}$.
\REQUIRE Real-time ranging data $I_\text{r}$.
\ENSURE 3-D position and velocity of the target(s), denoted as $\textbf{\textit{P}}=[p_x, p_y, p_z]^\text{T}$ and $\textbf{\textit{V}}=[v_x, v_y, v_z]^\text{T}$ respectively.
\STATE Initialize hyper-parameters: intrinsic matrices $\textbf{\textit{K}}_\text{L}/\textbf{\textit{K}}_\text{R}$, distortion parameters $\textbf{\textit{D}}_\text{L}/\textbf{\textit{D}}_\text{R}$, extrinsic parameters $\textbf{\textit{R}}_\text{b}$ and $\textbf{\textit{t}}_\text{b}$, segmentation confidence threshold $c_\text{s}^{\text{th}}$, epipolar tolerance $\varepsilon$, ranging data threshold $r_\text{max}^\text{th}$ and  $r_\text{min}^\text{th}$, fusion weight $\alpha$, process/measurement noise covariance matrix $\textbf{\textit{Q}}/\textbf{\textit{R}}$.
\WHILE{new measurement data}
\IF{ranging data $> r_\text{max}^\text{th}$ or $< r_\text{min}^\text{th}$}
\STATE Ineffective data. Implement extrapolation.
\ENDIF
\STATE Perform rectification using $\textbf{\textit{K}}_\text{L}$, $\textbf{\textit{K}}_\text{R}$, $\textbf{\textit{D}}_\text{L}$, $\textbf{\textit{D}}_\text{R}$, 
$\textbf{\textit{R}}_\text{b}$ and $\textbf{\textit{t}}_\text{b}$.
\STATE Perform instance segmentation by SAM using point prompts per Section~\mbox{\ref{mmloc}}.
\IF{mask confidence $< c_\text{s}^{\text{th}}$}
\STATE Image segmentation fails. Implement extrapolation.
\ENDIF
\STATE Calculate vertical disparities of key point pairs $\{d_v^i, i=1,...,5\}$ per Section~\mbox{\ref{mmloc}}.
\IF{$\exists \,i \text{ such that }d_v^i > \varepsilon$}
\STATE Frame segmentation fails. Implement extrapolation.
\ENDIF
\IF{only target ranging is desired}
\STATE Implement weighted average per Section~\mbox{\ref{sec:weightavg}}
\ELSIF{3-D target position and velocity are desired}
\STATE Implement EKF per Section~\mbox{\ref{sec:EKF}}
\ENDIF
\ENDWHILE
\end{algorithmic}
\end{algorithm}

\subsection{Target Detection With Large Vision Model and Ranging Sensor Prompts}
\label{mmloc}
The proposed sensing module consists of a binocular camera and several ultrasonic ranging sensors. The binocular camera is selected as the primary sensor for close-range target localization while ultrasonic ranging sensors provide rough inference of the target location. Comprehensively considering the sensing accuracy performance and the practical limitations in the computational resource and the transmission bandwidth, the number of the ultrasonic ranging sensors is selected to achieve the balance therebetween, specifically eight sensors in this paper as an example.

This paper adopts a large vision model --- SAM in the target detection process aiming to achieve a high generalization capability with no training data required. SAM is applicable either with or without input prompts. Without users' input prompts, SAM automatically segments everything in an image. To leverage SAM through prompting, either points or bounding boxes are required to help SAM in locating potential target locations. In our scenario, point prompts are provided by the acoustic ranging sensors.

The transmission and reception angles of the ultrasonic ranging sensors are typically restricted to a certain acute angle, which is correlated to the transducer structure and the transmission frequency and should be adjusted or even tailored to suit experimental requirements. As long as the target is within the reception field, the following procedure applies.

Align the world coordinate frame with the camera coordinate frame. Set an upper limit $r_\text{max}^\text{th}$ and a lower limit $r_\text{min}^\text{th}$ for the ranging data. Monitor all the ranging sensor outputs in real time. When one or more of the range measurements are within the distance threshold, project the 3-D target position onto the 2-D pixel coordinate frame. Define the coordinates of a detected point $\textbf{\textit{P}}$ on the target in the camera coordinate frame as $[X_\text{c}, Y_\text{c}, Z_\text{c}]^\text{T}$. $X_\text{c}$ and $Y_\text{c}$ are typically constant according to the structural design of the sensing module and $Z_\text{c}$ is acquired by the ranging sensor(s). The 2-D coordinates $[u, v]^\text{T}$ of point $\textbf{\textit{P}}$ in the pixel coordinate frame are obtained per Section~\ref{imagerectify}. The 2-D point(s) is then used as prompt input(s) for the SAM model to begin the segmentation process.

With images and proper prompt(s), SAM is applied to obtain segmentation masks, after which minimum bounding box is achieved for each target. Define the center point pair and four corner point pairs as the five key point pairs set in the left and right images for each target. Obtain coordinates of the five key point pairs of the target, denoted as $\{\left [\textbf{\textit{P}}_\text{L}^\text{1}=(u_\text{L}^\text{1}, v_\text{L}^\text{1}), \textbf{\textit{P}}_\text{R}^\text{1}=(u_\text{R}^\text{1}, v_\text{R}^\text{1})\right ],$ $..., \left [\textbf{\textit{P}}_\text{L}^\text{5}=(u_\text{L}^\text{5}, v_\text{L}^\text{5}), \textbf{\textit{P}}_\text{R}^\text{5}=(u_\text{R}^\text{5}, v_\text{R}^\text{5})\right ]\}$. Calculate vertical disparities of key point pairs, denoted as $\{d_v^\text{1}, ..., d_v^\text{5}\}$. Check whether or not the EMC is satisfied for each pair per Eq.~(\ref{EMC}). If EMC is satisfied for all five key point pairs, the segmentation masks in paired images are matched successfully, indicating the same target.  Distance $Z_\text{i}$ where $i=1,...,5$, of each point is calculated by Eq.~(\ref{calcdistance}). Averaged over the five distances, the estimate of the distance of the target is obtained. Otherwise, if any of the five vertical disparities fails EMC, the segmentation masks of the target in current frame are considered ineffective.

To validate the proposed ranging sensor prompt method, we conduct comparative experiments of one-shot prompt locating. The experiment proceeds by manually labeling desired targets in reference images, encoding  reference and test images using SAM's encoder, dividing encoded images into patches, and calculating patch similarities and probability distributions.

\subsection{Multi-modal Target Localization}
\label{mmtrack}
With pre-processed sensory data, both optical and acoustic sensing modalities are used in underwater target localization. Two types of filtering are designed to assimilate the multi-modal sensor measurements including the weighted averaging filter and the extended Kalman filter (EKF) for target ranging and target motion state estimation, respectively. The filters are selected considering a balance between localization performance and the real-time computational cost.

\subsubsection{Target ranging with the weighted averaging filter}
\label{sec:weightavg}
\par A light intensity sensor is used to monitor the light intensity for further investigation of segmentation and localization success rate under different light intensities. A factor $\alpha\in[0,1]$ that balances between the measurement from optical sensor and from acoustic sensors is designed
\begin{equation}
Z_\text{f} = \alpha*Z_\text{b} + (1-\alpha)*Z_\text{r},
\label{weight_avg}
\end{equation}
where $Z_\text{b}, Z_\text{r}$ and $Z_\text{f}$ represent the distance estimates by the binocular camera alone, the ultrasonic ranging sensors alone, and the multi-modal sensor fusion.

Define the ground truth distance and the estimated distance as $Z_\text{gt}$ and $Z_\text{m}$ respectively. The estimation percentage error is then defined as $e=\left ( \left | Z_\text{m}-Z_\text{gt} \right |/Z_\text{gt}\right )\times100\%$. Parameter $\alpha$ is designed and calculated following an intuitive formula utilizing the averaged estimation percentage error of both sensor modalities. The average distance estimation percentage error of the ranging sensor $\bar{e_\text{r}}$ is obtained from the datasheet. Under a certain illumination intensity, to estimate the distance estimation percentage error of the binocular camera denoted as $\bar{e_\text{b}}$, we take $M$ frames with $N$ targets each frame and calculate $\bar{e_\text{b}}$ by
\begin{equation}
\label{eq:e_b}
    \bar{e_\text{b}} = \frac{1}{M\times N}{\textstyle \sum_{1}^{M}} {\textstyle \sum_{1}^{N}}e_\text{b}^{m,n},
\end{equation}
where $e_\text{b}^{m,n}$ represents the distance estimation percentage error of the $n^{\text{th}}$ target in the $m^{\text{th}}$ frame.
 
The weight $\alpha$ is then calculated as
\begin{equation}
\label{calalpha}
    \alpha = \frac{\bar{e_\text{r}}}{\bar{e_\text{b}}+{\bar{e_\text{r}}}}.
\end{equation}

Furthermore, if target segmentation in frame $t$ fails, the current distance value from binocular camera $Z_\text{b}^t$ is then extrapolated from the previous distance estimates $Z_\text{b}^{t-1},...,Z_\text{b}^{t-n}$. The same extrapolation method is applicable to ultrasonic ranging data as well.

\subsubsection{Target motion state estimation using EKF}
\label{sec:EKF}
This paper establishes the estimation model, including the dynamic system state equation and the observation equation.
We define the estimation state vector $\textbf{\textit{x}}$ and input vector $\textbf{\textit{u}}$ as follows

\begin{equation}
\textbf{\textit{x}}=\begin{bmatrix}
p_x&p_y&p_z&v_x&v_y&v_z
\end{bmatrix}^\text{T},
\end{equation}
\begin{equation}
\textbf{\textit{u}}=\begin{bmatrix}
a_x&a_y&a_z
\end{bmatrix}^\text{T},
\end{equation}
where the first three elements represent the position states, and the last three elements represent the velocity states in the $x$, $y$, and $z$ directions, respectively.

The state equation is given in a compact form as
\begin{equation}
\textbf{\textit{x}}_{\textit{k}}=\textbf{\textit{A}}\textbf{\textit{x}}_{\textit{k}\text{-1}}+\textbf{\textit{B}}\textbf{\textit{u}}_{\textit{k}\text{-1}}+\textbf{\textit{w}}_{\textit{k}}.
\end{equation}
where $\textbf{\textit{w}}$ represents the process noise, and the probability distribution $p(\textbf{\textit{w}})\sim N(\textbf{{\textit{0}}},\textbf{{\textit{Q}}})$, with $\textbf{\textit{Q}}$ being the process noise covariance matrix.

We define the system matrix
$    \textbf{\textit{A}}=\begin{bsmallmatrix}
        \textbf{\textit{A}}_\text{11} &
        \textbf{\textit{A}}_\text{12} \\
        \textbf{\textit{A}}_\text{21} &
        \textbf{\textit{A}}_\text{22}
    \end{bsmallmatrix}$
and the input matrix
$    \textbf{\textit{B}}=\begin{bsmallmatrix}
        \textbf{\textit{B}}_\text{1} \\
        \textbf{\textit{B}}_\text{2}
    \end{bsmallmatrix}$
, where
$\textbf{\textit{A}}_\text{11}$, $\textbf{\textit{A}}_\text{12}$, $\textbf{\textit{A}}_\text{21}$, $\textbf{\textit{A}}_\text{22}$, $\textbf{\textit{B}}_\text{1}$ and $\textbf{\textit{B}}_\text{2}$ are all 3-by-3 block matrices. 
Define position vector $\textbf{\textit{s}}_\text{p}=[p_x, p_y, p_z]^\text{T}$, velocity vector $\textbf{\textit{s}}_\text{v}=[v_x, v_y, v_z]^\text{T}$ and acceleration vector $\textbf{\textit{s}}_\text{a}=[a_x, a_y, a_z]^\text{T}$. Denote the position state vector, velocity state vector and acceleration state vector at time $k$ as $\textbf{\textit{s}}_\text{p}^k$, $\textbf{\textit{s}}_\text{v}^k$ and $\textbf{\textit{s}}_\text{a}^k$, respectively. 
Considering the focus of this paper is to investigate the feasibility and evaluate the performance of the multi-modal close-range underwater target localization using a large vision model, we select and implement a representative motion in which the target travels at a constant speed along the $z$-axis.
In a small sampling interval $\Delta t$, we have
\begin{equation}
\label{calp}
    \textbf{\textit{s}}_\text{p}^k = \textbf{\textit{s}}_\text{p}^{k\text{-1}}+\Delta t \textbf{\textit{s}}_\text{v}^{k\text{-1}}+ \frac{1}{2}(\Delta t)^2\textbf{\textit{s}}_\text{a}^{k\text{-1}}
\end{equation}
\begin{equation}
\label{calv}
    \textbf{\textit{s}}_\text{v}^k = \textbf{\textit{s}}_\text{v}^{k\text{-1}}+\Delta t \textbf{\textit{s}}_\text{a}^{k\text{-1}}
\end{equation}

Based on Eqs. (\mbox{\ref{calp}}) and (\mbox{\ref{calv}}), $\textbf{\textit{A}}_\text{11}=\textbf{\textit{A}}_\text{22}=\textbf{\textit{I}}_\text{3}$, $\textbf{\textit{A}}_\text{12}=\textbf{\textit{B}}_\text{2}=\Delta t\textbf{\textit{I}}_\text{3}$, 
$\textbf{\textit{A}}_\text{21}=\textbf{\textit{0}}_\text{3}$ and 
$\textbf{\textit{B}}_\text{1}=\frac{1}{2}(\Delta t)^2\textbf{\textit{I}}_\text{3}$, where $\textbf{\textit{I}}$ and $\textbf{\textit{0}}$ represent identity matrix and zero matrix, respectively.

Define the measurement vector as follows
\begin{equation}
    \textbf{\textit{z}}=\begin{bmatrix}
        u&v&d_u&d_r
    \end{bmatrix}^\text{T},
\end{equation}
where $[u,v]^\text{T}$ is the 2-D coordinate in the image coordinate system, $d_u$ is the disparity value calculated from the binocular camera and $d_r$ is the distance value from the ranging sensor.

The measurement equation is in the form of
\begin{equation}
\textbf{\textit{z}}_{\textit{k}}=\textbf{\textit{h}}(\textbf{\textit{x}}_{\textit{k}})+\textbf{\textit{v}}_{\textit{k}},
\end{equation}
where nonlinear function vector $\textbf{\textit{h}}$ is described as
\begin{equation}
\textbf{\textit{h}}(\textbf{\textit{x}}_{\textit{k}})=\begin{bmatrix}
\dfrac{f_up_x}{p_z}+c_u, \dfrac{f_vp_y}{p_z}+c_v, \dfrac{bf_u}{p_z}, p_z
\end{bmatrix}^\text{T},
\end{equation}
$\textbf{\textit{v}}$ is the measurement noise and the distribution $p(\textbf{\textit{v}})\sim N(\textbf{{\textit{0}}},\textbf{{\textit{R}}})$, where $\textbf{\textit{R}}$ is the measurement noise covariance matrix.

The Jacobian matrix $\textbf{\textit{H}}$ is the partial derivatives of $\textbf{\textit{h}}$ with respect to $\textbf{\textit{x}}$, i.e.,
\begin{equation}
\renewcommand\arraystretch{1.8}
     \textbf{\textit{H}}=\begin{bmatrix}
        \dfrac{f_u}{p_z}&0&-\dfrac{f_up_x}{p_z^2}&0&0&0 \\
        0&\dfrac{f_v}{p_z}&-\dfrac{f_vp_y}{p_z^2}&0&0&0 \\
        0&0&-\dfrac{bf_u}{p_z^2}&0&0&0 \\
        0&0&1&0&0&0
    \end{bmatrix},
\end{equation}

The process noise $\textbf{\textit{w}}$ in our experiment mainly comes from the vibration of the guiding system travelling through water and is modeled as follows. We consider the vibration in the system results in a non-zero  acceleration of the sensing module which follows a Gaussian process with zero mean and a constant variation. The variances of the position and the velocity as well as the covariance between the position and the velocity are then calculated accordingly. Assuming that the external forces are independent along the $x,y,z$ axes, and within a small time interval $\Delta t$ the accelerations $a_x, a_y, a_z$ in the $x,y,z$ axes are all constant, the resulting velocity and position $v_x,v_y,v_z$ and $p_x,p_y,p_z$ follow Gaussian distributions. The variances of $v_x$ and $p_x$ are calculated as follows and $v_y,v_z$ and $p_y,p_z$ compute similarly.
\begin{equation}
\sigma(v_x)=\sigma(a_x\Delta t) = (\Delta t)^2\sigma(a_x),
\end{equation}
\begin{equation}
\sigma(p_x)=\sigma(\frac{1}{2}a_x(\Delta t)^2)=\frac{1}{4}(\Delta t)^4\sigma(a_x).
\end{equation}

The covariance between the position and the velocity along the same axis is then calculated as
\begin{equation}
\sigma(p_x,v_x)=\sqrt{\sigma(p_x)}\sqrt{\sigma(v_x)}=\frac{1}{2}(\Delta t)^3\sigma(a_x),
\end{equation}

The process noise covariance matrix $\textbf{\textit{Q}}$ is given by
\begin{equation}
\renewcommand\arraystretch{1.8}
\textbf{\textit{Q}}=\begin{bmatrix}
    \dfrac{1}{4}(\Delta t)^4\textbf{\textit{Q}}_\text{b} &
    \dfrac{1}{2}(\Delta t)^3\textbf{\textit{Q}}_\text{b} \\
    \dfrac{1}{2}(\Delta t)^3\textbf{\textit{Q}}_\text{b} &
    (\Delta t)^2\textbf{\textit{Q}}_\text{b}
\end{bmatrix},
\end{equation}
where $\textbf{\textit{Q}}_\text{b} = \text{diag}(\sigma(a_x), \sigma(a_y), \sigma(a_z))$ is the 3-by-3 building block diagonal matrix.

\section{Experiments}

This section presents the implementation setup of the physical experiments, the experimental results, and the corresponding analyses.
\subsection{Implementation Setup}
The binocular camera was calibrated using a checkerboard with $9\times 12$ square grids of 2~cm by 2~cm dimension. The calibration was completed in water using the open source computer vision library --- OpenCV. Intrinsic parameters and distortion parameters of the binocular camera are listed in Table \ref{bino-intrinsic-geometry}. The extrinsic parameters of the binocular system are listed in Table \ref{external-para}. The transmission frequency of the ultrasonic ranging sensors is 1~MHz. To ensure target visibility in camera field of view (FOV) and exclusive target detection by ultrasonic ranging sensors, we adopted trial and error and selected a cone angle of 4 degrees in the experiment. To mitigate cross-talk issue among different ultrasonic ranging sensors, we implemented a software-based synchronization method. This approach ensures simultaneous signal transmission through multi-threading on Raspberry Pi, leveraging various Python libraries. Subsequently, we applied an extended Kalman Filter to the measurement data to enhance accuracy and reliability.
\begin{table*}[thpb]
\caption{Intrinsic parameters}
\label{bino-intrinsic-geometry}
\begin{center}
\renewcommand\arraystretch{1.4}
\begin{tabular}{cccccccccc}
\toprule
\multirow{2}*{\textbf{View}} & \multicolumn{4}{c}{Intrinsic Parameters} & \multicolumn{3}{c}{Radial Distortion Parameters} & \multicolumn{2}{c}{Tangential Distortion Parameters} \\
  \cmidrule(lr){2-5}\cmidrule(lr){6-8}\cmidrule(lr){9-10} &
$\textbf{\textit{f}}_\textbf{\textit{x}}$ & $\textbf{\textit{f}}_\textbf{\textit{y}}$ & $\textbf{\textit{c}}_\textbf{\textit{x}}$ & $\textbf{\textit{c}}_\textbf{\textit{y}}$ &
$\textbf{\textit{k}}_\textbf{\textit{1}}$ & $\textbf{\textit{k}}_\textbf{\textit{2}}$ & $\textbf{\textit{k}}_\textbf{\textit{3}}$ &
$\textbf{\textit{p}}_\textbf{\textit{4}}$ & $\textbf{\textit{p}}_\textbf{\textit{5}}$ \\
\midrule
\textbf{Left} & $1.241\times10^{3}$ & $1.187\times10^{3}$ & $6.61\times10^{2}$ & $5.06\times10^{2}$ & $2.92\times10^{-1}$ & $9.98\times10^{-1}$ & $-1.74$ & $3.30\times10^{-3}$ & $-2.64\times10^{-3}$\\
\textbf{Right} & $1.242\times10^{3}$ & $1.184\times10^{3}$ & $6.93\times10^{2}$ & $5.25\times10^{2}$ & $2.92\times10^{-1}$ & $8.97\times10^{-1}$ & $-1.13$ & $1.12\times10^{-3}$ & $-1.56\times10^{-3}$\\
\bottomrule
\end{tabular}
\end{center}
\end{table*}

\begin{table}[thpb]
\caption{Extrinsic Parameters}
\begin{center}
\renewcommand\arraystretch{1.4}
\begin{tabular}{cc}
\toprule
\textbf{Rotation} & $\begin{bmatrix}
9.99\times10^{-1} & 2.08\times10^{-3} & -9.35\times10^{-3}\\
-2.03\times10^{-3} & 9.99\times10^{-1} & 5.62\times10^{-3}\\
9.36\times10^{-3} & -5.60\times10^{-3} & 9.99\times10^{-1}
\end{bmatrix}$\\
\midrule
\textbf{Translation}
& $\begin{bmatrix}
-59.02 & 0.17 & -0.43
\end{bmatrix}$\\
\bottomrule
\end{tabular}
\end{center}
\label{external-para}
\end{table}

\begin{figure}[thpb]
\centering
\subfigure[Scene 1.]{
\label{11@50}
\includegraphics[width=0.30\textwidth]{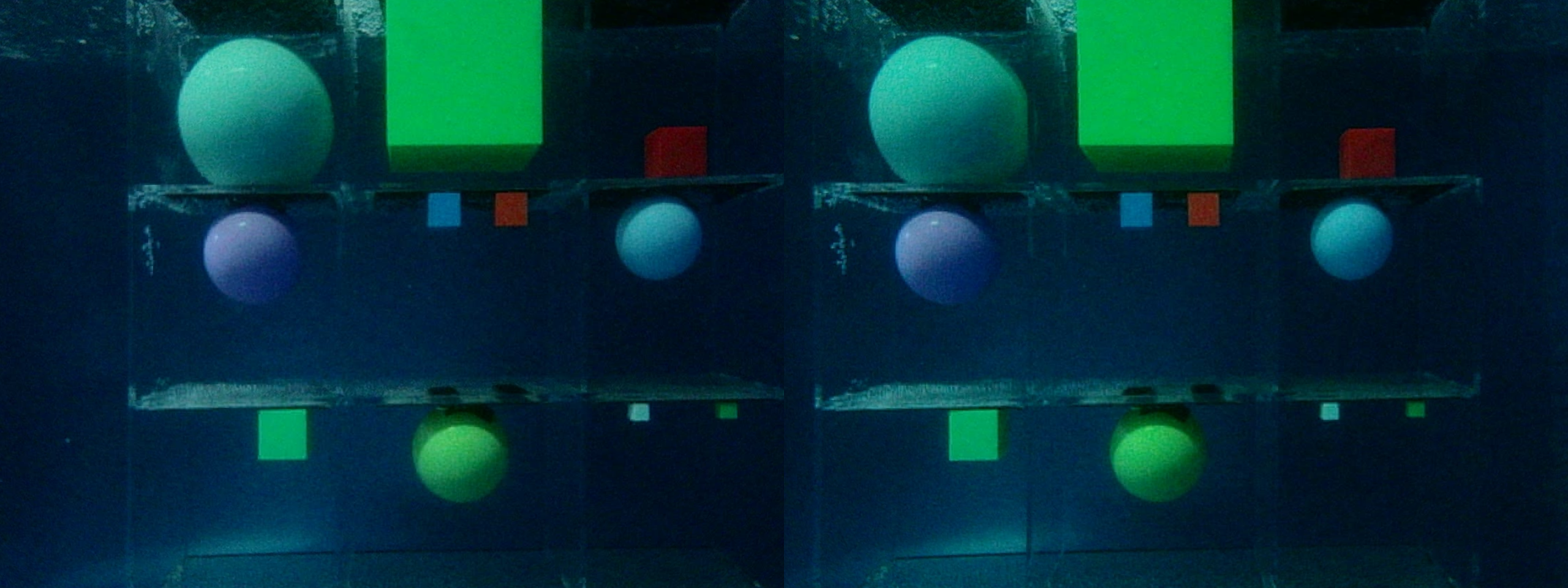}}
\subfigure[Scene 2.]{
\label{10@55}
\includegraphics[width=0.15\textwidth]{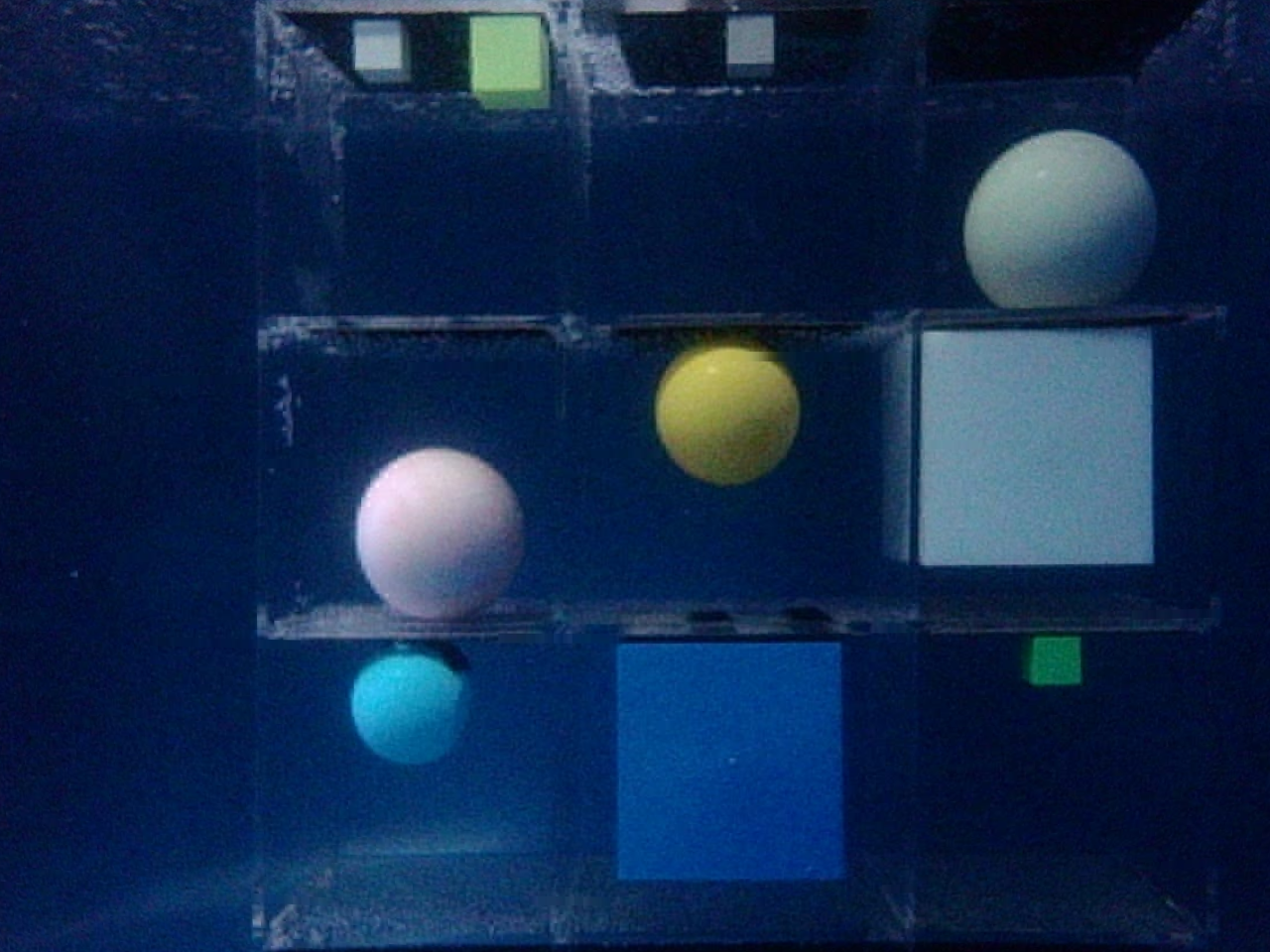}}
\subfigure[Scene 3.]{
\label{9@60}
\includegraphics[width=0.15\textwidth]{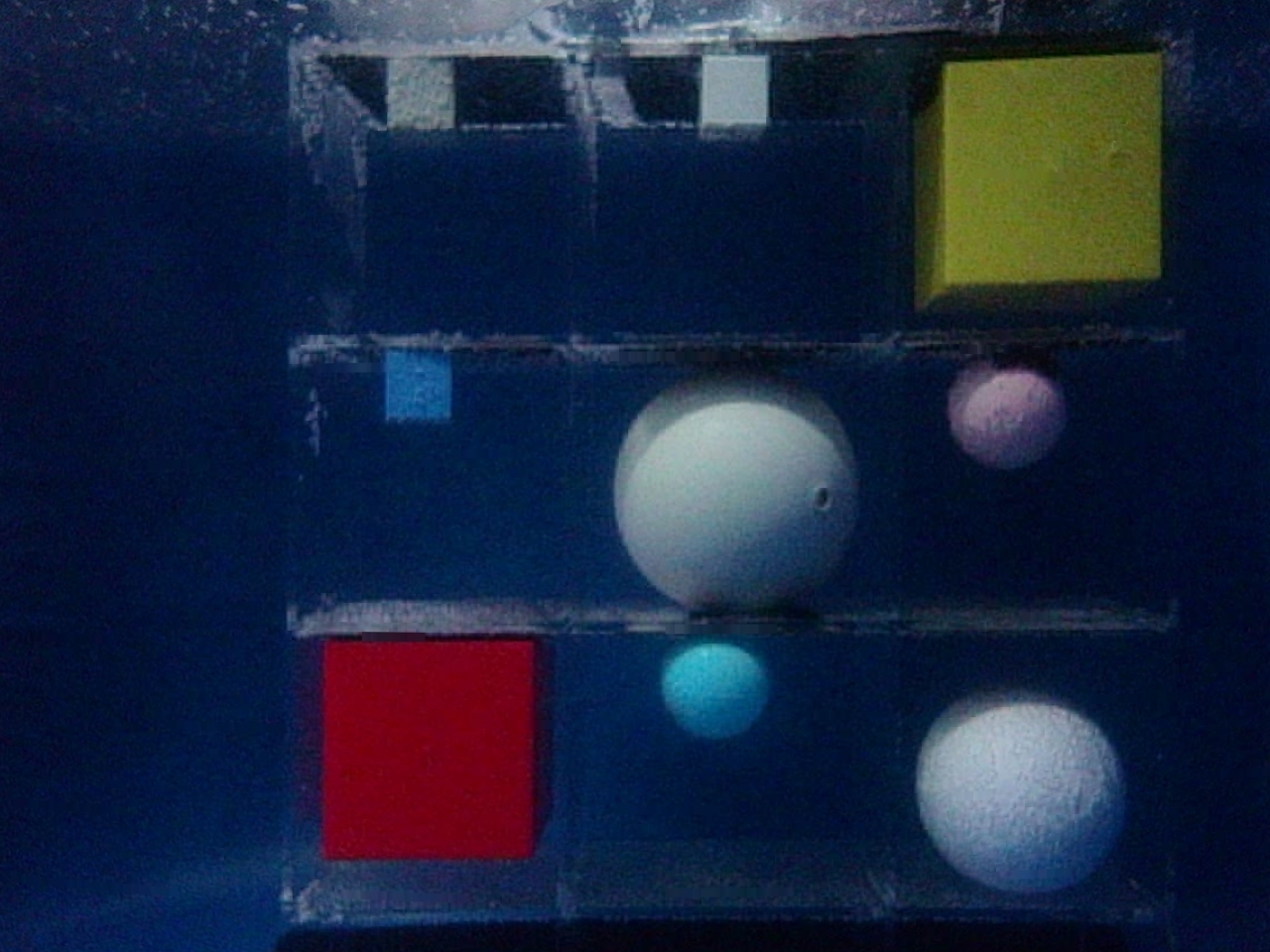}}
\subfigure[Scene 4.]{
\label{dynamic_scene4}
\includegraphics[width=0.15\textwidth]{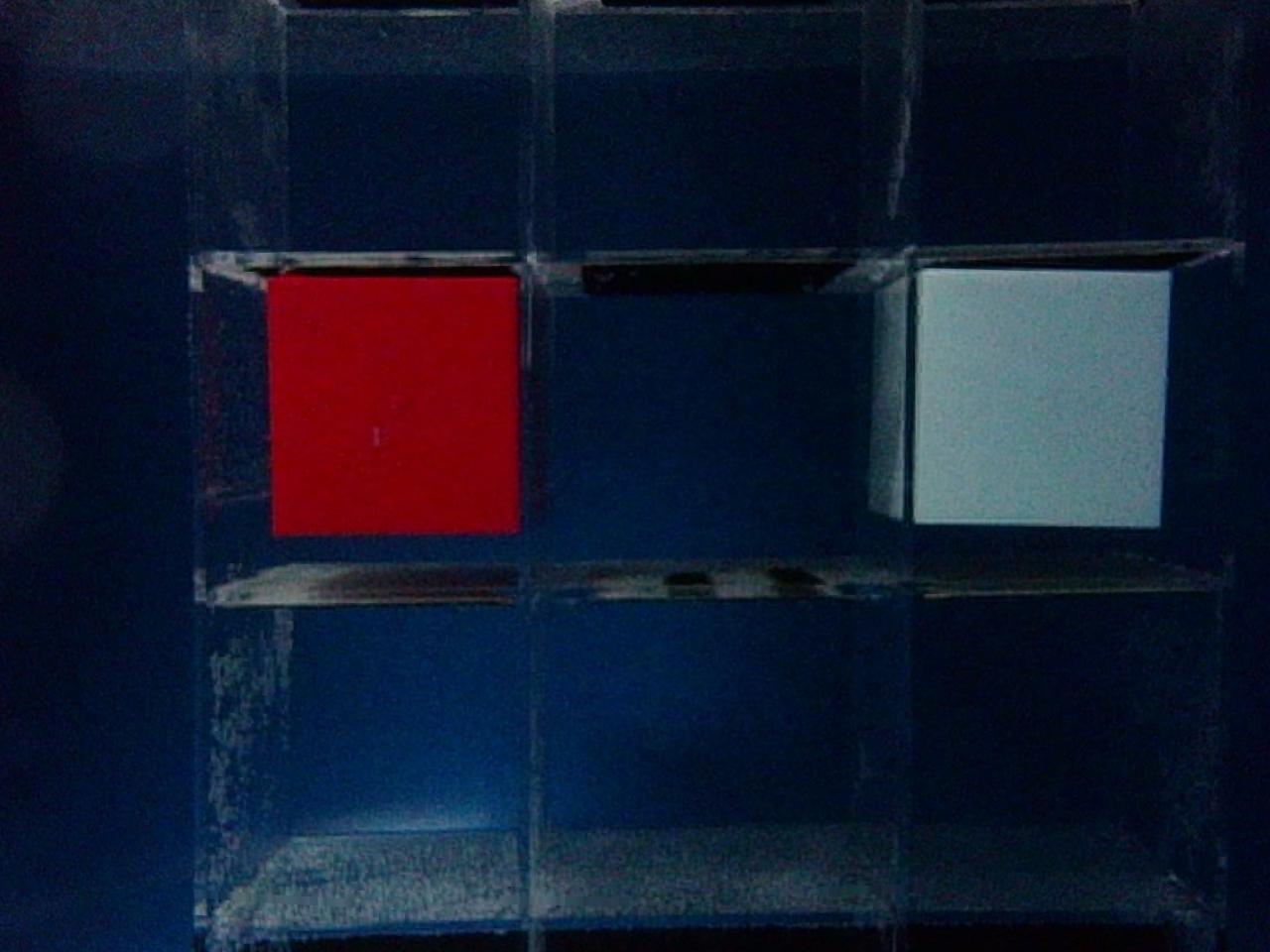}}
\subfigure[Scene 5.]{
\label{dynamic_scene5}
\includegraphics[width=0.15\textwidth]{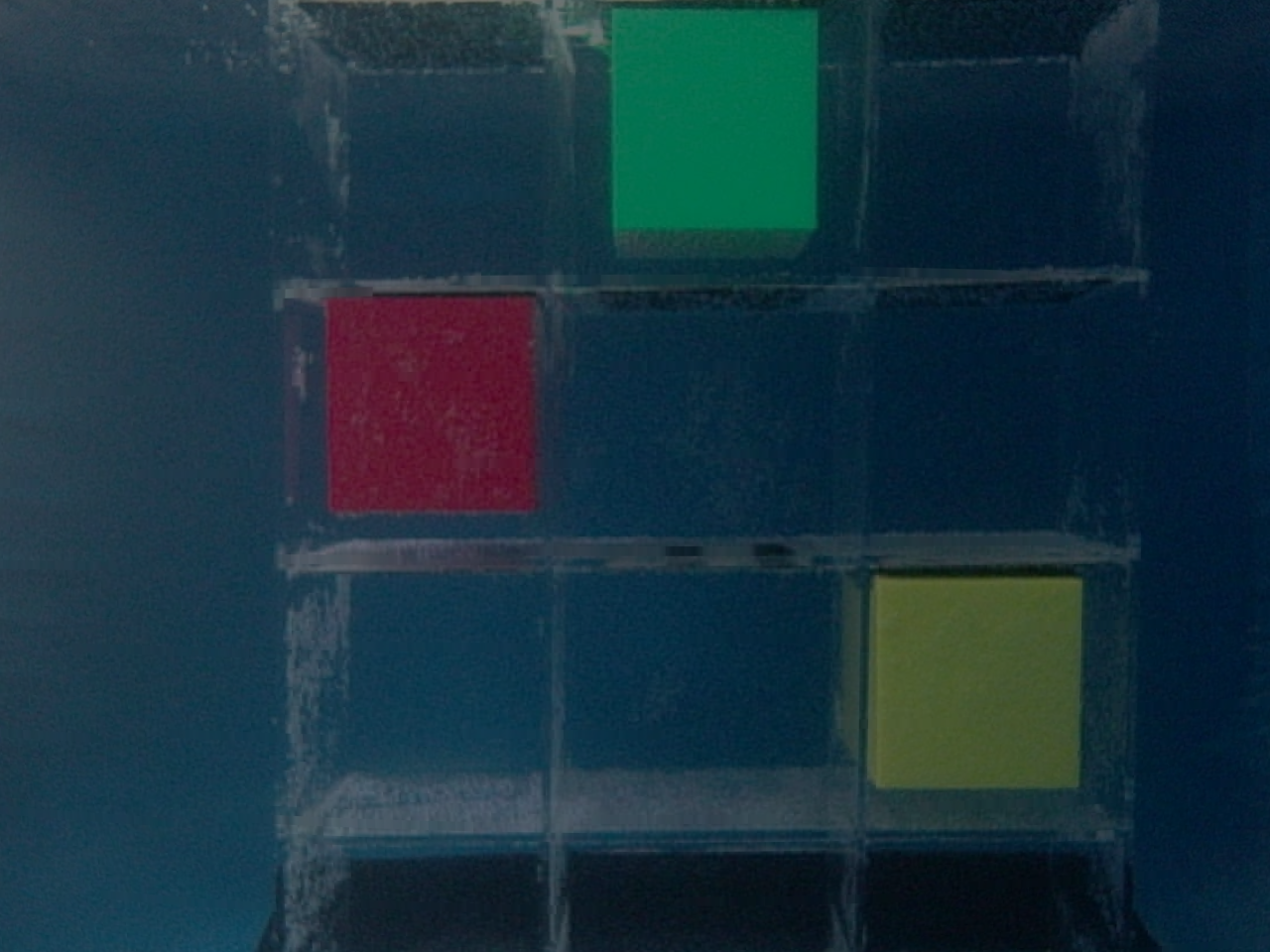}}
\subfigure[Scene 6.]{
\label{marinelife_scene6@50}
\includegraphics[width=0.15\textwidth]{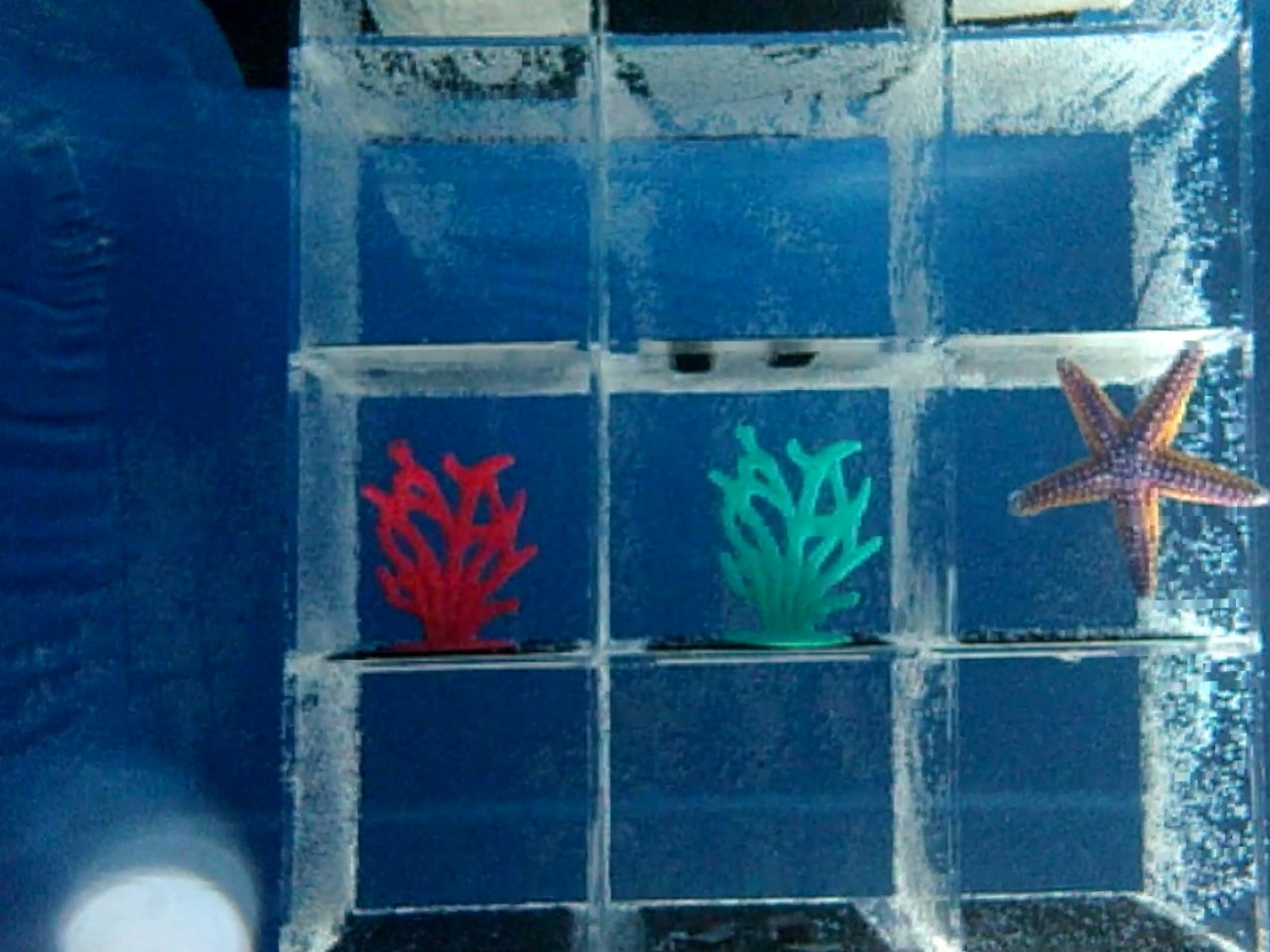}}
\subfigure[Scene 7.]{
\label{marinelife_scene7@55}
\includegraphics[width=0.15\textwidth]{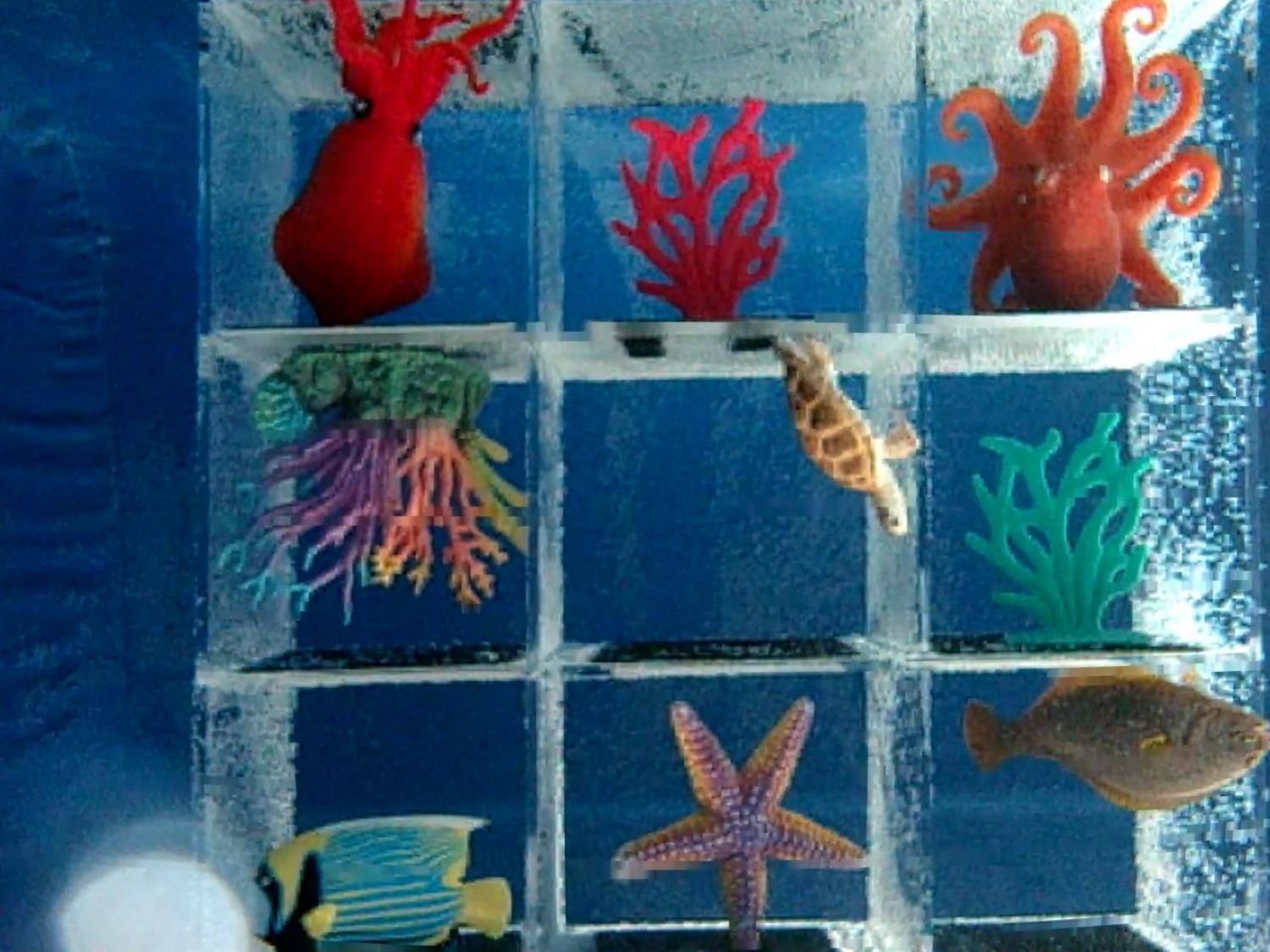}}
\subfigure[Scene 8.]{
\label{marinelife_scene8@60}
\includegraphics[width=0.15\textwidth]{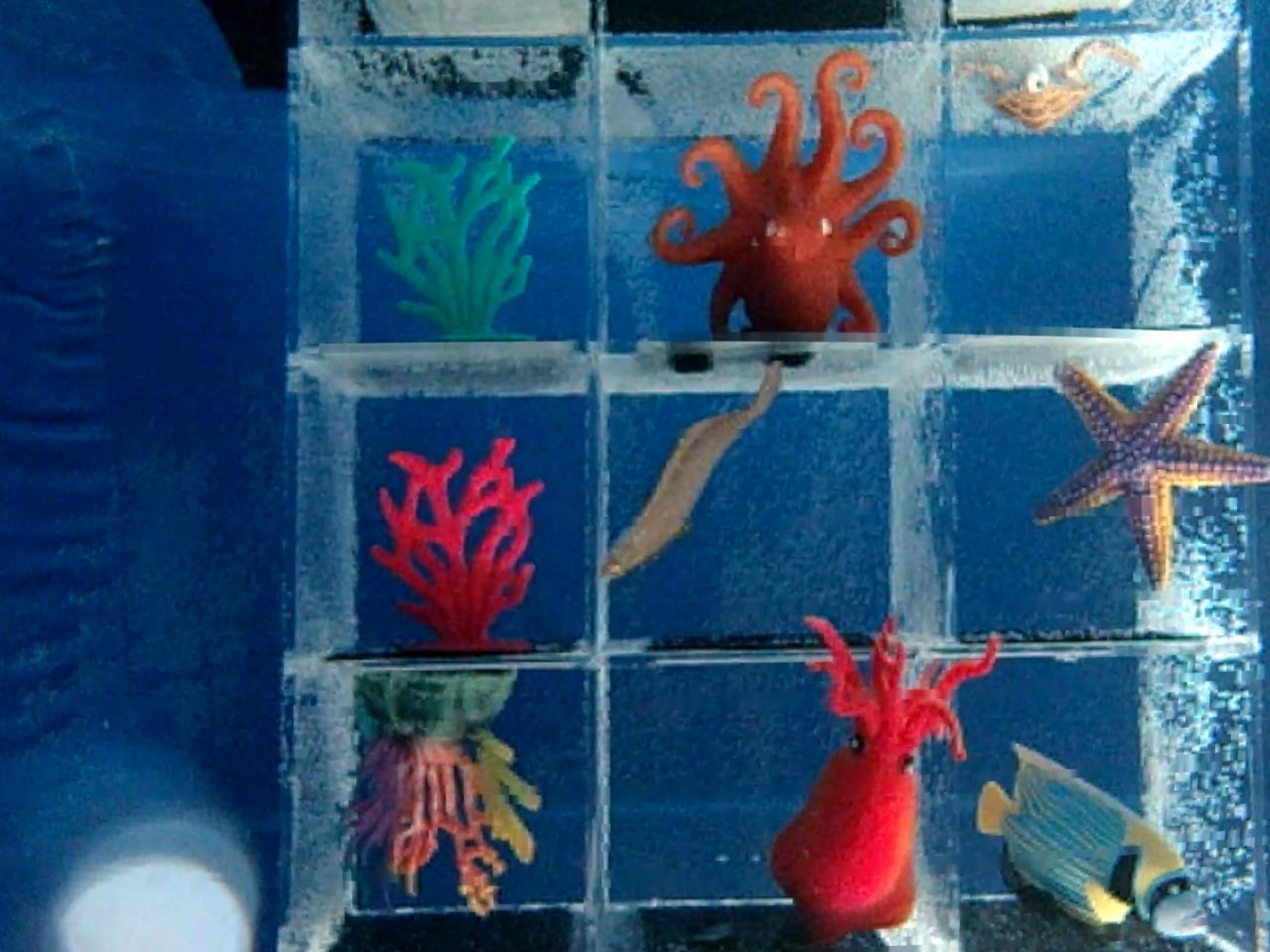}}
\subfigure[Scene 9.]{
\label{oneshot-1}
\includegraphics[width=0.15\textwidth]{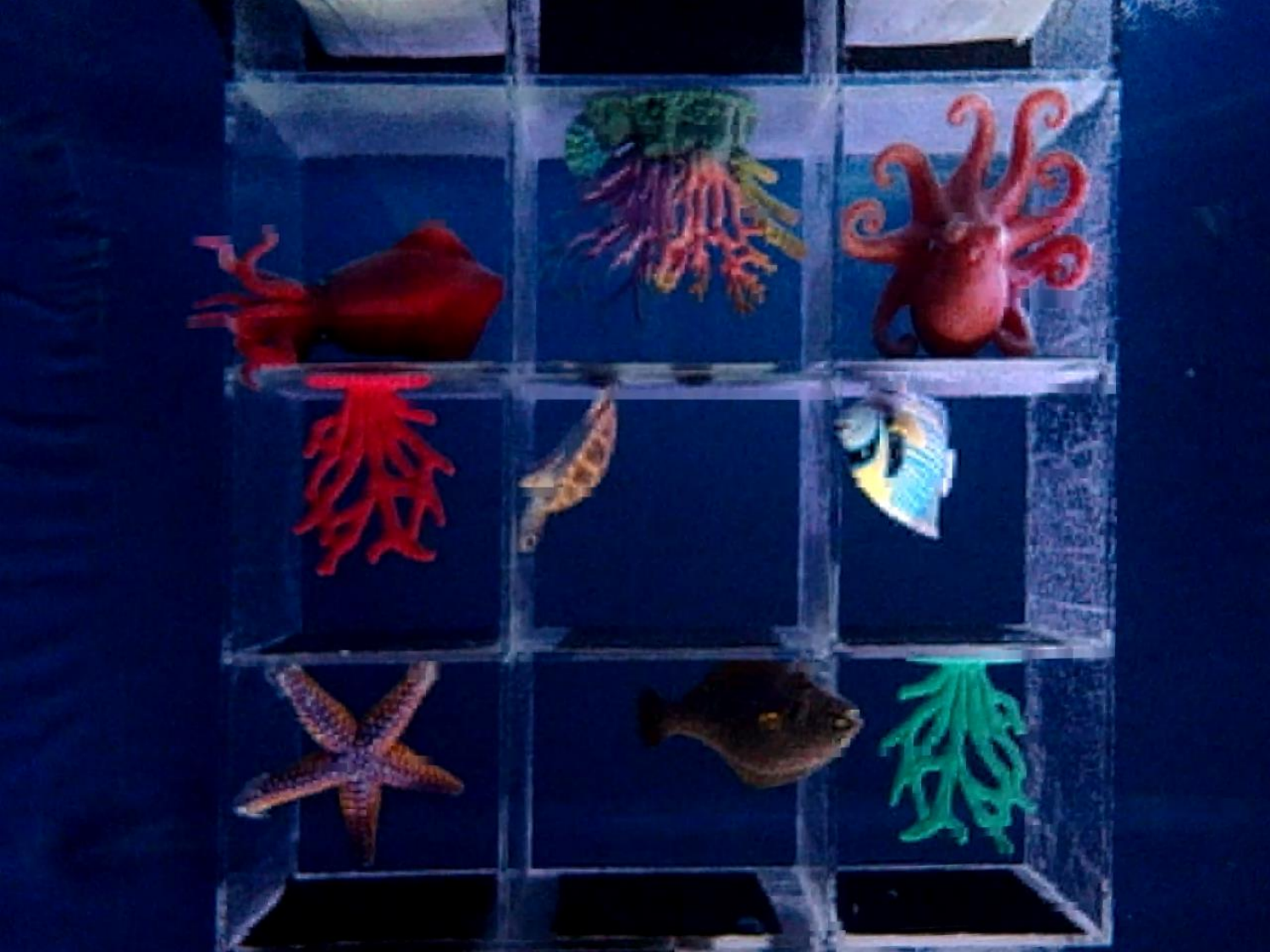}}
\subfigure[Scene 10.]{
\label{oneshot-2}
\includegraphics[width=0.15\textwidth]{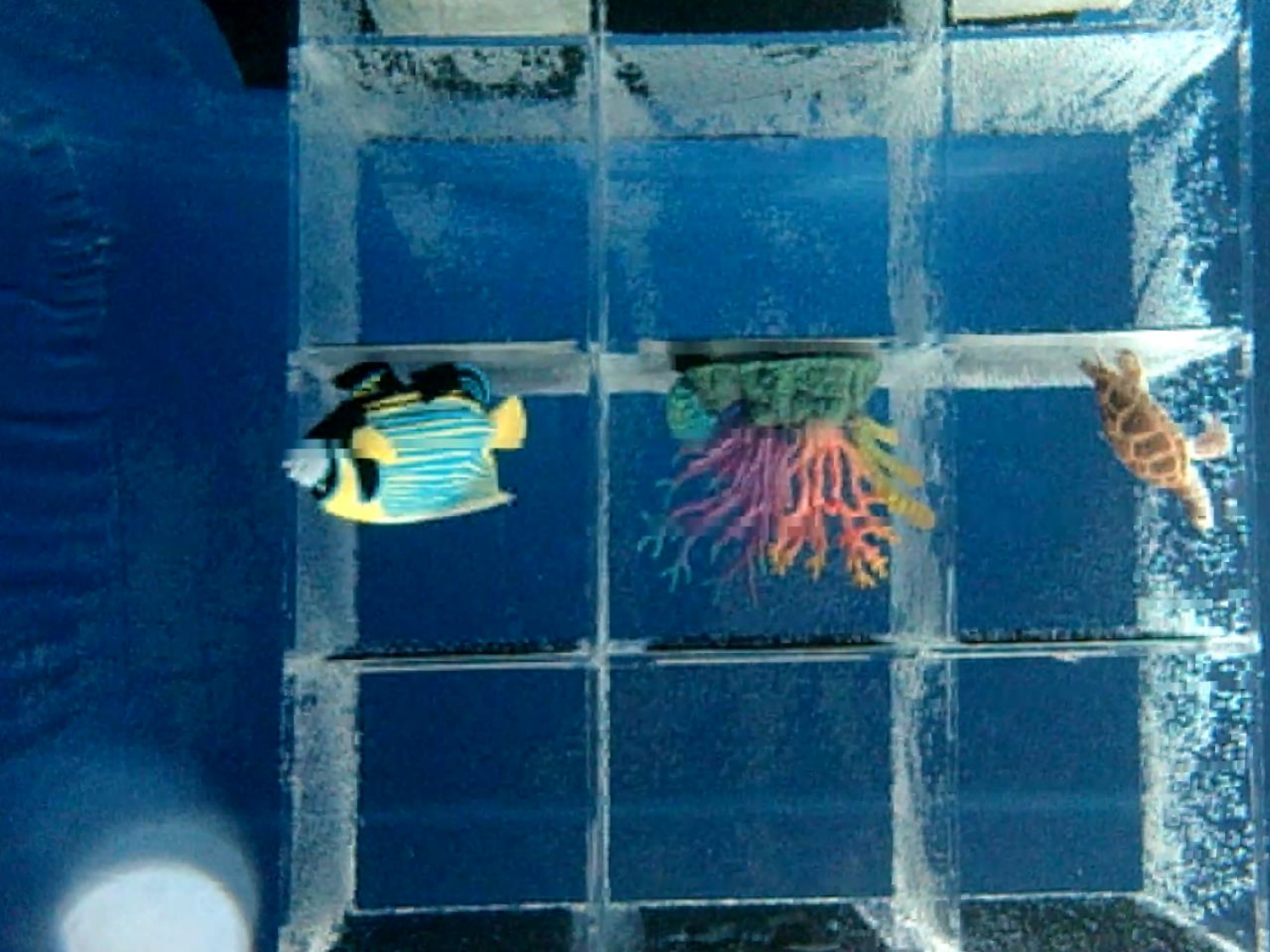}}
\subfigure[Scene 11.]{
\label{oneshot-3}
\includegraphics[width=0.15\textwidth]{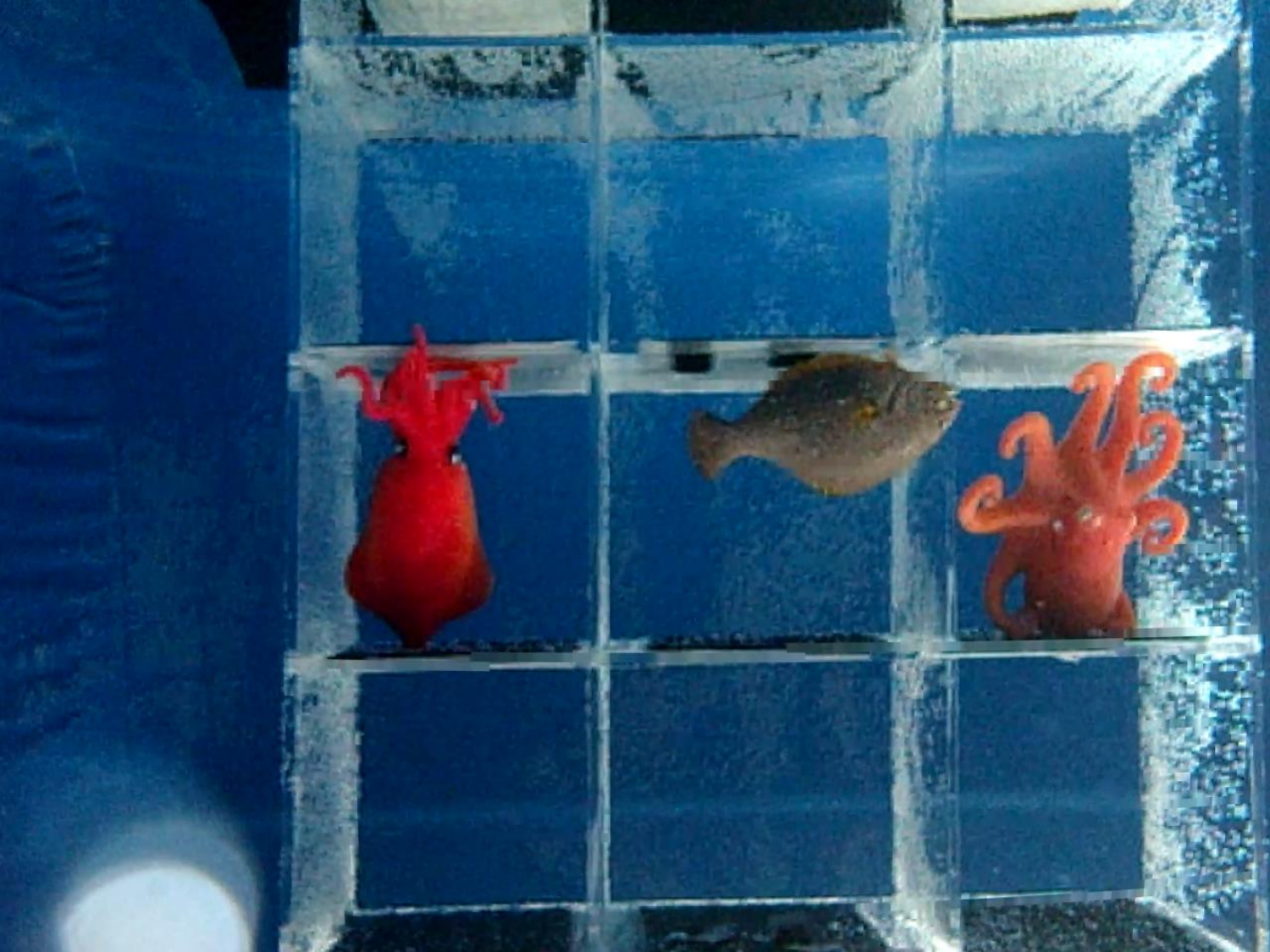}}
\caption{Layout of the 11 test scenes. Scene~1 includes 11 targets placed at a distance of 0.5~m from the sensing module. Scene~2 includes 10 targets placed at a distance of 0.55~m. Scene~3 includes 9 targets at a distance of 0.6~m. Scenes~4 and 5 include 2 and 3 targets of dynamic motion, respectively. Scenes~6-8 incorporate the same set of aquatic life model targets shown in Fig.~\mbox{\ref{marinelife}} but setup at different distances. Scenes~6-8 are placed at a distance of 0.5~m, 0.55~m and 0.6~m, respectively. Scenes~9-11 are used for one-shot prompt locating process. Scene 1 shows the paired left and right view images of the binocular camera while Scenes 2-11 show only the left view images. All targets in Scenes~4-5 are included in Scenes~1-3.}
\label{scenes}
\end{figure}

As shown in Fig. \ref{scenes}, a total of 11 scenes with 30 regular-shaped targets and 9 aquatic life model targets were designed and used in the experiment. All the images were rectified with camera calibration parameters. 

The experiment mainly includes three tasks. In the first task, Scenes~1-3 and 6-8 are used, where the targets are stationary. The segmentation capability of SAM and the influence of the light intensity on the ranging estimation performance are comprehensively studied with extensive experiments. In the second and third tasks, Scenes~4-5 are used, where 1-D ranging and 3-D position \& velocity estimation of dynamic targets are investigated. For dynamic targets, the sensing module moves along the sliding guide at a constant speed of either $1.25\times10^{-2}$~m/s or $5\times10^{-3}$~m/s. The weighted averaging and EKF filters are used in Tasks 2 and 3, respectively.

\subsection{Experimental Results}
\subsubsection{Task 1}
\begin{figure}[thpb]
\centering
\subfigure[4~lux illumination]{
\label{4-lux}
\includegraphics[width=0.15\textwidth]{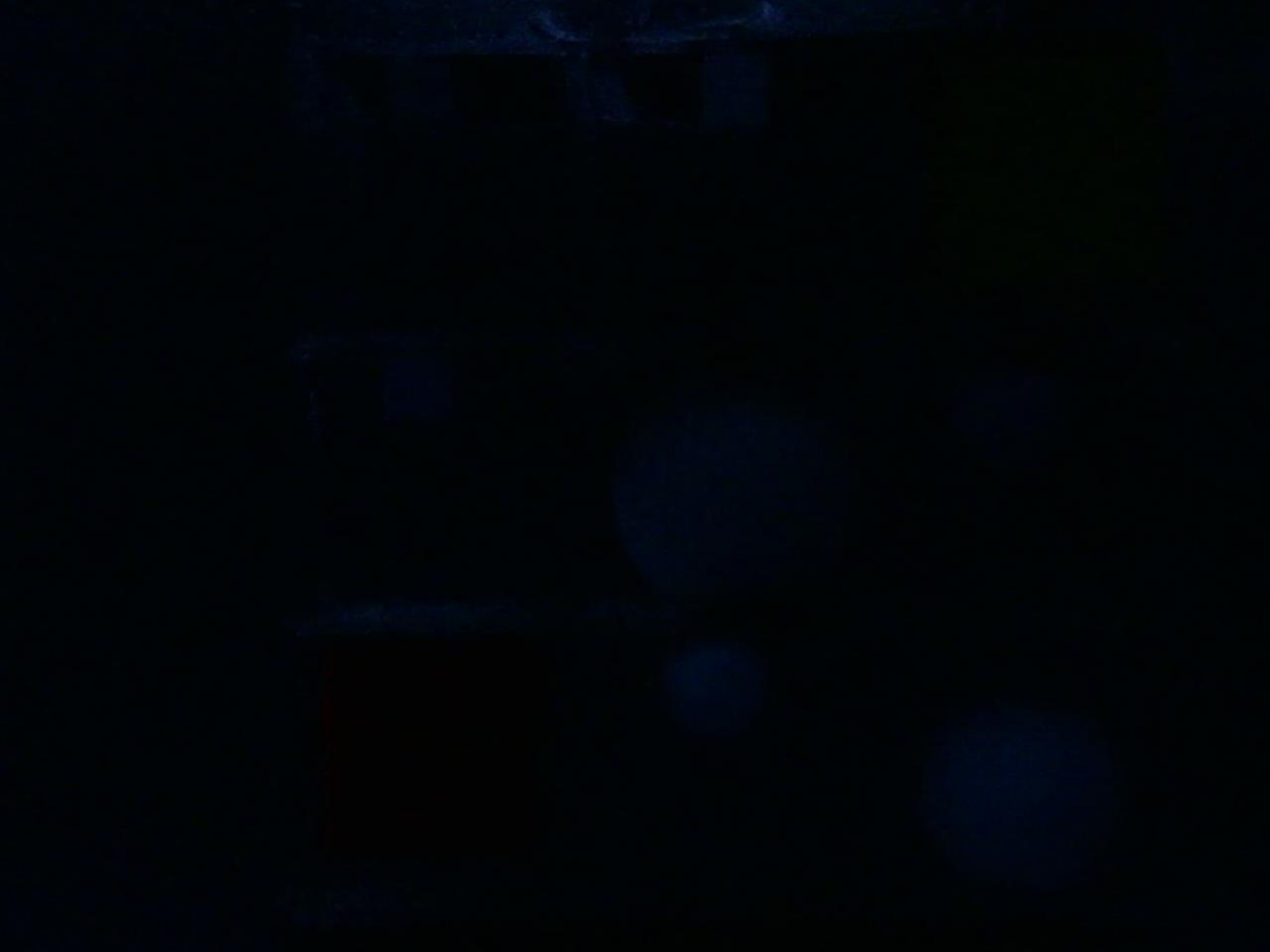}}
\subfigure[6~lux illumination]{
\label{6-lux}
\includegraphics[width=0.15\textwidth]{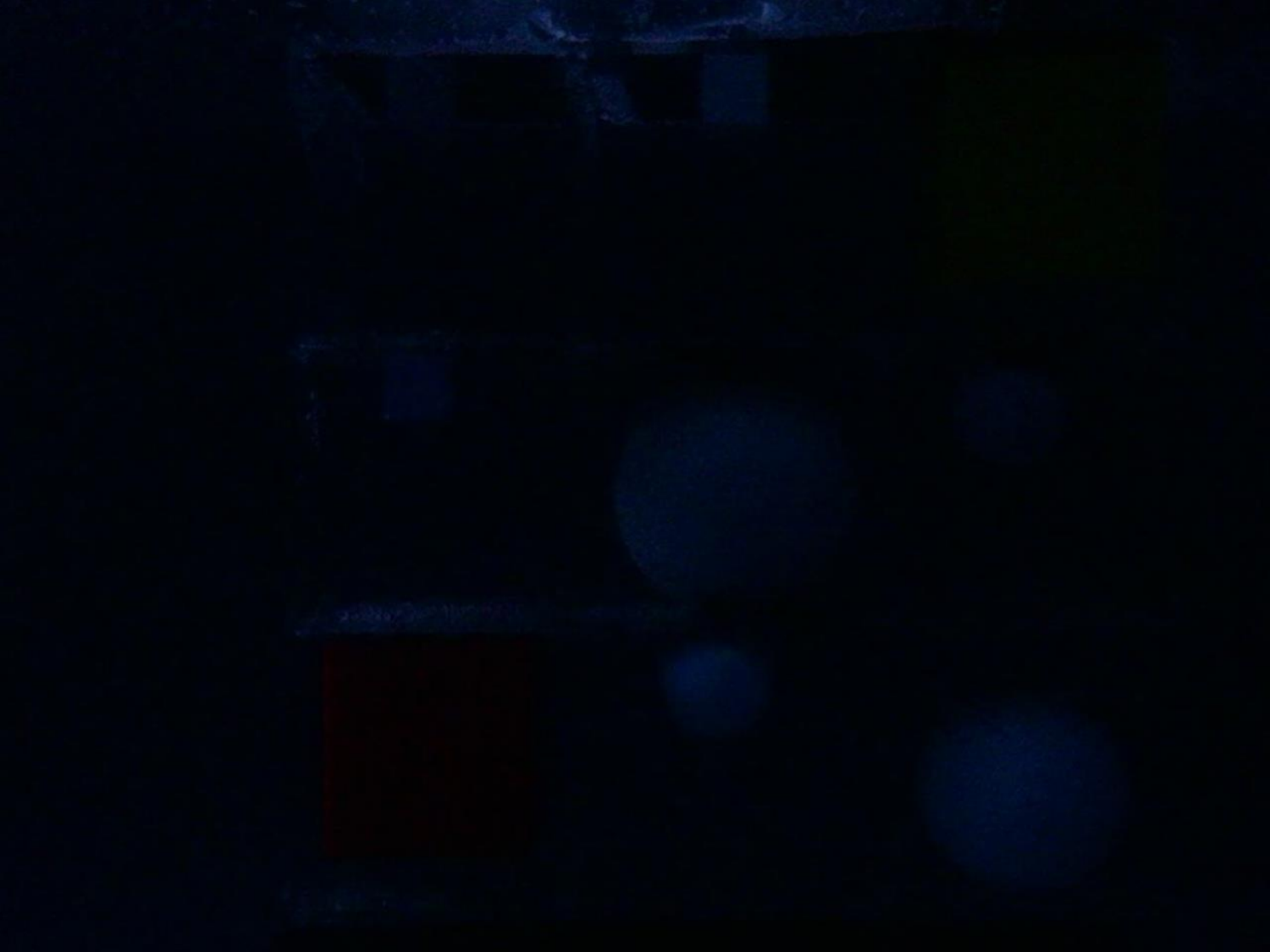}}
\subfigure[8~lux illumination]{
\label{8-lux}
\includegraphics[width=0.15\textwidth]{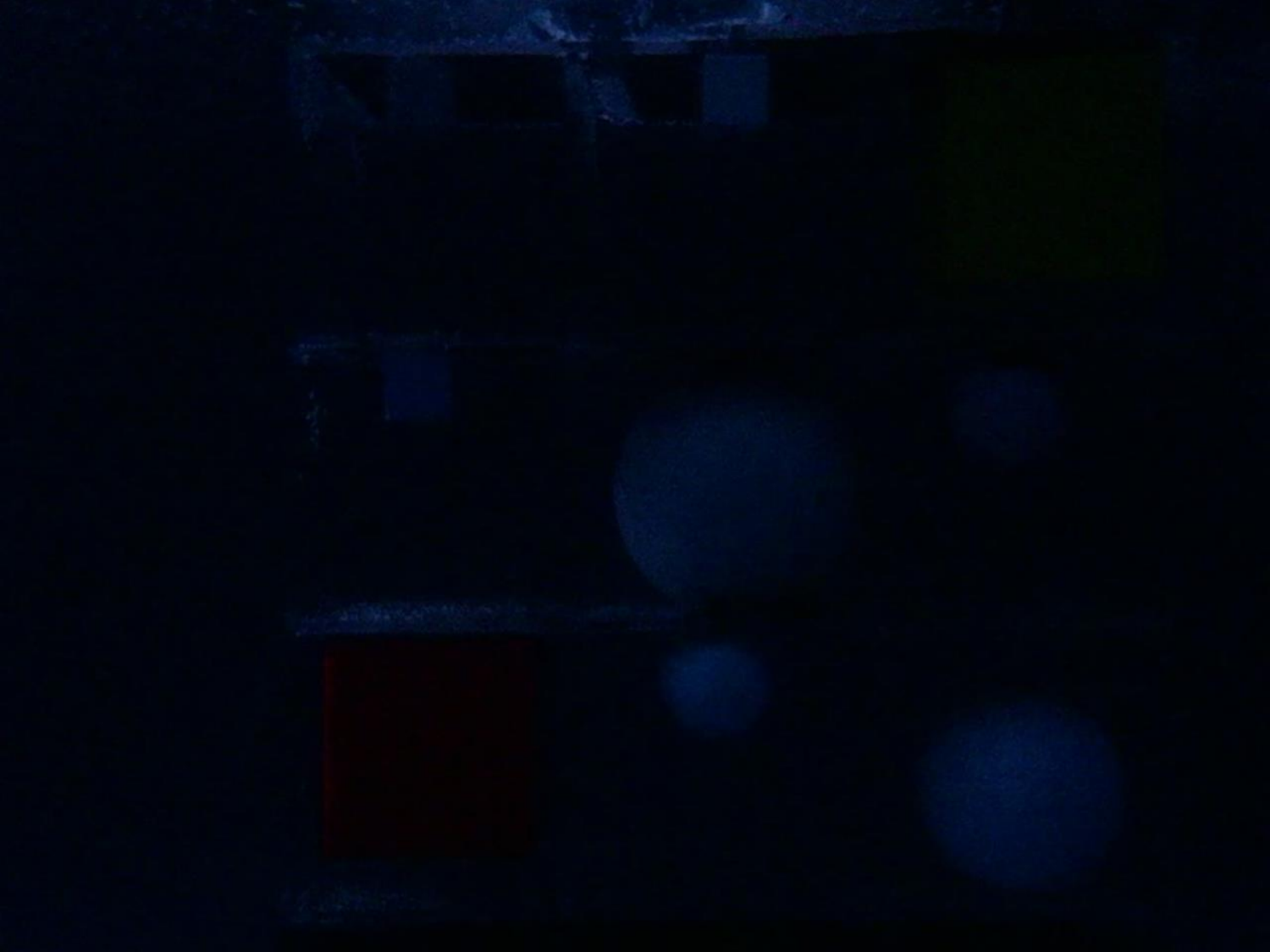}}
\subfigure[10~lux illumination]{
\label{10-lux}
\includegraphics[width=0.15\textwidth]{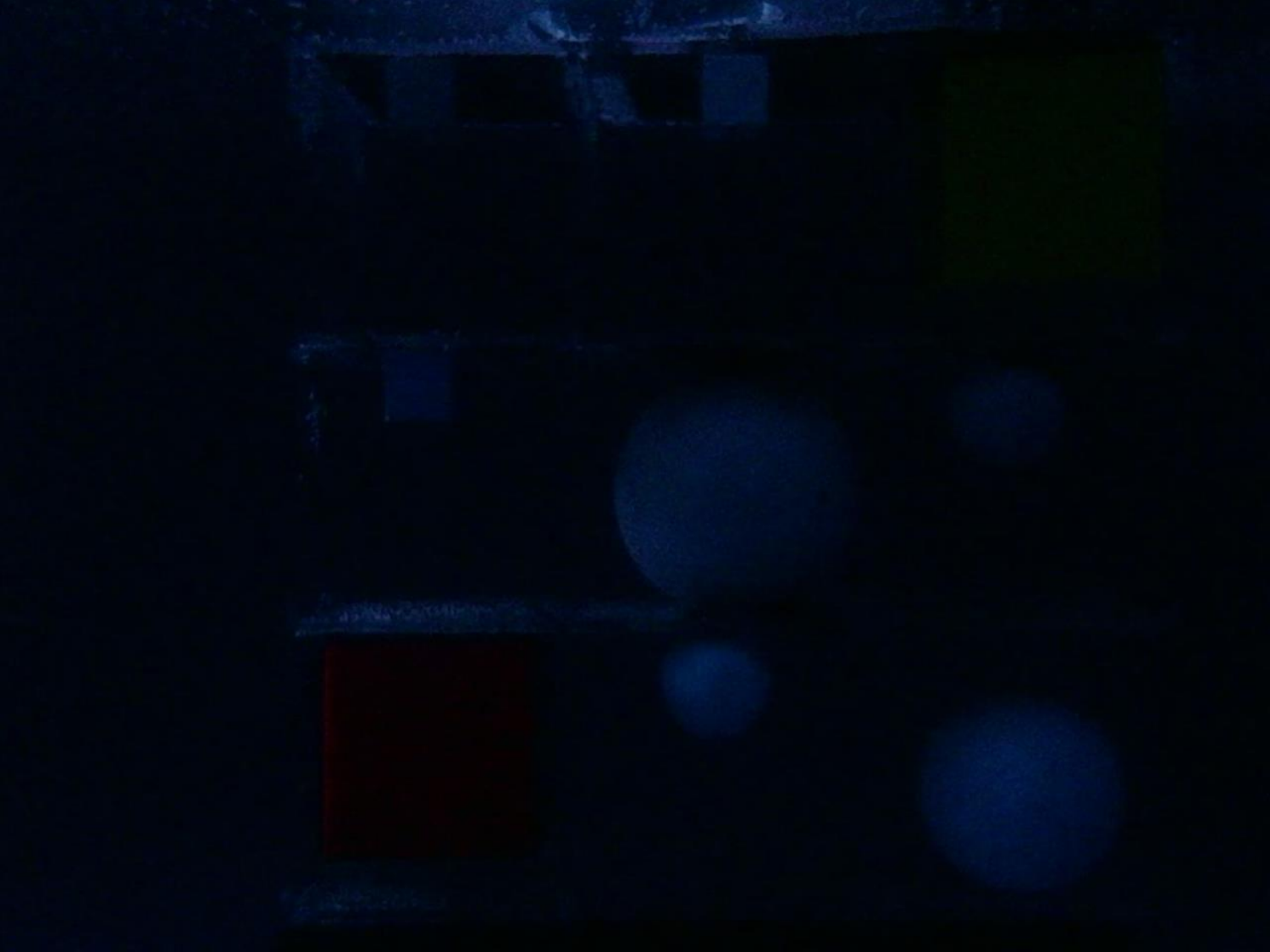}}
\subfigure[12~lux illumination]{
\label{12-lux}
\includegraphics[width=0.15\textwidth]{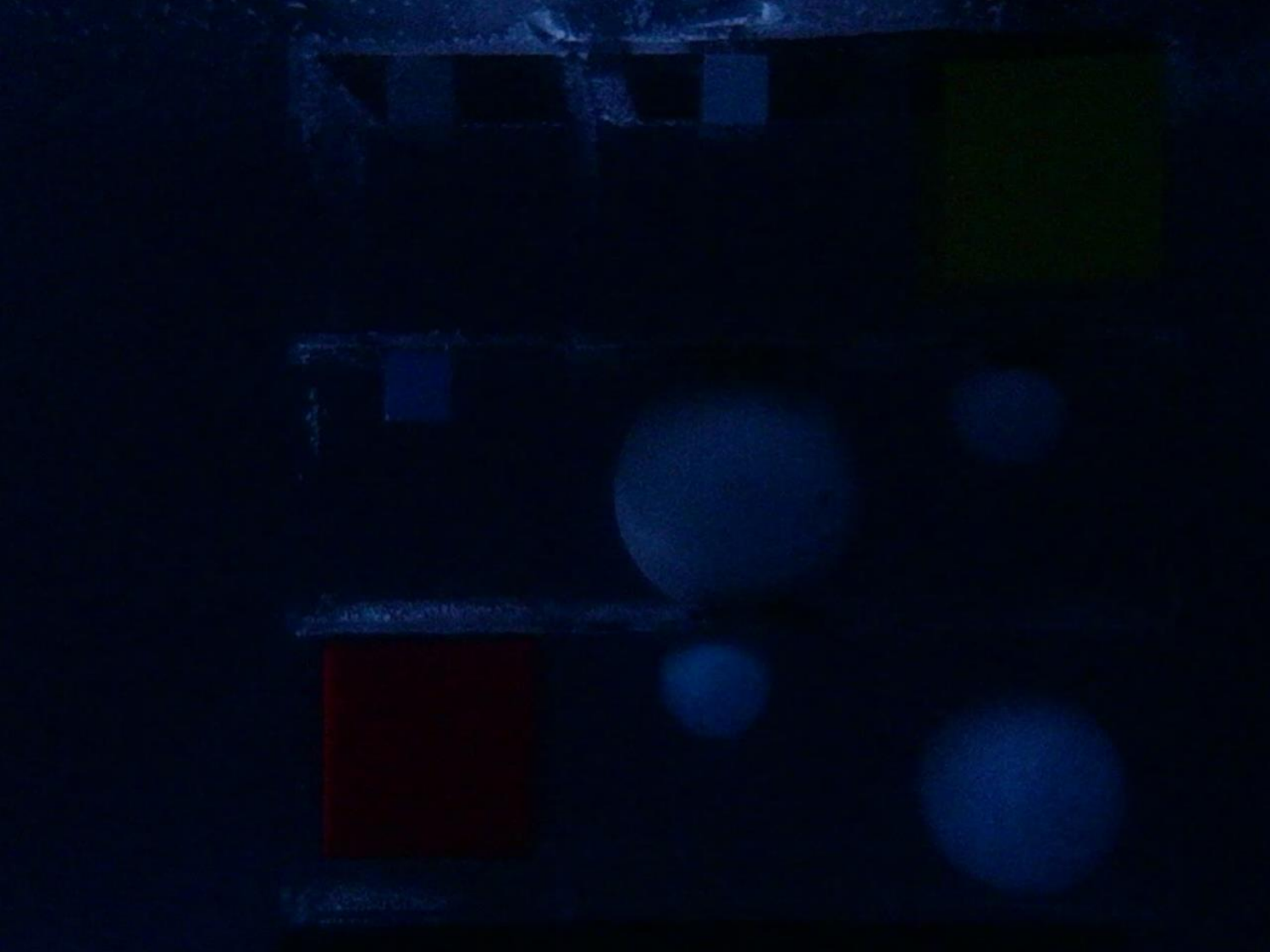}}
\subfigure[25~lux illumination]{
\label{25-lux}
\includegraphics[width=0.15\textwidth]{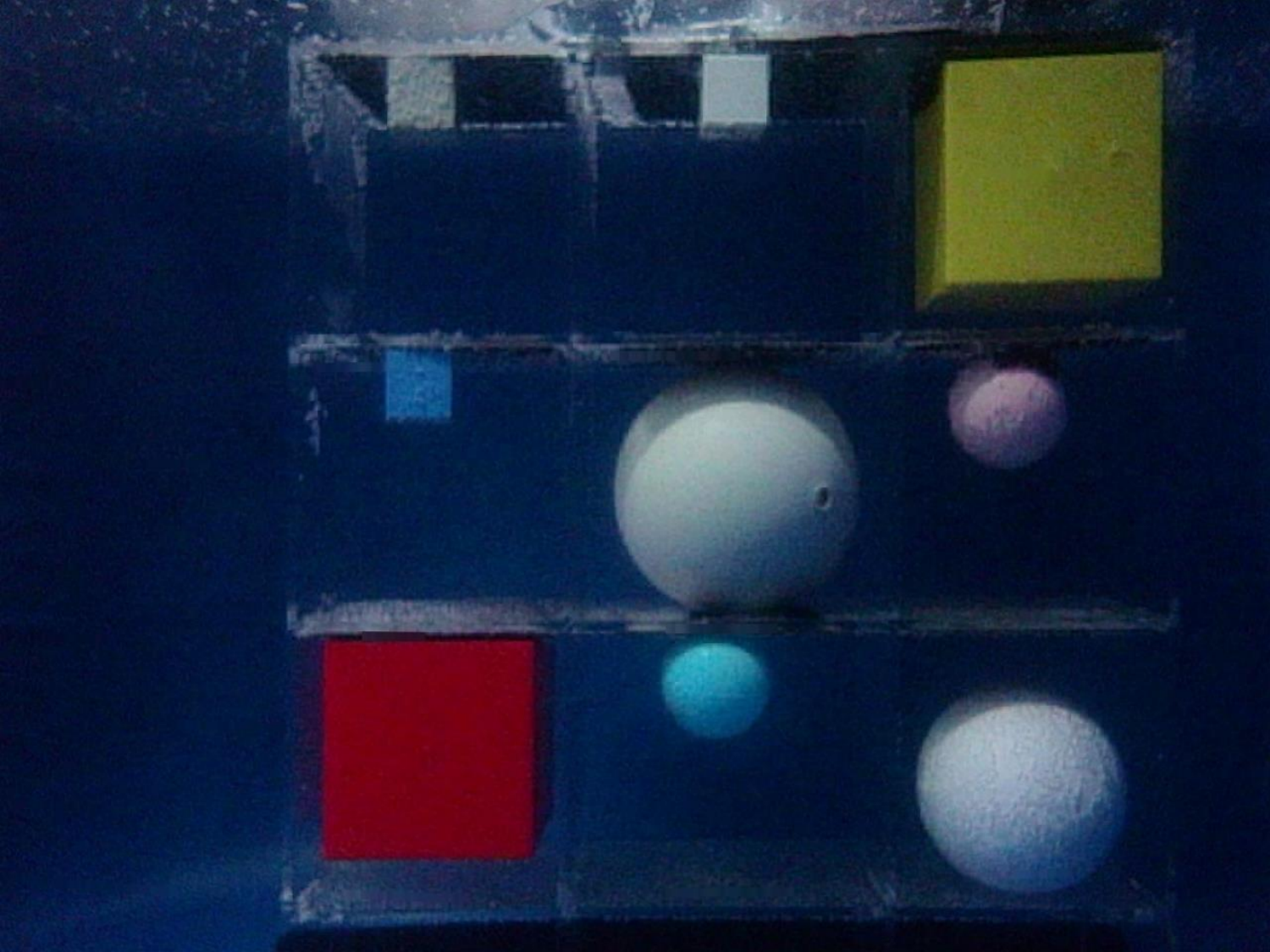}}
\caption{Left view images acquired in Scene~3 under 4~lux, 6~lux, 8~lux, 10~lux, 12~lux and 25~lux illumination conditions are demonstrated. Image under 2~lux illumination is not covered since it is barely visually distinguishable with its 4~lux counterpart. The gradual increment in image brightness from Fig.~\mbox{\ref{4-lux}} to Fig.~\mbox{\ref{25-lux}} is visually observable.}
\label{dif_illu}
\end{figure}

Scenes~1-3 feature a total of 30 static regular-shaped targets, while Scenes~6-8 encompass a total of 9 static sea animal figure-shaped targets used in  the experiment. We leveraged the controllable LED tubes to adjust the environmental illumination and a light intensity sensor to quantify the lighting conditions (Fig.~\ref{PhyicalExperiment}). Seven illumination conditions were created including 25~lux, 12~lux, 10~lux, 8~lux, 6~lux, 4~lux and 2~lux while 25~lux represents the normal daylight environment and others mimic different illumination levels in the underwater environment. 
We take Scene~3 as an example, and Fig.~\mbox{\ref{dif_illu}} demonstrates Scene~3 under various illumination conditions.

\textbf{Target segmentation with a large vision model.} To quantitatively evaluate the performance of the SAM-based segmentation in our underwater environments, we adopted the Intersection over Union (IoU) metric. This metric is calculated by defining the number of pixels that appear both in Ground Truth (GT) and the predicted Segmentation mask (S) as True Positive (TP), the number of pixels that appear in S but not in GT as False Positive (FP), and the number of pixels that appear in GT but not in S as False Negative (FN). The IoU is then calculated as

\begin{equation}
    {IoU} = \frac{TP}{TP+FP+FN}.
\end{equation} 

\begin{figure}[thpb]
\centering
\subfigure{
\label{seg-cube1}
\includegraphics[width=0.15\textwidth]{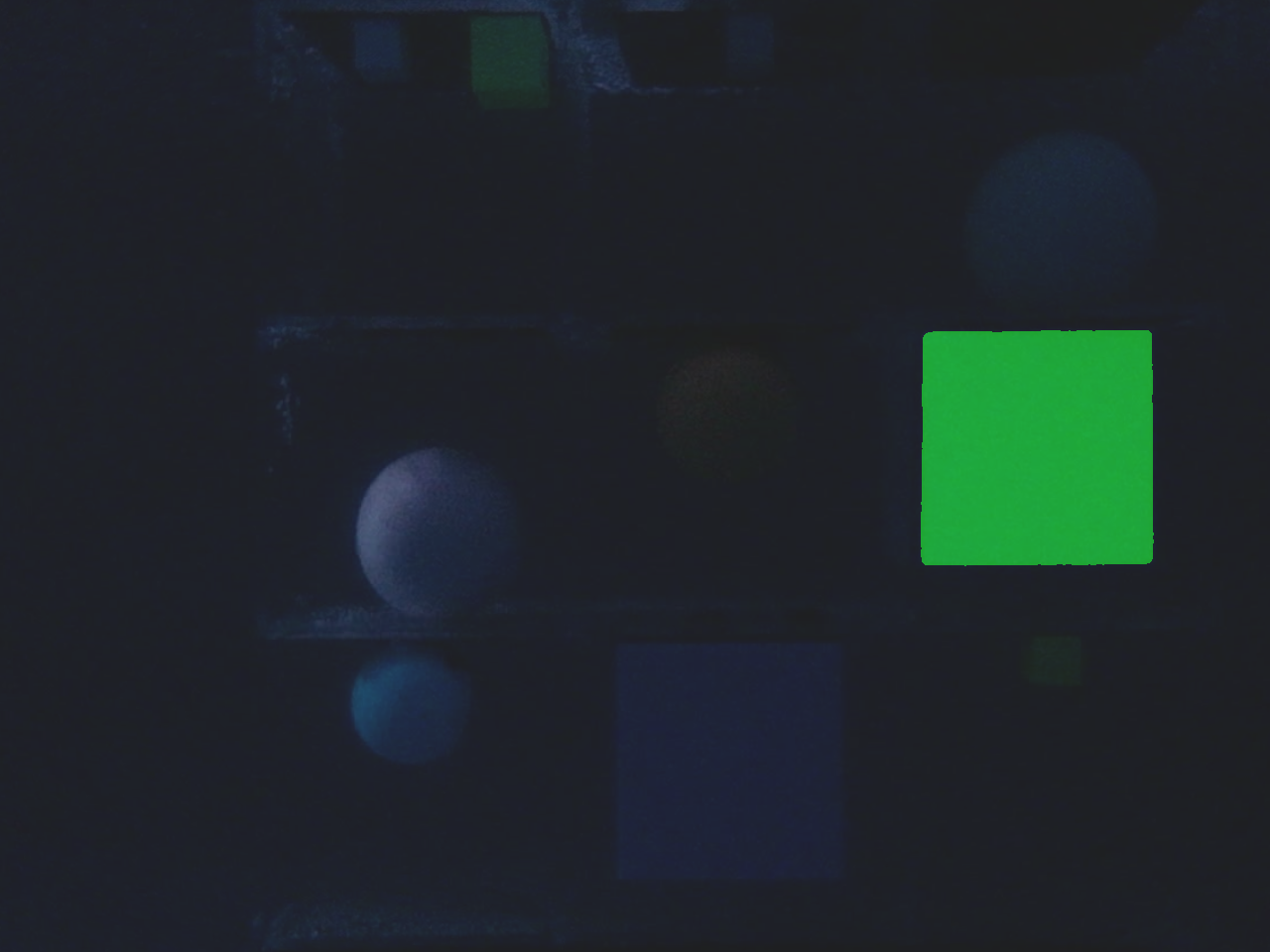}}
\subfigure{
\label{seg-cube2}
\includegraphics[width=0.15\textwidth]{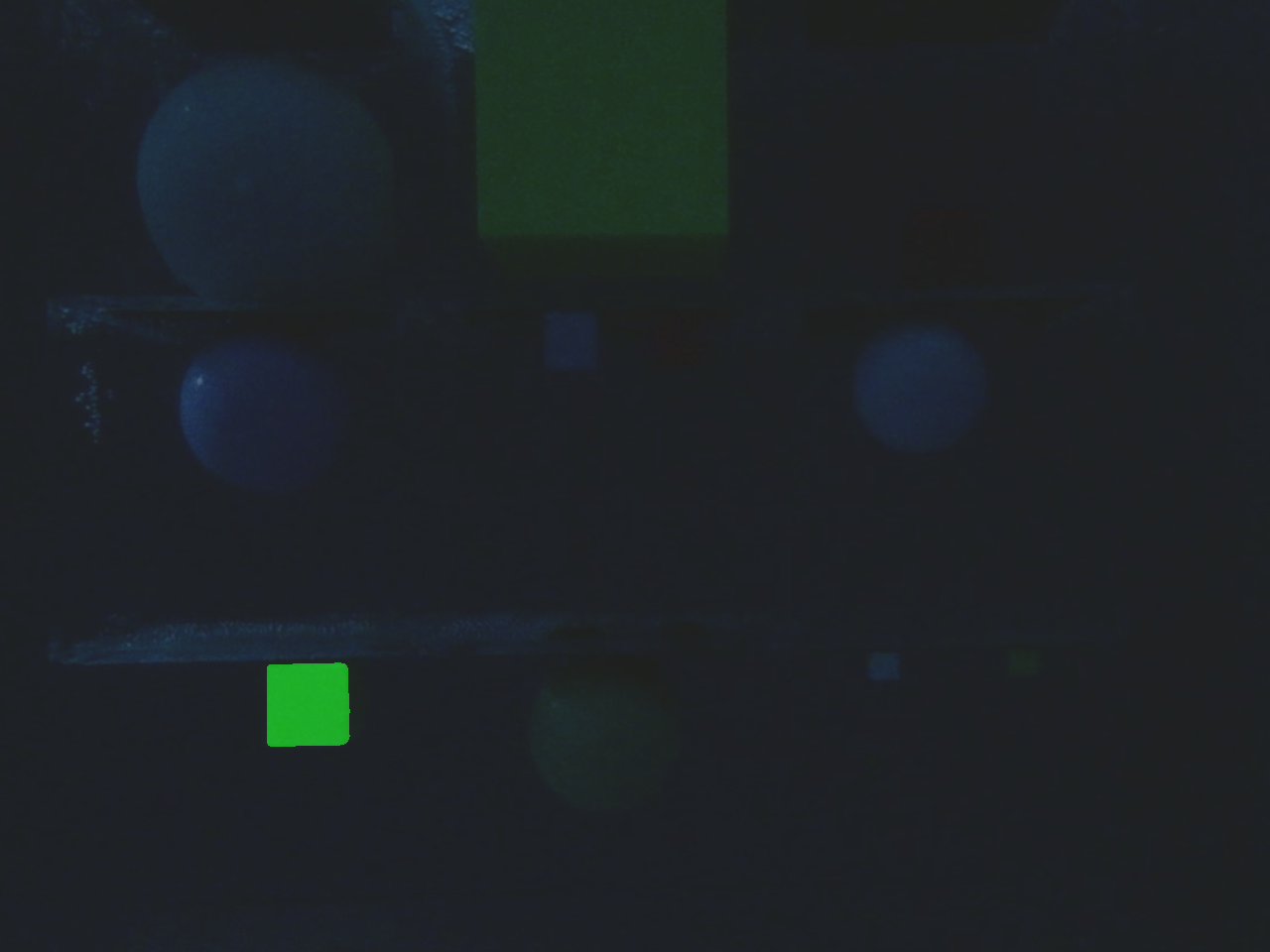}}
\subfigure{
\label{seg-cube3}
\includegraphics[width=0.15\textwidth]{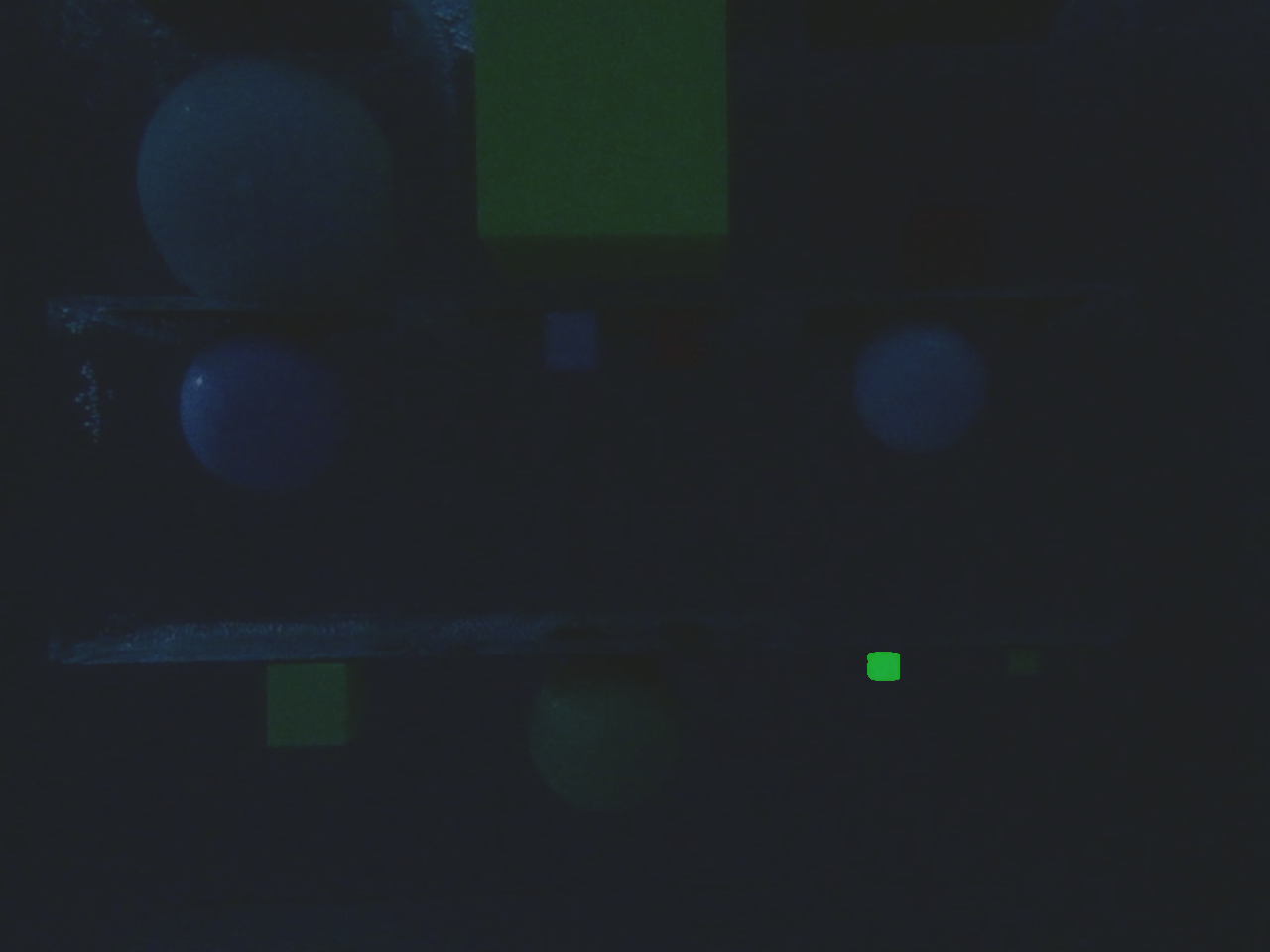}}
\subfigure{
\label{seg-sphere1}
\includegraphics[width=0.15\textwidth]{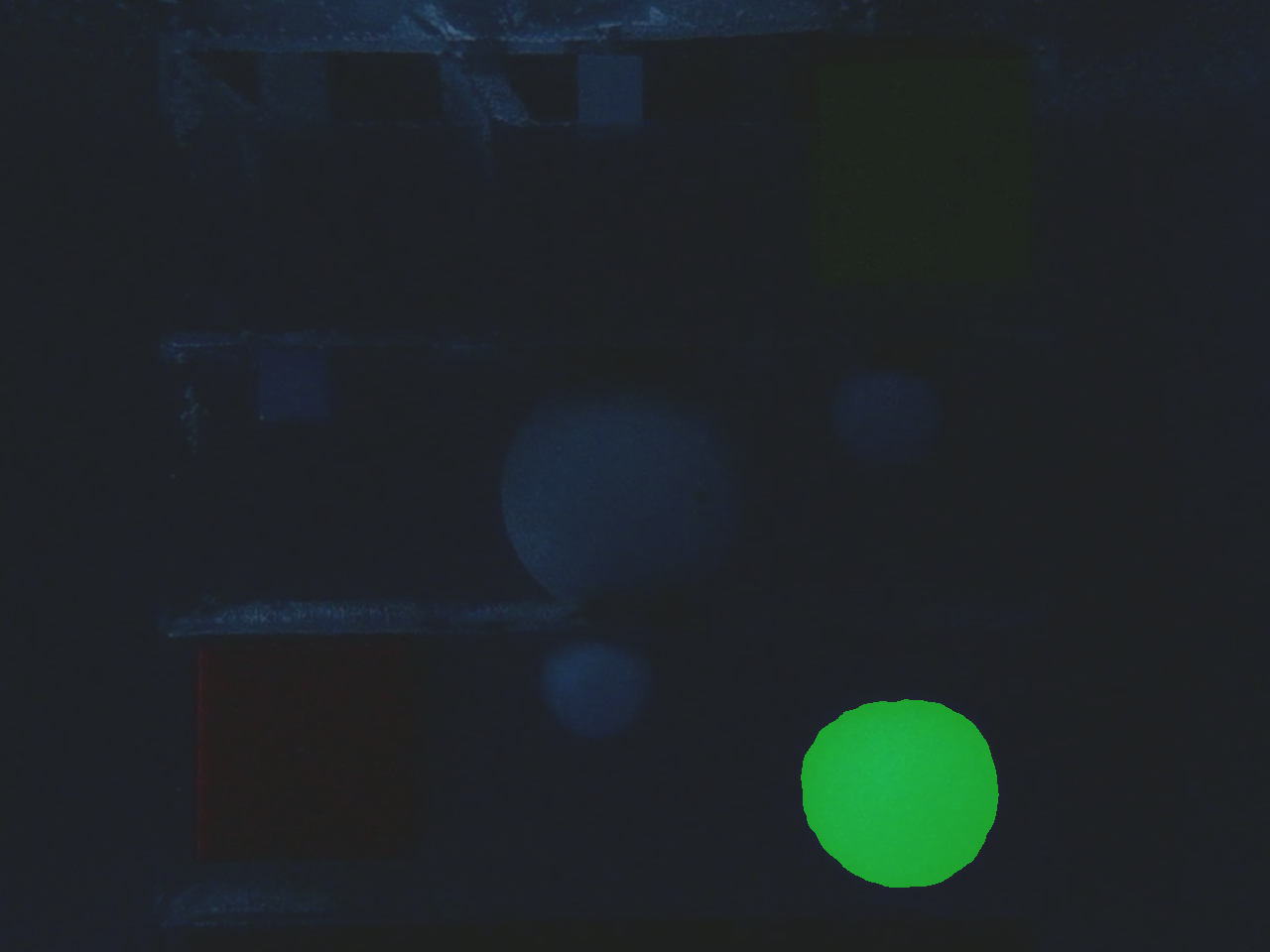}}
\subfigure{
\label{seg-sphere2}
\includegraphics[width=0.15\textwidth]{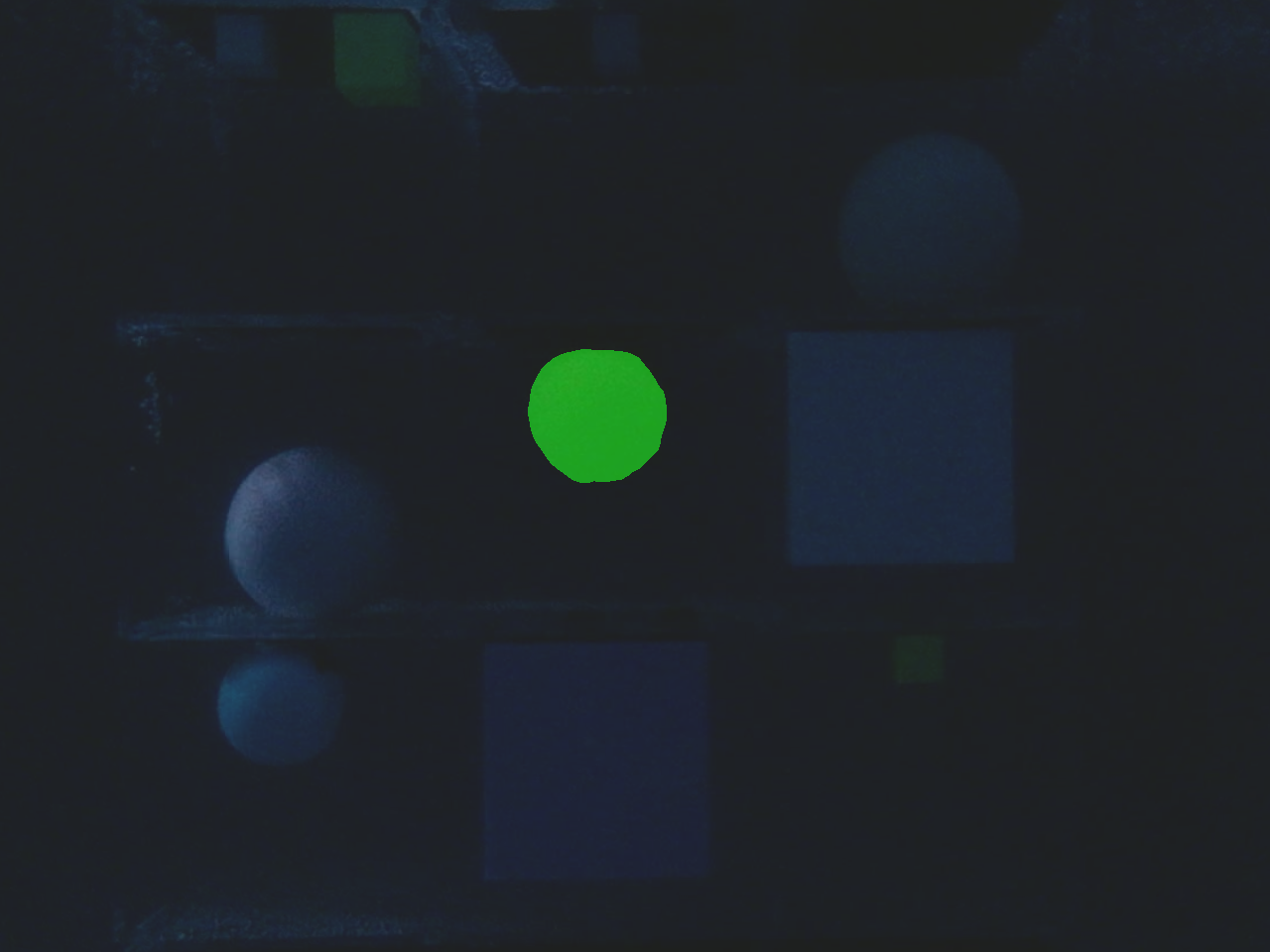}}
\subfigure{
\label{seg-sphere3}
\includegraphics[width=0.15\textwidth]{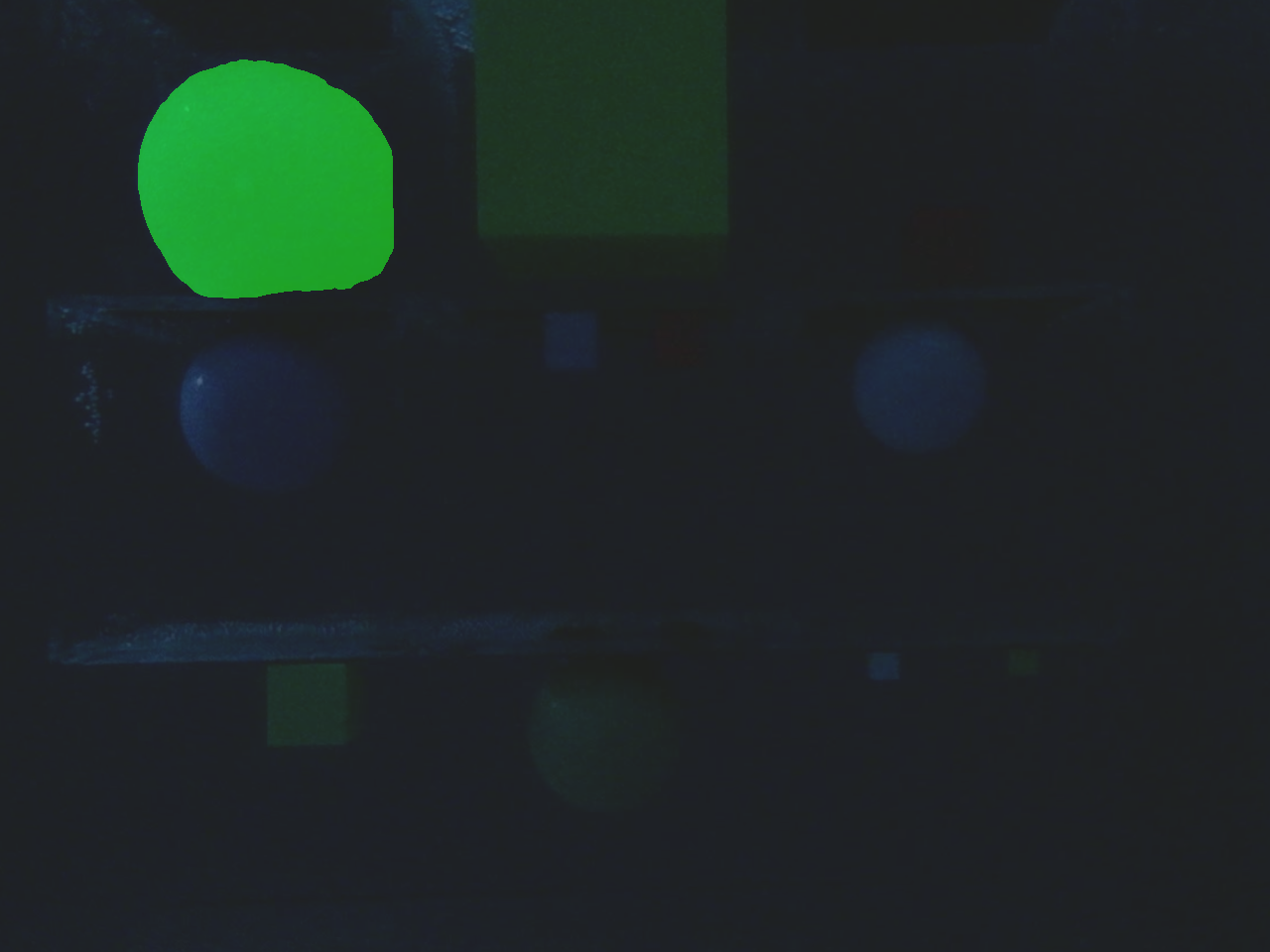}}
\subfigure{
\label{seg-marinelife1}
\includegraphics[width=0.15\textwidth]{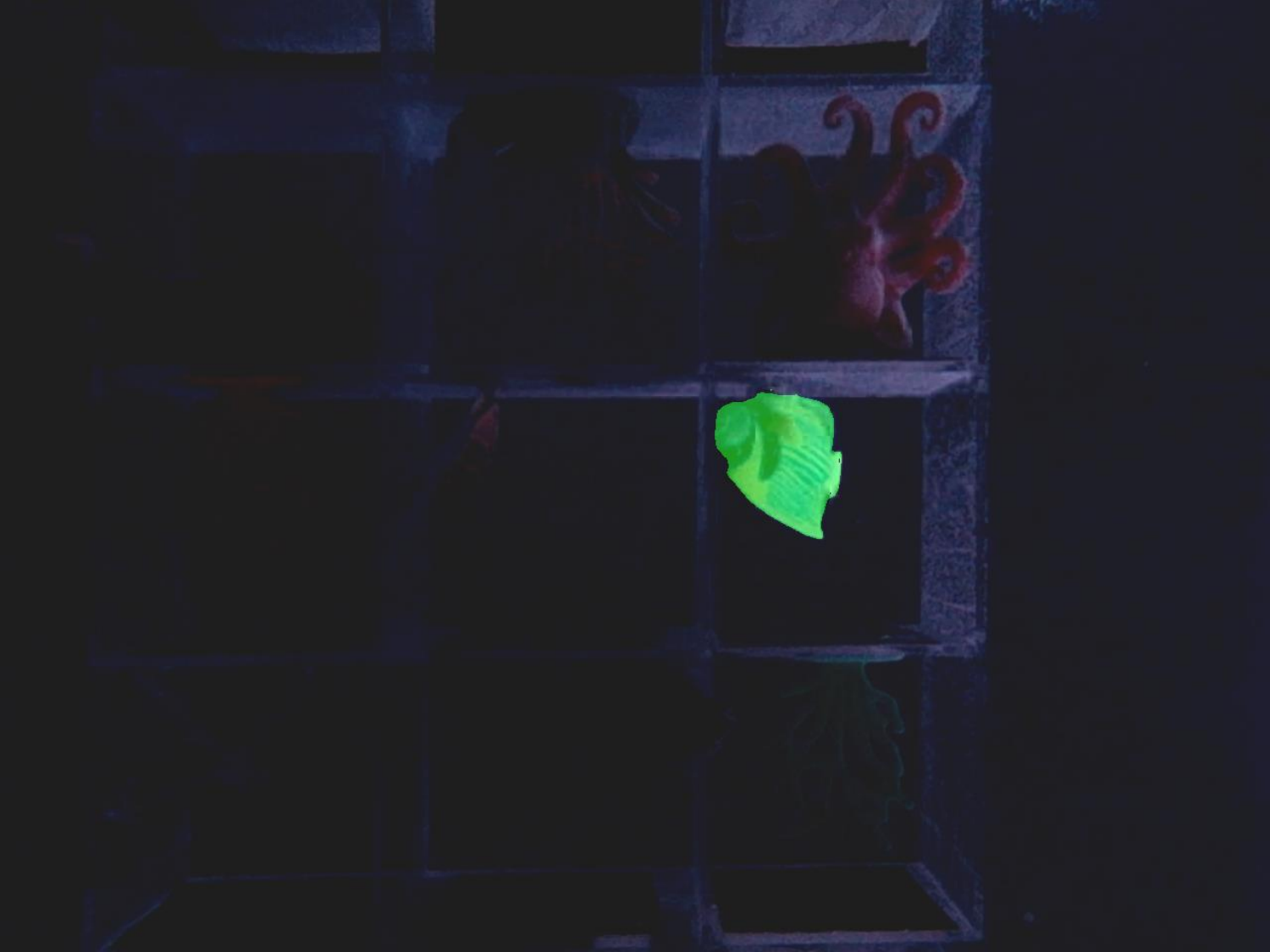}}
\subfigure{
\label{seg-marinelife2}
\includegraphics[width=0.15\textwidth]{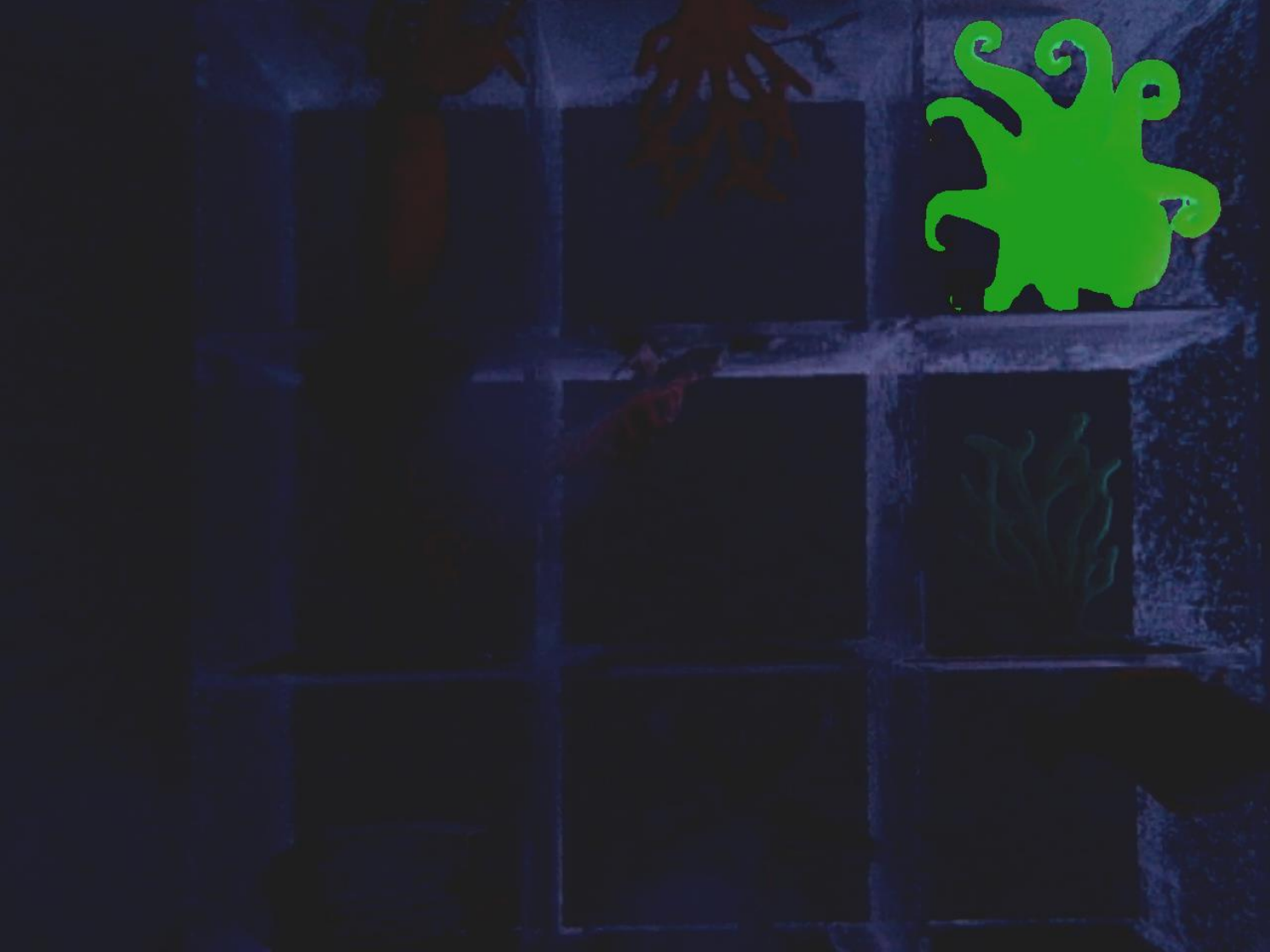}}
\subfigure{
\label{seg-marinelife3}
\includegraphics[width=0.15\textwidth]{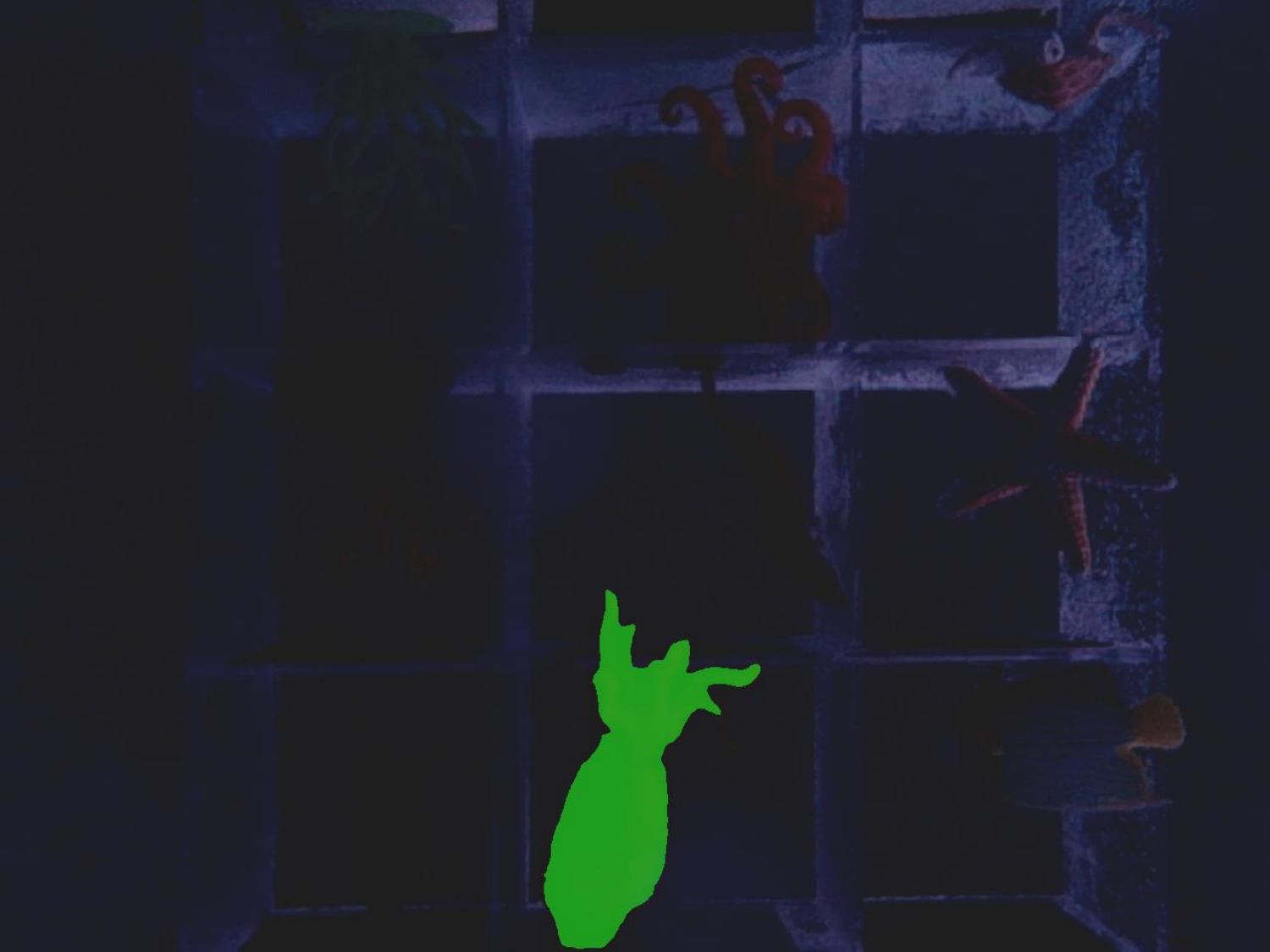}}
\caption{Illustration of the segmentation experimental results using the large vision model --- SAM with the ranging sensor measurements as prompt inputs. The segmentation masks are superimposed on the original images. Three segmented cube cases in the first row, three segmented sphere cases in the second row and three segmented aquatic life model cases in the third row 
are demonstrated.}
\label{seg-example}
\end{figure}

When the IoU between the segmentation mask and the ground truth is lower than 50$\%$ for any single target, we consider the target segmentation as a failure. In addition, if the EMC (Eq. \ref{EMC}) is not satisfied which indicates a target matching failure in the left and right view images, the segmentation is considered as a failure. Otherwise, we have a successful segmentation. Examples of segmentation are demonstrated in Fig. \ref{seg-example} with masks superimposed on the original images.

Figure~6 demonstrates that segmentation failure is absent at illumination levels of 8~lux or higher and 10~lux or higher for regular-shaped targets and sea animal figure-shaped targets, respectively. However, as illumination intensity decreases, failure rates increase significantly --- approximately 2\% at 6~lux, 12\% at 4~lux, and over 70\% at 2~lux for regular-shaped targets. Similarly, for sea animal figure-shaped targets in Scenes~4-6, segmentation failure rates rise slightly at illumination levels ranging from 2~lux to 8~lux compared with regular-shaped targets, but exhibit a similar trend with decreasing illumination.

\begin{figure}[htpb]
\centering
\subfigure[Segmentation failure rate for regular-shaped targets.]{
\label{failure_regular}
\includegraphics[width=0.425\textwidth]{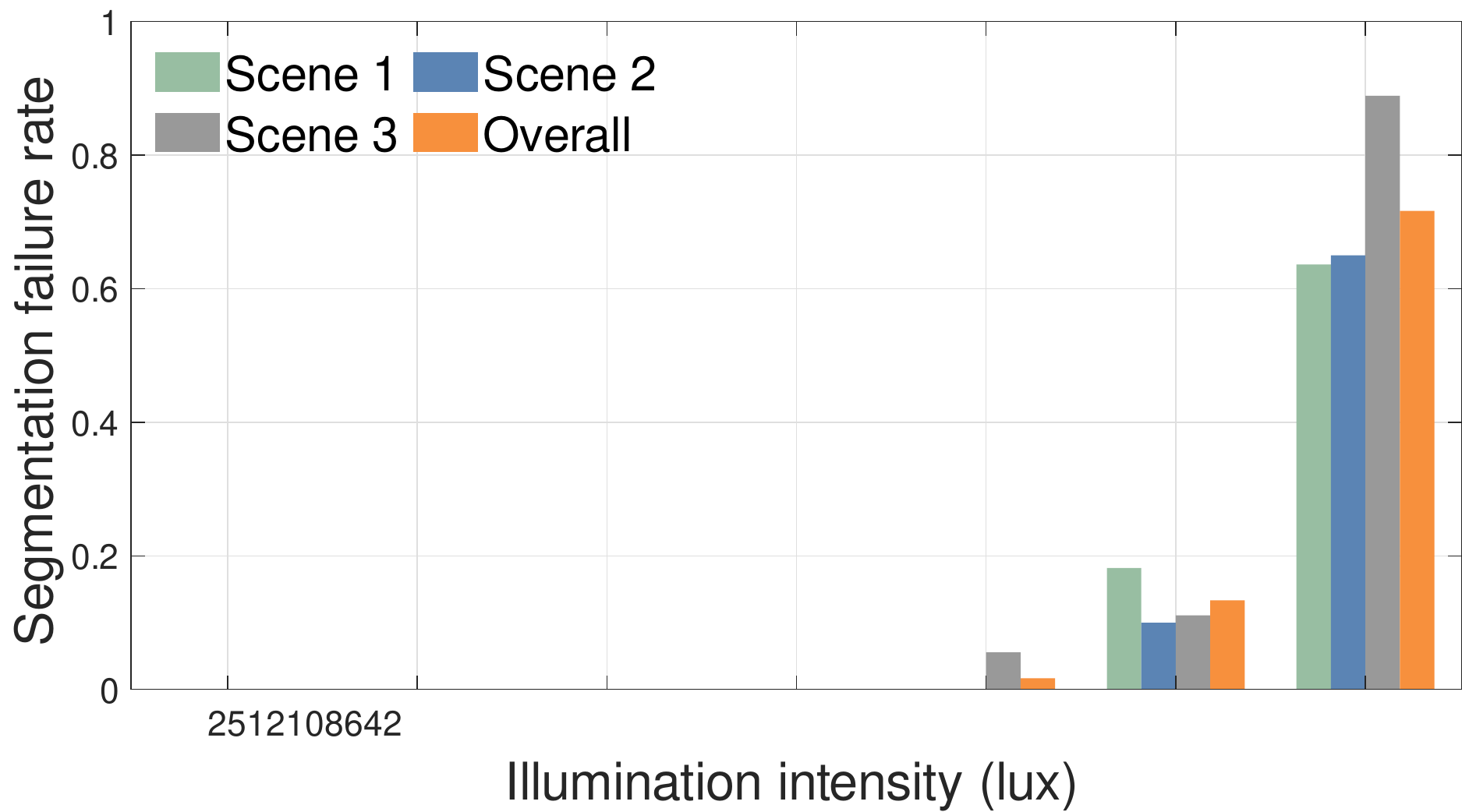}
}
\subfigure[Segmentation failure rate for aquatic life model targets.]{
\label{failure_marinelife}
\includegraphics[width=0.425\textwidth]{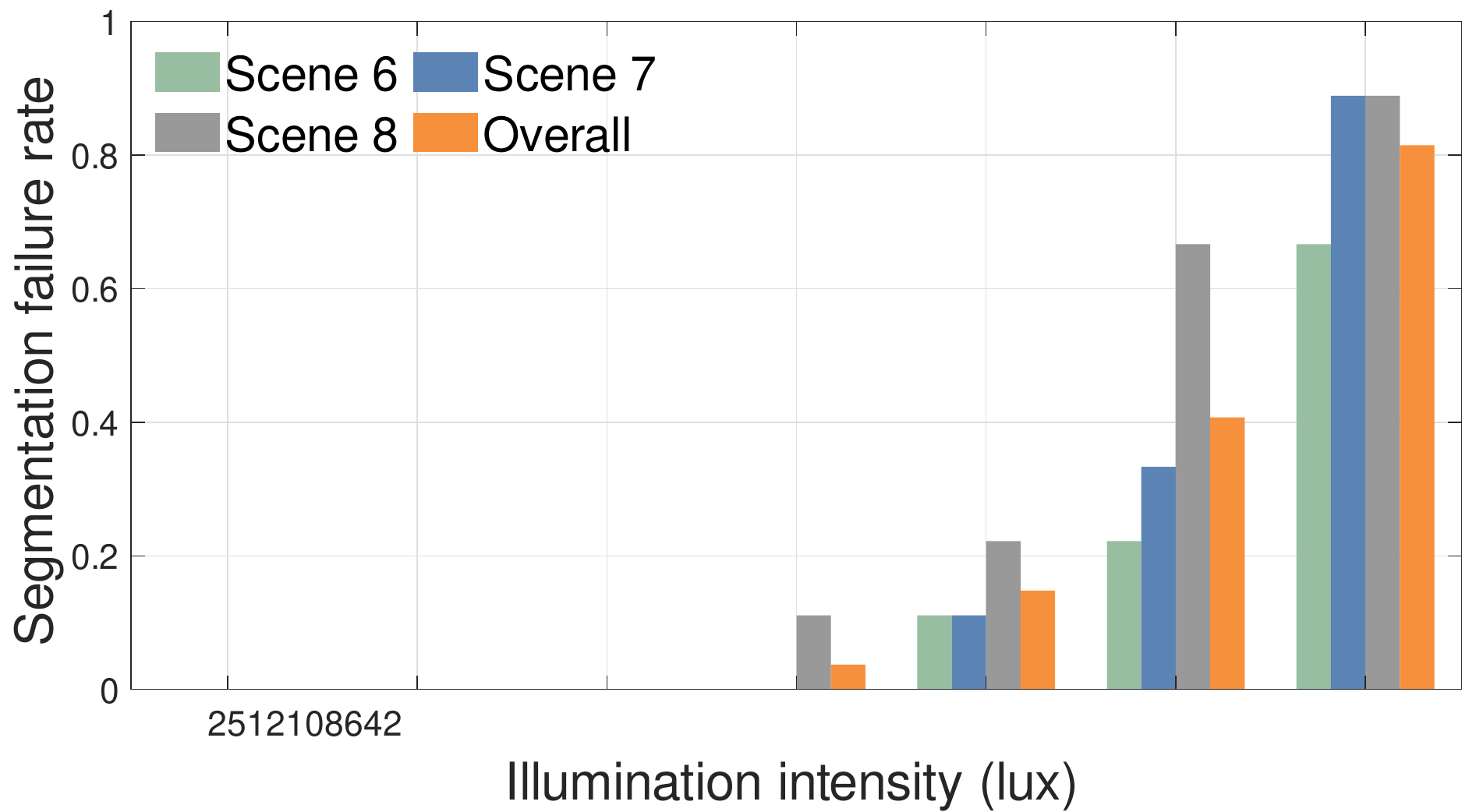}
}
\caption{Experimental results of the statistical segmentation failure rate with respect to the illumination intensity in the testing scenes. (a) presents the statistical results for Scenes~1-3, and (b) presents the statistical results of Scenes~6-8.}
\label{failure}
\end{figure}

\begin{figure}[thpb]
\centering
\subfigure[Segmentation IoU for regular-shaped targets.]{
\label{iou_regular}
\includegraphics[width=0.425\textwidth]{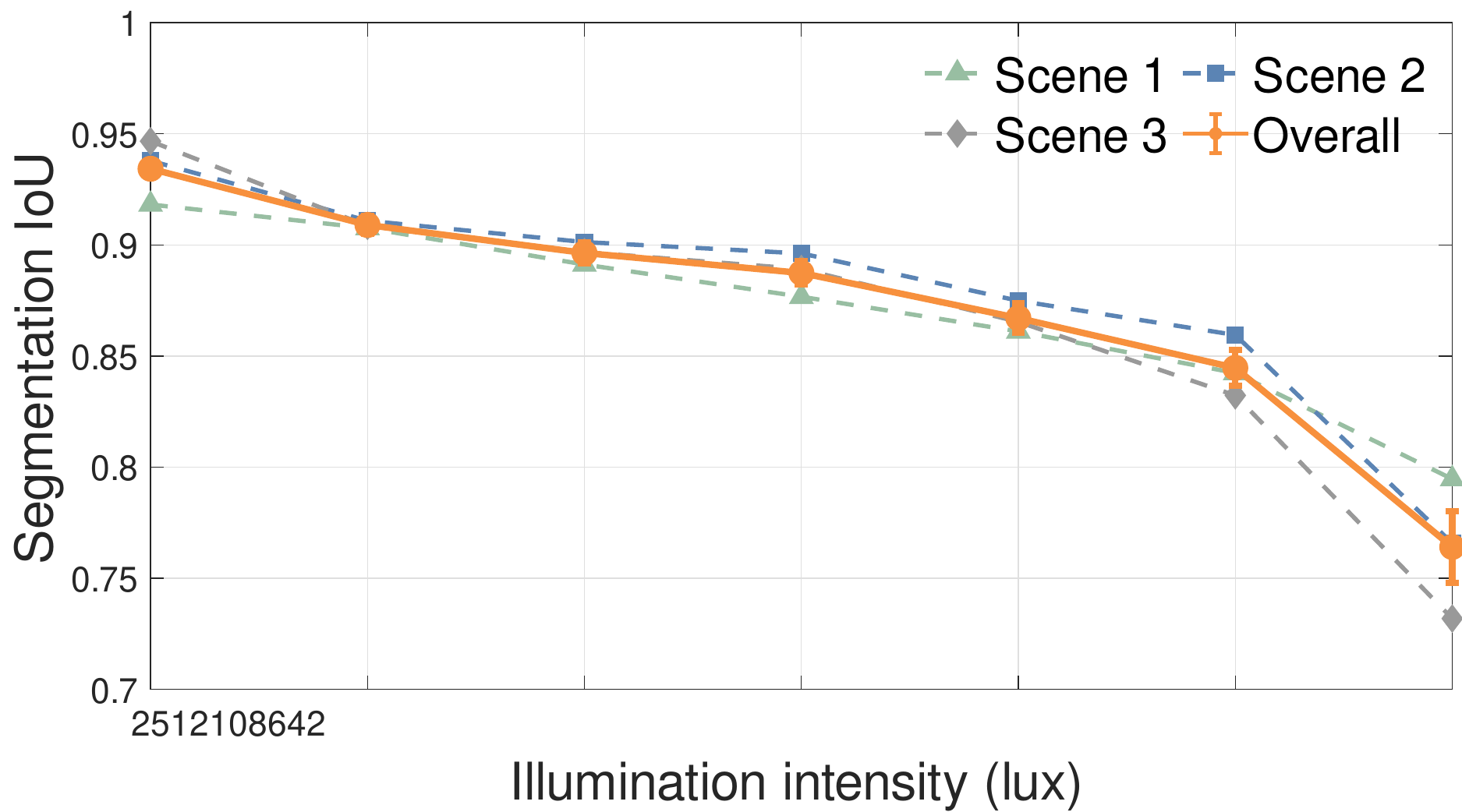}
}
\subfigure[Segmentation IoU for aquatic life model targets.]{
\label{iou_marinelife}
\includegraphics[width=0.425\textwidth]{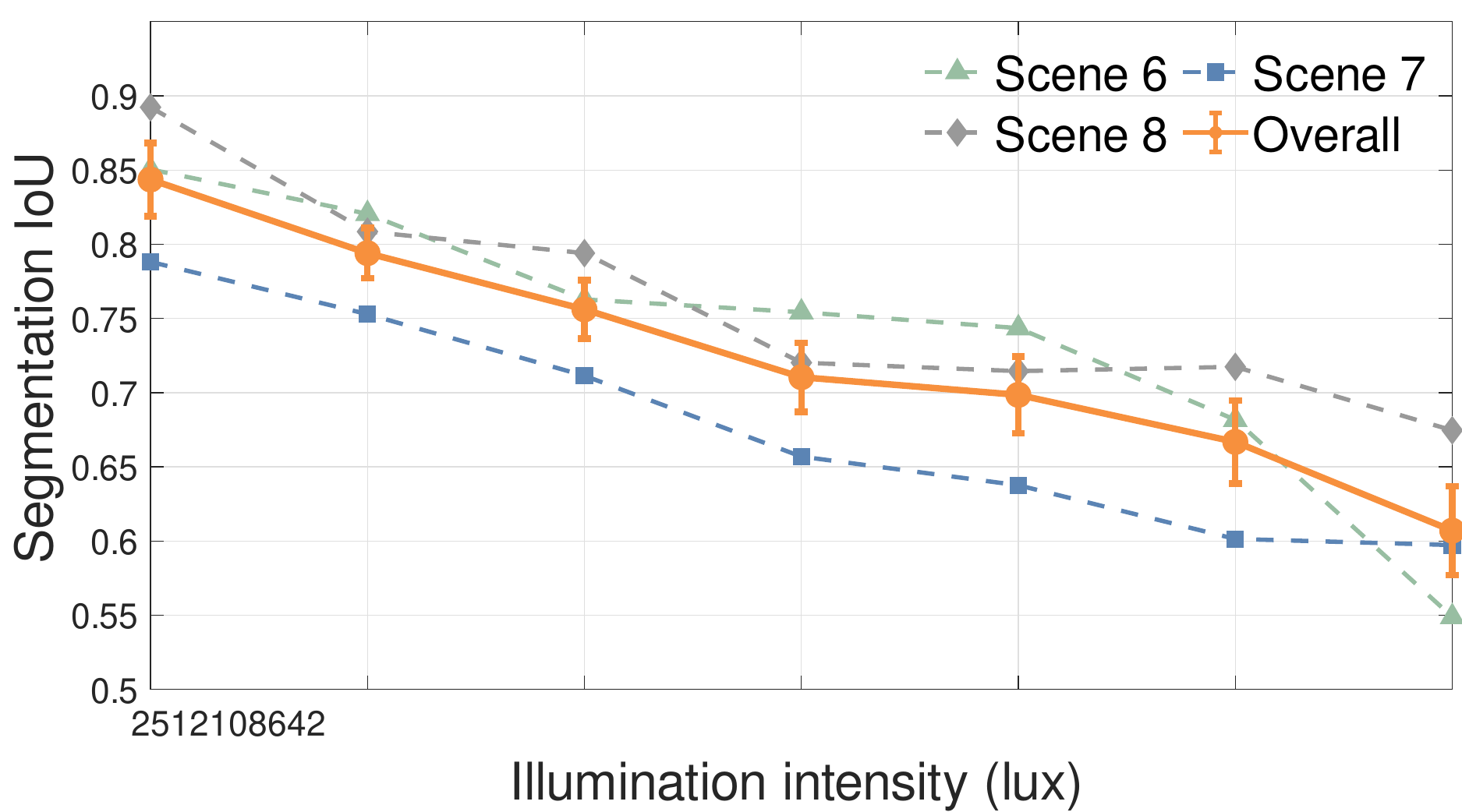}
}
\caption{The experimental results of the segmentation IoU with respect to the illumination intensity. The dash lines in (a) and (b) represent the averaged IoU of all the targets in each scene from Scenes~1-3 and 6-8, respectively. The error bars of Overall in (a) and (b) represent the averaged value and the standard deviation of the IoU of all the targets combined in Scenes~1-3 and 6-8, respectively.}
\label{Seg_IoU}
\end{figure}

\begin{figure}[thpb]
\centering
\subfigure[Distance estimate error for regular-shaped targets.]{
\label{diserr_regular}
\includegraphics[width=0.425\textwidth]{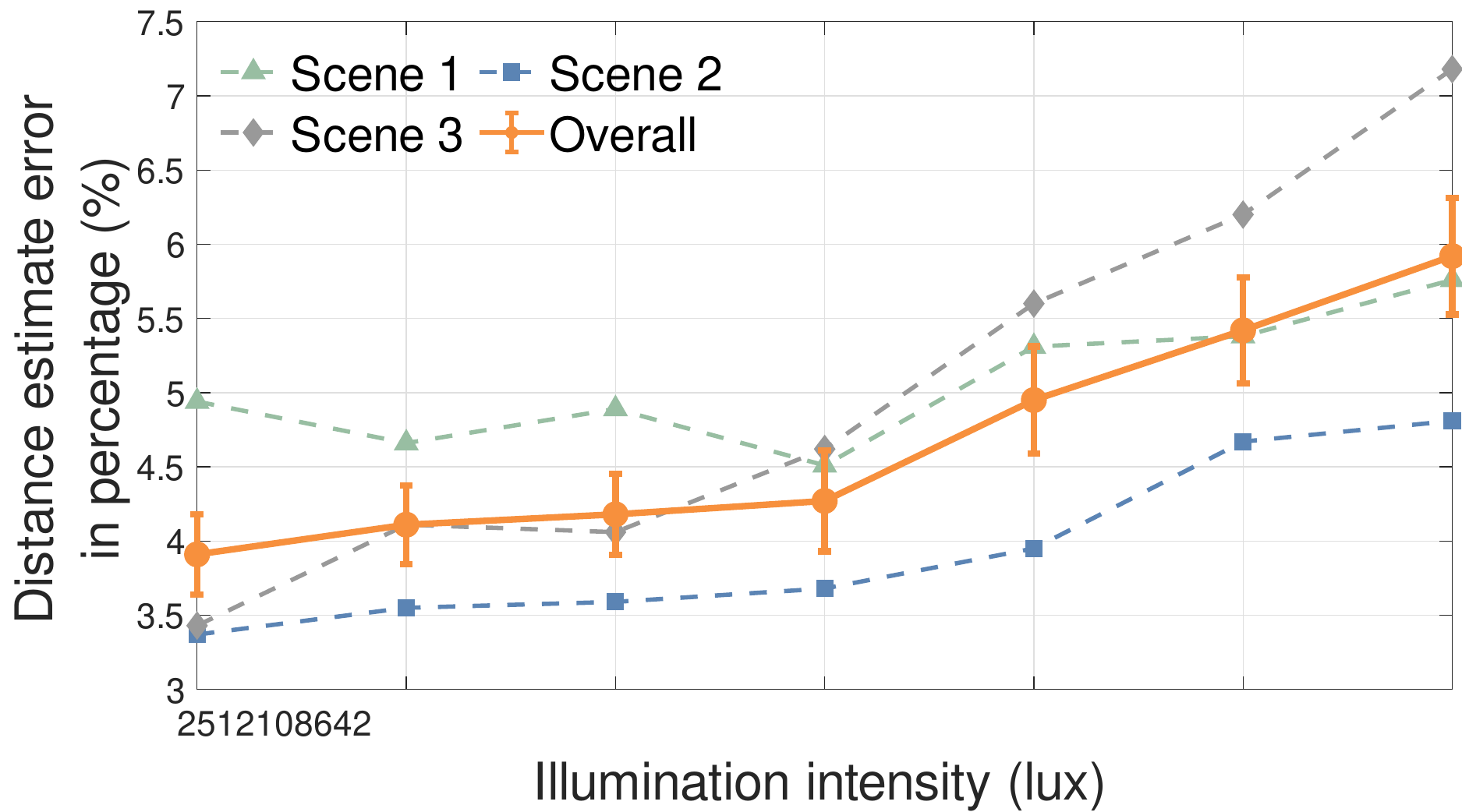}
}
\subfigure[Distance estimate error for aquatic life model targets.]{
\label{diserr_marinelife}
\includegraphics[width=0.425\textwidth]{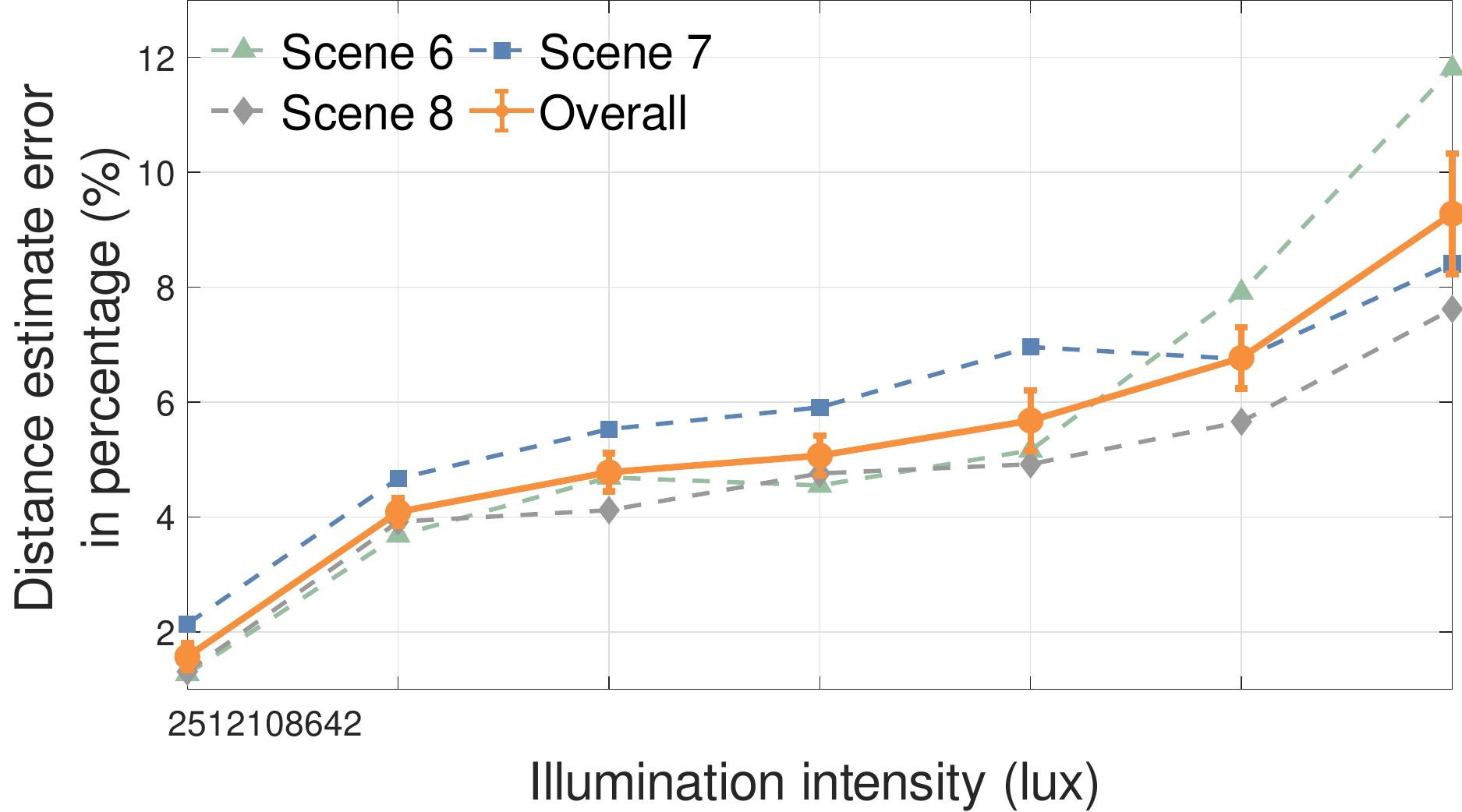}
}
\caption{The experimental results of the distance estimation percentage error $\bar{e_b}$ with respect to the illumination intensity. (a) and (b) illustrate the distance estimation errors for Scenes~1-3 and 6-8, respectively. $\bar{e_b}$ is calculated based on all the targets in one testing scene. The error bars of Overall represent the averaged value and the standard deviation of $\bar{e_b}$ among all three testing scenes.}
\label{dis_err}
\end{figure}

\textbf{Distance estimation by stereo vision.} As the illumination decreases, not only the SAM-based segmentation success rate decreases but also its performance in terms of the segmentation accuracy, which results in an increased distance estimation error $\bar{e_b}$ (Eq.~\ref{eq:e_b}). Figures~\ref{Seg_IoU} and \ref{dis_err} present the experimental results of the segmentation IoU and the distance estimation percent error with respect to the illumination intensity, respectively. In all the testing scenes, the averaged segmentation accuracy consistently decreases when the illumination intensity decreases. We observe that the IoU exceeds 90\% for regular-shaped targets and over 80\% for sea animal figure-shaped targets at 25~lux, dropping to approximately 75\% for regular-shaped targets and 60\% for sea animal figure-shaped targets at 2~lux (calculated based on successful target segmentation only). Regarding distance estimation using stereo vision, the percentage error consistently increases from 4\% for regular-shaped targets and 1.5\% for sea animal figure-shaped targets at 25~lux to 5.5\% for regular-shaped targets and 9\% for sea animal figure-shaped targets at 2~lux. In addition, target size and shape influence segmentation accuracy, with larger targets and cubes yielding more accurate segmentation than smaller ones and spheres, respectively. Segmentation results are also impacted by illumination angle due to the existence of shadows, particularly noticeable for spheres. Sea animal figure-shaped targets exhibit more sensitivity in distance estimation to varying illumination conditions compared to regular-shaped targets.

\begin{figure}[thpb]
\centering
\includegraphics[width=0.425\textwidth]{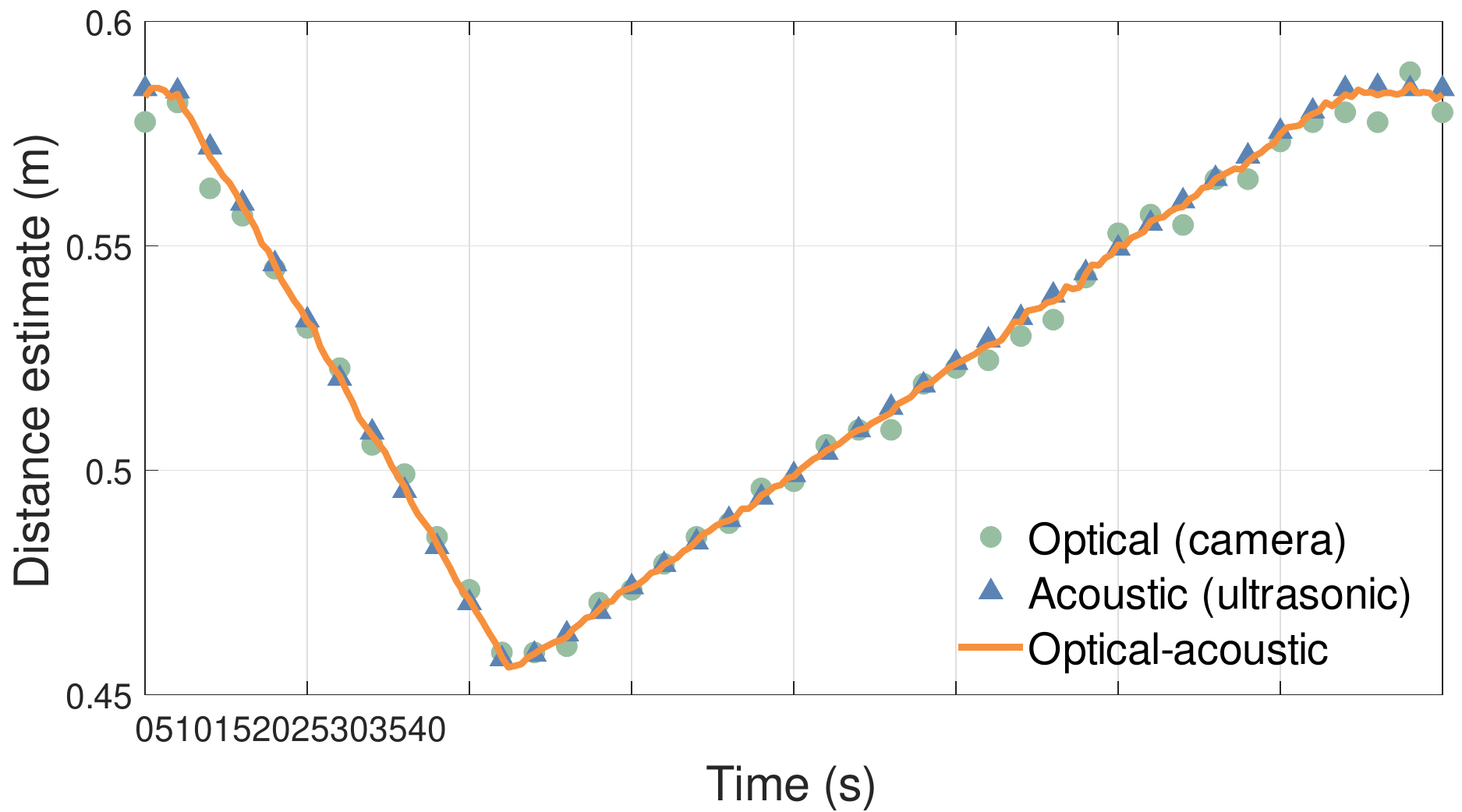}
\caption{Trajectories of the multi-modal estimated distance between the sensing module and the dynamic target in the experiment. The ranging estimation using only the optical measurements (camera) and the acoustic measurements  (ultrasonic) are provided as a comparison. The experimental results shows an enhanced estimation performance with a higher accuracy and a lower variance using both sensing modalities.}
\label{mmdepth}
\end{figure}

\begin{figure}[thpb]
\centering
\includegraphics[width=0.44\textwidth]{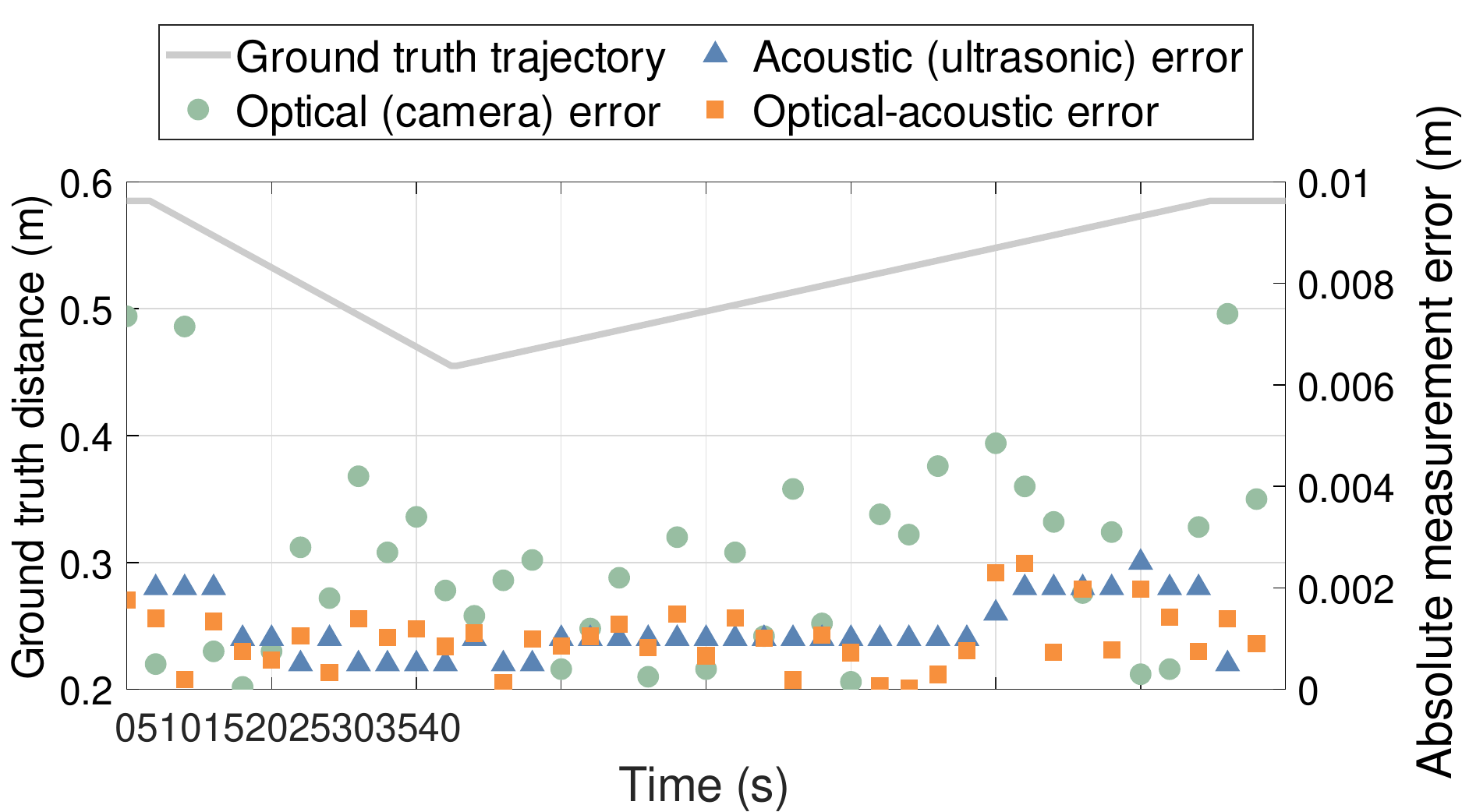}
\caption{The trajectories of absolute sensing errors using both single and multiple sensing modalities of optics and acoustics in Task 2, following a dynamic straight-line motion trajectory. Through utilization of both optical and acoustic sensing modalities, higher estimation accuracy is obtained.}
\label{mmerrcurve}
\end{figure}

\textbf{Comparative Experiments on Prompt Localization} For the comparative experiments of the one-shot prompt localization method, we utilize Scenes~6-8 in Fig.~\mbox{\ref{marinelife_scene6@50}}-\mbox{\ref{marinelife_scene8@60}} as test images and Scenes~9-11 in Fig.~\mbox{\ref{oneshot-1}}-\mbox{\ref{oneshot-3}} as reference images. When the prompt point input falls within the mask of the desired target, the segmentation result resembles the previously mentioned outcomes since they both utilize SAM for segmentation. We consider this process as a successful prompt localization. With a total of 27 targets in Scenes~6-8 taken into account, the accuracy of the prompt localization is 62.96\% (17/27) and 74.07\% (20/27) under 12 lux and 25 lux, respectively. In our proposed method, prompts are provided by acoustic measurements and the localization accuracy, without any filtering method, is 96.17\% and 96.50\% under 25~lux and 12~lux, respectively. The calculated accuracy is based on the same scenes used for the camera-based one-shot method with a total of 120-second measurement length. The accuracy comparison between the pure image-based one-shot method and our proposed acoustic-based method are illustrated in Table \mbox{\ref{loc-compare}}. The comparison results demonstrate that our proposed acoustic-based prompt-localization method consistently provides more accurate and reliable input prompts regardless of the illumination condition.

\begin{table}[thpb]
\caption{Accuracy of prompt localization methods}
\begin{center}
\renewcommand\arraystretch{1.4}
\begin{tabular}{ccc}
\toprule
\textbf{Illumination} & one-shot & \textbf{acoustic-based (ours)}\\
\midrule
\textbf{25~lux} & 74.07\% & \textbf{96.17\%}\\
\textbf{12~lux} & 62.96\% & \textbf{96.50\%}\\
\bottomrule
\end{tabular}
\end{center}
\label{loc-compare}
\end{table}

\subsubsection{Task 2}
This experimental task estimates the time-varying distance between a moving target and the sensing module per Section~\ref{sec:weightavg}. A weighted averaging filter balances between the acoustic and optical sensor measurements with the design parameter $\alpha$ calculated based on the stereo camera ranging accuracy and the acoustic ranging sensor accuracy. The experiment adopted the 4~lux illumination condition where the large vision model SAM occasionally fails the image segmentation. We selected two large cubic targets (as shown in Fig. \ref{dynamic_scene4}) that typically lead to a higher segmentation success rate than other sized and/or shaped objects. Such an experimental setup is expected to alleviate overwhelming segmentation failures and help us to focus on the performance evaluation of the multi-modal ranging design for dynamic targets.
\par The distance estimation percentage error of the binocular camera under 4~lux illumination  (Fig. \ref{dis_err}) is $\bar{e_\text{b}}=5.42\%$ and that of the acoustic ranging sensor is $\bar{e_\text{r}}=1.75\%$. By Eq.~(\ref{calalpha}), we calculate the weighting parameter $\alpha=0.24$. In this experiment, we selected two travelling speeds of the moving target, $1.25\times10^{-2}$~m/s and $5\times10^{-3}$~m/s when moving towards and farther away from the sensing module, respectively. Fig.~\ref{mmdepth} shows the trajectory of the averaged ranging estimation error using the stereo vision, acoustic ranging, and both sensing modalities. The difference among the absolute measurement errors of the three sensing methods are demonstrated in Fig.~\mbox{\ref{mmerrcurve}}, where the measurement error data are resampled at a frequency of 1 Hz. From the experimental result, the averaged ranging estimation error over time using the binocular camera and the ultrasonic ranging sensors separately are 0.45\% and 0.21\%, respectively. The averaged ranging estimation error of the fused optical and acoustic measurements is 0.18\%, thus providing a more accurate estimate than using either sensing modality alone.

\begin{table*}[htbp]
\caption{The estimation errors in the target position and velocity states averaged over the entire motion process.}
\label{part3-errorcomparisonp}
\begin{center}
\renewcommand\arraystretch{1.4}
\begin{tabular}{cccccccc}
\toprule
\textbf{Scene} & \textbf{Estimation} & $\boldsymbol{p_x}$ (m) & $\boldsymbol{p_y}$ (m) & $\boldsymbol{p_z}$ (m) & $\boldsymbol{v_x}$ (m/s) &
$\boldsymbol{v_y}$ (m/s) & $\boldsymbol{v_z}$ (m/s) \\
\midrule
\multirow{2}*{\textbf{Scene 4}} & \textbf{Prior} & $5.3\times10^{-4}$ & $3.1\times10^{-4}$ & $2.8\times10^{-3}$ & $9.3\times10^{-4}$& $6.2\times10^{-4}$ & $1.5\times10^{-3}$\\ & \textbf{Posterior} & $4.2\times10^{-4}$ & $2.3\times10^{-4}$ & $2.3\times10^{-3}$ & $8.3\times10^{-4}$ & $5.1\times10^{-4}$ & $1.1\times10^{-3}$\\ 
\multirow{2}*{\textbf{Scene 5}} & \textbf{Prior} & $2.1\times10^{-4}$ & $1.3\times10^{-4}$ & $1.1\times10^{-3}$ & $3.2\times10^{-4}$ & $2.8\times10^{-4}$& $7.1\times10^{-4}$ \\ & \textbf{Posterior} & $1.8\times10^{-4}$ & $1.0\times10^{-4}$ & $7.2\times10^{-4}$ & $2.8\times10^{-4}$ & $2.1\times10^{-4}$ & $6.6\times10^{-4}$\\
\bottomrule
\end{tabular}
\end{center}
\end{table*}

\subsubsection{Task 3}
\begin{figure}[thpb]
\centering
\subfigure{
\label{ekf-p-s4}
\includegraphics[width=0.42\textwidth]{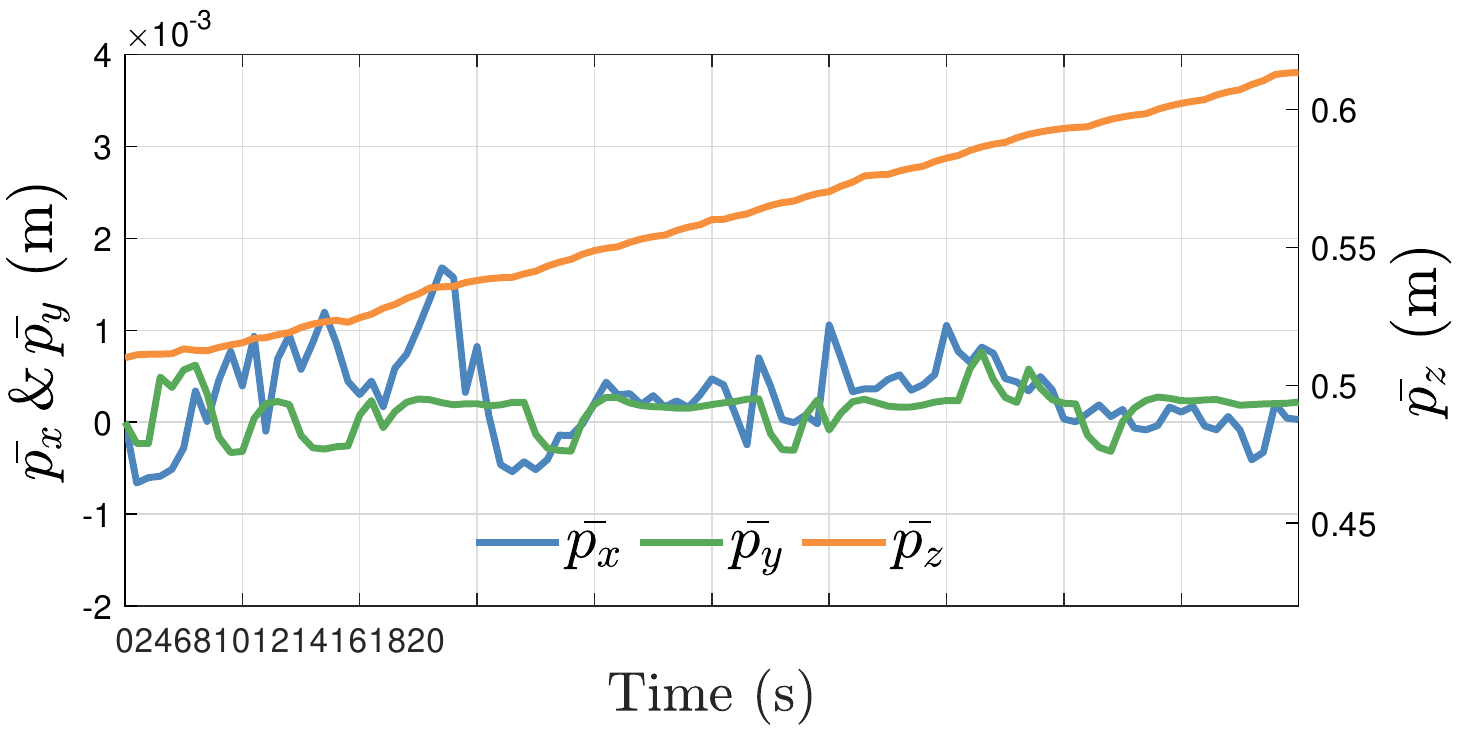}}
\subfigure{
\label{ekf-v-s4}
\includegraphics[width=0.425\textwidth]{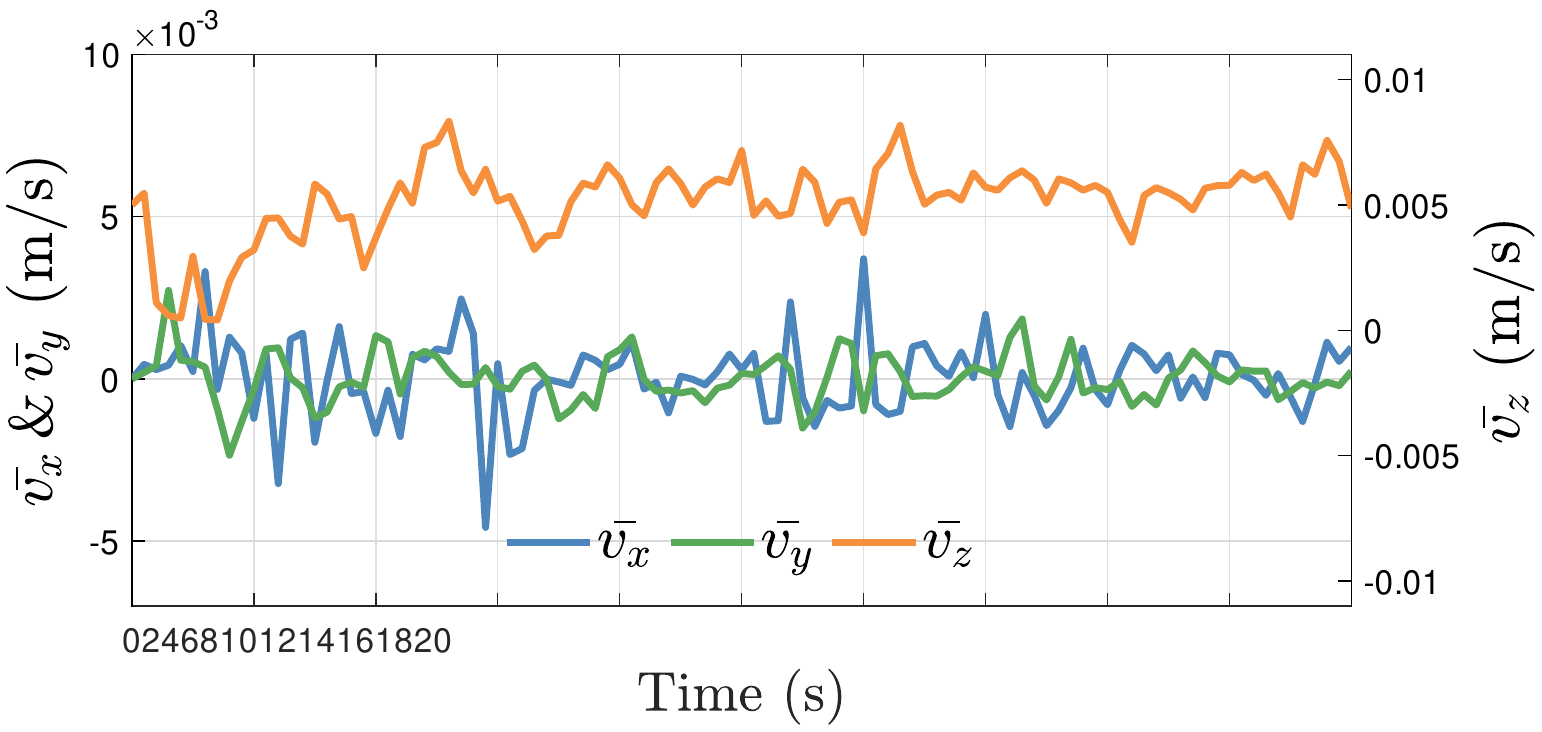}}
\caption{Trajectories of the position and velocity motion state estimates in the experiment in Scene 4 where the targets move father away from the sensing module at the speed of $5\times10^{-3}$~m/s.}
\label{ekf-result-scene4}
\end{figure}

\begin{figure}[thpb]
\centering
\subfigure{
\label{ekf-p-s5}
\includegraphics[width=0.42\textwidth]{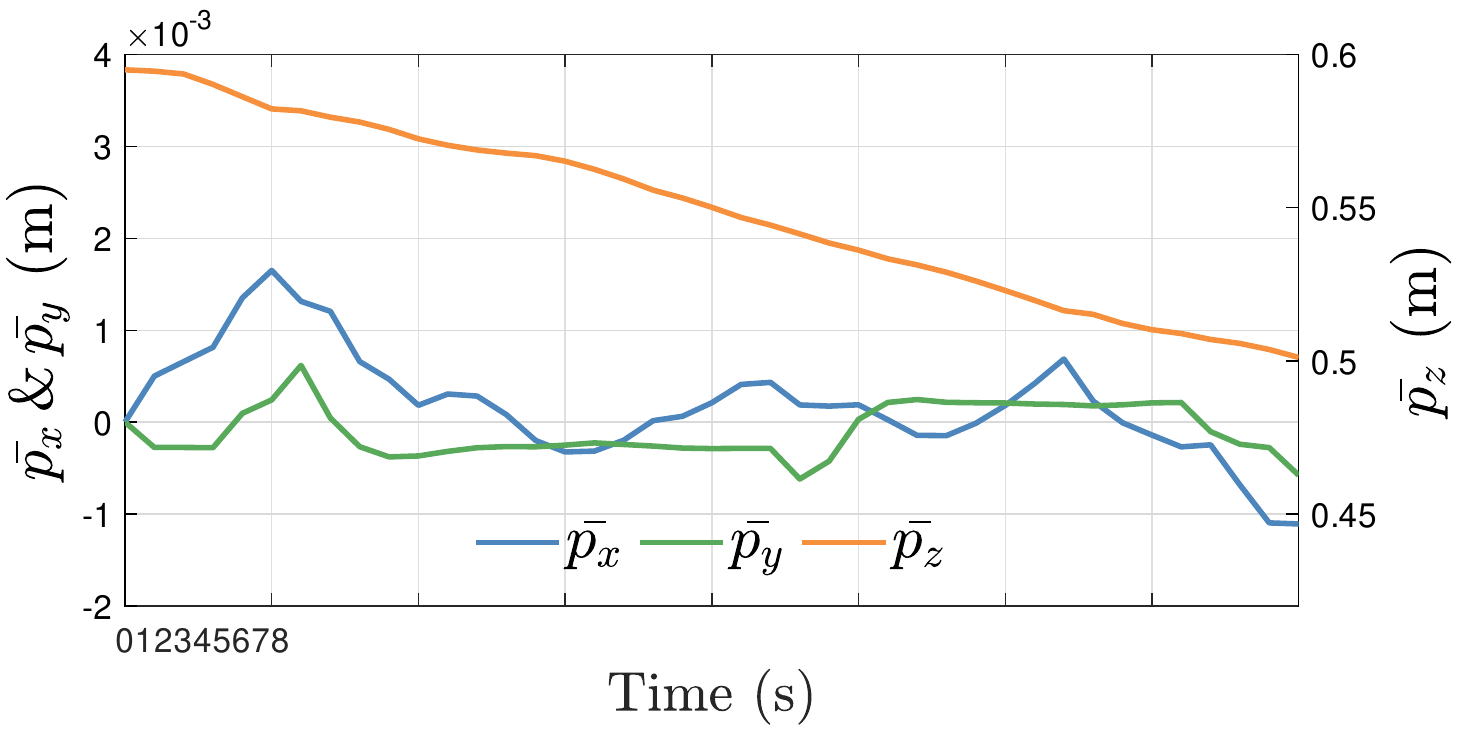}}
\subfigure{
\label{ekf-v-s5}
\includegraphics[width=0.425\textwidth]{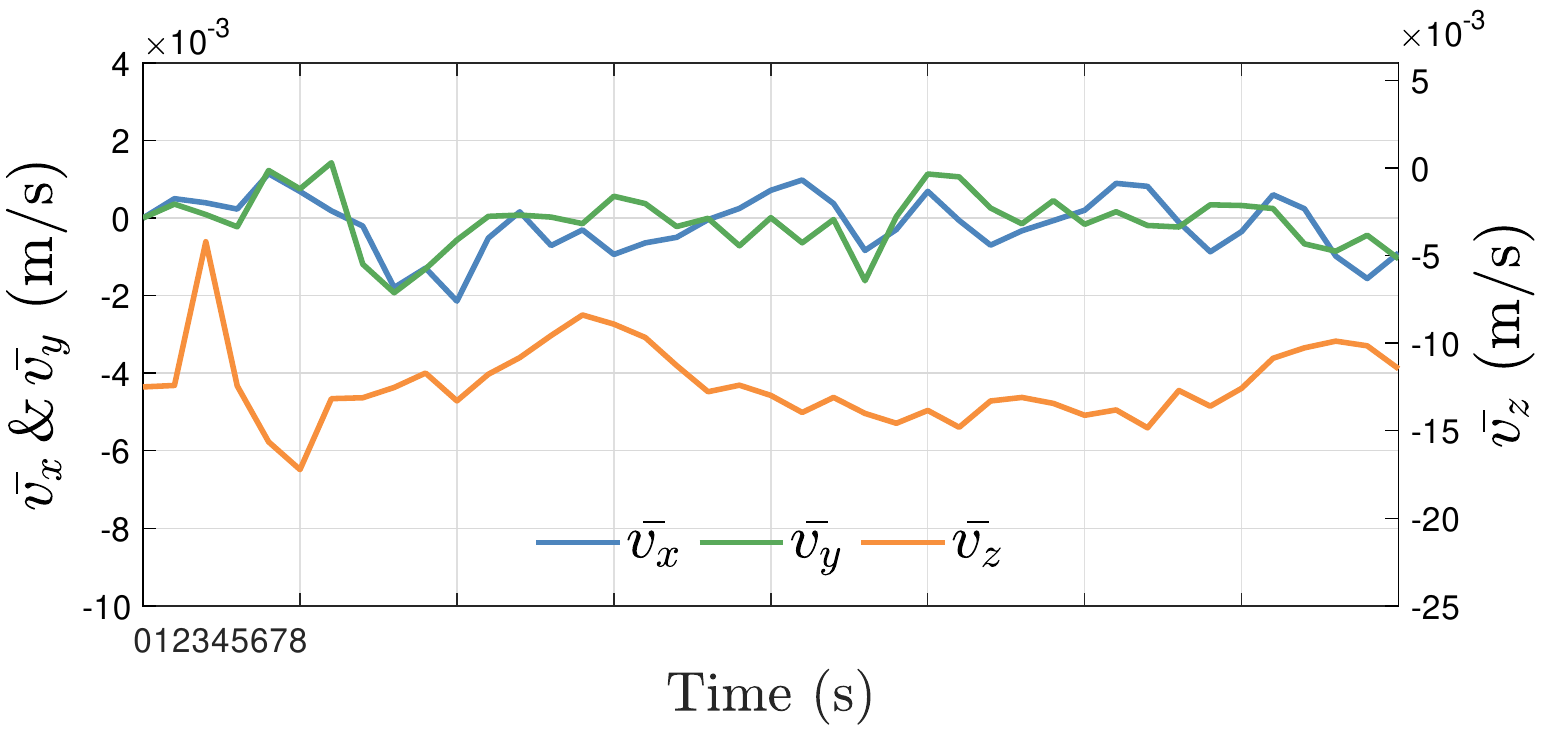}}
\caption{Trajectories of the position and velocity motion state estimates in the experiment in Scene 5 where the targets move towards the sensing module at the speed of $1.25\times10^{-2}$~m/s.}
\label{ekf-result-scene5}
\end{figure}

This task assimilates optical and acoustic sensor measurements via EKF to estimate the 3-D motion states including the position and the velocity of a moving target of interest. The covariance matrix $\textbf{\textit{Q}}$ of the EKF is determined per Section \ref{sec:EKF}. We adopt Scenes 4 and 5 in Fig. \ref{scenes} and use the sliding guide to move the targets along the $z$-axis as shown in Fig. 2. For the convenience of presentation and analysis, we subtract the $x$ and $y$ coordinates of the estimated target position by the actual constant coordinates and redefine the difference as $p_x$ and $p_y$. Consequently, in our experimental setup, the estimated target motion states along the $x$ and $y$ directions, i.e., $p_x$, $p_y$, $v_{x}$ and $v_{y}$ conform to normal distributions with a zero mean. In Scene 4, the acrylic frame that holds two cubic targets moves farther away from the sensing module at a speed of $0.5\times10^{-3}$~m/s for 20 seconds. In Scene 5, the acrylic frame holding three cubic targets moves towards the sensing module at a faster speed of $1.25\times10^{-2}$~m/s for 8 seconds. The averaged position estimates $\bar{p_x}$, $\bar{p_y}$, $\bar{p_z}$ and the averaged velocity estimates $\bar{v_x}$, $\bar{v_y}$, $\bar{v_z}$ over all the targets in each experimental setup are presented in Figs. \ref{ekf-result-scene4} and \ref{ekf-result-scene5} for Scenes 4 and 5, respectively. Furthermore, the estimation errors for each motion state in Scenes 4 and 5 are provided in Table \ref{part3-errorcomparisonp} with both priors and posteriors of the EKF estimation. From the experimental results, we observe a consistent estimation error in both the position and velocity motion states in the magnitude of $10^{-3}$m or less along the $x$ and $y$ axes perpendicular to the direction of travel, and a slightly increased estimation error along the direction of travel. Averaged over all time instants, all the position and velocity state estimation errors are bounded by $2.8 \times 10^{-3}$m and $1.5\times 10^{-3}$~m/s. In addition, by comparing the prior and posterior state estimates, we find that incorporating the multi-modal sensor measurements aligns with our design expectations and generally improves the estimation accuracy, reducing the averaged estimation error by 10\% to 25\%.

\section{Conclusion}

This paper proposed a multi-modal sensing framework to resolve the close-range underwater target localization problem with generalization capability.  A sensing module consisting a stereo vision camera and eight acoustic ranging sensors was designed and developed along with a testing platform. A target localization algorithm was proposed which incorporated image segmentation through a large vision model (SAM) and multi-modal sensor fusion through weighted averaging and the EKF according to the sensing tasks. Extensive experiments were conducted, the results of which validated the effectiveness of the proposed multi-modal sensing framework in 1-D ranging and 3-D motion state estimation for both static and dynamic underwater targets. Furthermore, we experimentally investigated and quantitatively evaluated the influence of the illumination intensity on the target localization performance, aiming to provide important insights into the multi-modal sensing design in underwater environments.

For future work, we will explore the feasibility of replacing SAM with semantic and lightweight large vision models in image segmentation to improve real-time performance. In addition, we plan to install the multi-modal sensing module onto a lab-developed underwater robot and explore the application of the proposed estimation algorithm in the closed-loop motion control of underwater robots.

\bibliography{ref}

\end{document}